\pdfoutput=1

\RequirePackage[l2tabu, orthodox]{nag}

\documentclass[12pt,phd,a4paper,twoside]{ucl_thesis}

\usepackage{blindtext}

\usepackage{emptypage}

\usepackage{graphicx}

\usepackage{float}

\usepackage{amsmath}
\usepackage{amssymb}

\usepackage{gensymb}
\usepackage{textcomp}

\usepackage{setspace}

\usepackage{multirow}

\usepackage{bibentry}

\usepackage[format=hang,font=small,labelfont=bf]{caption}

\usepackage{etoolbox}

\usepackage{lipsum}

\usepackage{xcolor}
\definecolor{nice-red}{HTML}{E41A1C}
\colorlet{dark-red}{nice-red!80!black}
\definecolor{nice-orange}{HTML}{FF7F00}
\colorlet{dark-orange}{orange!85!black}
\definecolor{nice-yellow}{HTML}{FFC020}
\definecolor{nice-green}{HTML}{4DAF4A}
\definecolor{nice-blue}{HTML}{377EB8}
\definecolor{nice-purple}{HTML}{984EA3}

\usepackage[xindy]{glossaries}

\usepackage{booktabs}

\usepackage[linesnumbered,ruled,noend]{algorithm2e}

\usepackage{tikz}
\usetikzlibrary{calc,trees,positioning,arrows,chains,shapes.geometric,%
  decorations.pathreplacing,decorations.pathmorphing,shapes,%
  matrix,shapes.symbols,fit,decorations,arrows.meta}

\PassOptionsToPackage{hyphens}{url}
\usepackage{hyperref}
\hypersetup{
 unicode=false,
 pdftoolbar=true,
 pdfmenubar=true,
 pdffitwindow=false,
 pdfstartview={FitH},
 pdfauthor={Tim Rocktäschel},
 pdfnewwindow=true,
 colorlinks=true,
 linkcolor=black,
 citecolor=black,
 filecolor=black,
 urlcolor=black
}

\usepackage{subcaption}

\usepackage{microtype}

\usepackage{url}

\usepackage{natbib}

\usepackage{soul}

\usepackage{changepage}

\usepackage{xargs}
 
\usepackage{bm}

\usepackage[capitalize]{cleveref}
\crefformat{equation}{Eq.~#2#1#3}
\Crefformat{equation}{Equation~#2#1#3}
\crefrangeformat{equation}{Eqs.~#3#1#4 to~#5#2#6}
\Crefrangeformat{equation}{Equations~#3#1#4 to~#5#2#6}
\crefmultiformat{equation}{Eqs.~#2#1#3}%
{ and~#2#1#3}{, #2#1#3}{ and~#2#1#3}
\Crefmultiformat{equation}{Equations~#2#1#3}%
{ and~#2#1#3}{, #2#1#3}{ and~#2#1#3}

\usepackage{stmaryrd}

\usepackage[outline]{contour}

\usepackage[disable]{todonotes}

\let\newcite\citet

\usepackage{xcolor}
\colorlet{dark-blue}{blue!50!black}
\colorlet{dark-cyan}{cyan!75!black}
\colorlet{dark-purple}{purple!50!black}
\colorlet{dark-red}{red!75!black}
\colorlet{dark-green}{green!75!black}
\colorlet{dark-orange}{orange!50!black}
\colorlet{dark-gray}{black!75}
\colorlet{light-gray}{black!30}
\colorlet{hidden}{light-gray}
\colorlet{todo}{red!85!black}
\colorlet{todoref}{purple!70!black}
\definecolor{ucl-purple}{RGB}{80,7,120}
\definecolor{ucl-navy-blue}{RGB}{0,40,85}
\definecolor{ucl-mid-green}{RGB}{143,153,62}
\definecolor{ucl-burgundy}{RGB}{147,39,44}
\definecolor{ucl-mid-red}{RGB}{224,60,49}
\definecolor{ucl-orange}{RGB}{234,118,0}
\definecolor{ucl-yellow}{RGB}{246,190,0}
\colorlet{ucl-blue}{ucl-navy-blue}
\colorlet{ucl-green}{ucl-mid-green}
\colorlet{ucl-red}{ucl-burgundy}

\definecolor{nice-red}{HTML}{E41A1C}
\definecolor{nice-orange}{HTML}{FF7F00}
\definecolor{nice-yellow}{HTML}{FFC020}
\definecolor{nice-green}{HTML}{4DAF4A}
\definecolor{nice-blue}{HTML}{377EB8}
\definecolor{nice-purple}{HTML}{984EA3}

\colorlet{niced-red}{nice-red!75!black}
\colorlet{niced-orange}{nice-orange!75!black}
\colorlet{niced-yellow}{nice-yellow!75!black}
\colorlet{niced-green}{nice-green!75!black}
\colorlet{niced-blue}{nice-blue!75!black}
\colorlet{niced-purple}{nice-purple!75!black}

\usetikzlibrary{matrix, calc, arrows, decorations.markings}

\makeatletter
\tikzset{%
  prefix node name/.code={%
    \tikzset{%
      name/.code={\edef\tikz@fig@name{#1 ##1}}
    }%
  }%
}
\makeatother

\newcommand{\at}[3]{
  \begin{scope}[shift={(#1,#2)}]
    #3
  \end{scope} 
}

\newcommand{\name}[2]{
  \begin{scope}[local bounding box=#1]
    #2
  \end{scope}
}

\tikzset{ shorten <>/.style={ shorten >=#1, shorten <=#1 } }

\tikzstyle{arrow}=[draw, -latex, thick] 
\tikzstyle{axis}=[draw, -stealth, thick]
\tikzstyle{vector}=[arrow, very thick]
\tikzstyle{mapping}=[draw, -open triangle 45, very thick, color=hidden]

\makeatletter
\newcommand\getwidthofnode[2]{%
  \pgfextractx{#1}{\pgfpointanchor{#2}{east}}%
  \pgfextractx{\pgf@xa}{\pgfpointanchor{#2}{west}}
  \addtolength{#1}{-\pgf@xa}%
}

\tikzset{onslide/.code args={<#1>#2}{%
  \only<#1>{\pgfkeysalso{#2}} 
}}

\tikzset{temporal/.code args={<#1>#2#3#4}{%
  \temporal<#1>{\pgfkeysalso{#2}}{\pgfkeysalso{#3}}{\pgfkeysalso{#4}} 
}}

\tikzset{%
  database/.style={
    cylinder,
    cylinder uses custom fill,
    cylinder body fill=black!10,
    cylinder end fill=black!10,
    shape border rotate=90,
    aspect=0.25,
    thick,
    draw
  }
}

\usepackage{tikz}
\usetikzlibrary{calc,trees,positioning,arrows,chains,shapes.geometric,%
  decorations.pathreplacing,decorations.pathmorphing,shapes,%
  matrix,shapes.symbols,fit,decorations}

\usepackage{xcolor}
\usepackage{xparse}

\newcommand{\tikztensorx}{1}
\newcommand{\tikztensory}{1}
\newcommand{\tikztensorz}{1}

\newcommand{\setcoordinates}[3]{
  \renewcommand{\tikztensorx}{#1}
  \renewcommand{\tikztensory}{#2}
  \renewcommand{\tikztensorz}{#3}

  \coordinate (0) at (0,0,0);
  \coordinate (x) at (\tikztensorx,0,0);
  \coordinate (y) at (0,\tikztensory,0);
  \coordinate (z) at (0,0,-\tikztensorz);
  \coordinate (xy) at (\tikztensorx,\tikztensory,0);
  \coordinate (xz) at (\tikztensorx,0,-\tikztensorz);
  \coordinate (yz) at (0,\tikztensory,-\tikztensorz);
  \coordinate (xyz) at (\tikztensorx,\tikztensory,-\tikztensorz); 
}

\DeclareDocumentCommand{\tensorfront}{O{1} m}{
  \draw[black, fill = black, fill opacity=0.15, #2] (x) -- (xz) -- (xyz) -- (xy) -- cycle;
  \draw[black, fill = black, fill opacity=0.1, #2] (y) -- (xy) -- (xyz) -- (yz) -- cycle;
  \ifnum #1=1
  \draw[black, fill = black, fill opacity=0.05, #2] (0) -- (x) -- (xy) -- (y) -- cycle;
  \fi
}

\DeclareDocumentCommand{\tensor}{O{} O{} O{1} O{1} O{1} O{1} m m m}{
  \name{lhs}{

    #2
    
    \setcoordinates{#7 * 0.25}{#8 * 0.25}{#9 * 0.25}
    \tensorfront[#6]{#1}
    \tensorgrid[#3][#4][#5]{#7}{#8}{#9}
  }
}

\DeclareDocumentCommand{\tensorgrid}{O{1} O{1} O{1} m m m}{
    \ifnum #1=1
    \foreach \i in {0,...,#4} {
      \draw[color=white] (\i*0.25,0) -- (\i*0.25,#5*0.25);
      \draw[color=white] (\i*0.25,#5*0.25) -- (\i*0.25,#5*0.25,-#6*0.25); 
    }
    \fi
    
    \ifnum #2=1
    \foreach \i in {0,...,#5} {
      \draw[color=white] (0,\i*0.25) -- (#4*0.25,\i*0.25);
      \draw[color=white] (#4*0.25,\i*0.25) -- (#4*0.25,\i*0.25,-#6*0.25); 
    }
    \fi

    \ifnum #3=1
    \foreach \i in {0,...,#6} {
      \draw[color=white] (0,#5*0.25,-\i*0.25) -- (#4*0.25,#5*0.25,-\i*0.25);
      \draw[color=white] (#4*0.25,0,-\i*0.25) -- (#4*0.25,#5*0.25,-\i*0.25);
    }
    \fi
}

\DeclareDocumentCommand{\tensorop}{O{.5em} O{1.5em} m}{
  \node[right = #1 of lhs.south east, yshift=#2] (op) {#3};
  \node[right = #1 of 0 -| op] {};
  \pgfgetlastxy{\nextx}{\nexty};
}

\DeclareDocumentCommand{\calculatenext}{O{.5em} O{2em} m}{
  \node[right = #1 of #3.south east, yshift=#2] (tmp) {};
  \pgfgetlastxy{\nextx}{\nexty};
}

\DeclareDocumentCommand{\setnext}{O{} m}{
  \node[#1] at (#2) {};
  \pgfgetlastxy{\nextx}{\nexty};
}

\DeclareDocumentCommand{\labelaxis}{O{black} m m m m}{
  \draw[ultra thick, #1] (#2) -- node[midway, #4] {#5} (#3);
}

\DeclareDocumentCommand{\labelx}{O{dark-green} m}{
  \labelaxis[#1]{0}{x}{below}{#2}
}

\DeclareDocumentCommand{\labely}{O{red} m}{
  \labelaxis[#1]{0}{y}{left}{#2}
}

\DeclareDocumentCommand{\labelz}{O{blue} m}{
  \labelaxis[#1]{y}{yz}{above left}{#2}
}

\DeclareDocumentCommand{\tensorshift}{O{0} O{0} O{0} m}{
  \begin{scope}[shift={(#1 * 0.25,#2 * 0.25,-#3 * 0.25)}]
    #4
  \end{scope} 
}

\newcommand{\eg}{\emph{e.g.}}
\newcommand{\ie}{\emph{i.e.}}

\newcommand{\kb}{\mathfrak{K}}
\newcommand{\state}{S}
\renewcommand{\emptyset}{\varnothing}
\newcommand{\fail}{\verb~FAIL~}
\newcommand{\fun}[1]{\text{#1}}

\DeclareRobustCommand{\por}{\textsc{Or}}

\let\union\cup
\let\dom\mathcal
\let\set\dom 

\newcommand{\lss}[1]{\mathbb{#1}}
\let\ls\textsc
\let\lst\ls
\newcommand{\datadom}[1]{#1}
\newcommand{\tensordom}[1]{#1}

\newcommand{\xs}[1]{\bm{[} #1 \bm{]}}
\newcommand{\emptylist}{\bm{[}\ \bm{]}}

\def\R{\mathbb{R}}
\def\N{\mathbb{N}}
\def\E{\mathbb{E}}
\def\C{\mathbb{C}}
\renewcommand\vec[1]{{\bm{#1}}}
\let\mat\bm
\let\ten\mathcal

\let\grad\nabla

\makeatletter
\newcommand*\bdot{\mathpalette\bdot@{.5}}
\newcommand*\bdot@[2]{\mathbin{\vcenter{\hbox{\scalebox{#2}{$\m@th#1\bullet$}}}}}
\makeatother

\def\loss{\mathcal{L}}
\def\params{{\vec{\theta}}}
\def\globalloss{\mathfrak{L}}

\newcommand{\module}[1]{\verb~#1~}
\def\ndot{\module{dot}}
\def\sigm{\module{sigm}}

\def\relu{\module{ReLU}}
\def\rnn{\module{RNN}}
\def\lstm{\module{LSTM}}

\DeclareMathOperator{\softmax}{softmax}
\DeclareMathOperator*{\argmax}{arg\,max}
\DeclareMathOperator*{\argmin}{arg\,min}
\let\sigmoid\sigma

\DeclareMathOperator{\real}{real}
\DeclareMathOperator{\imag}{imag}

\newcommand{\rel}[1]{\verb~#1~}
\newcommand{\const}[1]{\textsc{#1}}
\let\ent\const
\newcommand{\var}[1]{\textsc{#1}}
\def\lif{\ \text{:--}\ }
\newcommand{\ldiff}[1]{\left\llbracket #1 \right\rrbracket}
\let\lrule\textsc
\def\lneg{\neg\,}
\let\land\wedge
\let\lor\vee

\let\subs\Psi

\newcommand{\pred}[1]{\text{$\verb~#1~$}}

\let\success\tau

\newcommand{\rparam}[1]{\bm{\theta}_{#1}}

\tikzstyle{cnode} = [draw, circle, minimum size=1cm]
\tikzstyle{op} = [color=nice-red]
\tikzstyle{input} = [cnode, fill=black!20]
\tikzstyle{param} = [cnode, color=nice-blue, fill=nice-blue!20]
\tikzstyle{output} = [cnode, color=nice-red, fill=nice-red!20, dashed]
\tikzstyle{grad} = [color=nice-red!80!black]
\tikzstyle{op left} = [below left=0.4cm]
\tikzstyle{op right} = [below right=0.4cm]
\tikzstyle{op below} = [below=0.75cm]
\tikzstyle{op above} = [above=0.75cm]
\tikzstyle{op above left} = [above left=0.4cm]
\tikzstyle{op above right} = [above right=0.4cm]

\let\todonote\todo

\colorlet{fixme}{red!85!black}
\colorlet{fixme-bright}{fixme!25}

\colorlet{todo}{orange!85!black}
\colorlet{todo-bright}{todo!25}

\colorlet{review}{nice-yellow!50!nice-orange}
\colorlet{review-bright}{review!25}

\colorlet{maybe}{nice-yellow}
\colorlet{maybe-bright}{maybe!25}

\colorlet{toref}{purple!70!black}
\colorlet{toref-bright}{toref!25}

\colorlet{info}{green!50!black}
\colorlet{info-bright}{info!25}

\colorlet{cut}{black!60}
\colorlet{cut-bright}{cut!25}

\colorlet{extend}{blue!85!black}
\colorlet{extend-bright}{extend!25}

\colorlet{discuss}{nice-red!50!black}
\colorlet{discuss-bright}{discuss!25}

\renewcommand{\todo}[1]{\todonote[linecolor=todo, backgroundcolor=todo-bright, bordercolor=todo]{\thesection{} {\bf\color{todo}TODO} #1}{}}

\let\check\review

\newcommand{\info}[1]{}
\newcommand{\hlinfo}[2]{}

\newcommand{\maybe}[1]{}
\newcommand{\hlmaybe}[2]{}

\newacronym{AKBC}{AKBC}{Automated Knowledge Base Construction}
\newacronym{AUC}{AUC}{Area Under Curve}

\newacronym{BPR}{BPR}{Bayesian Personalized Ranking}

\newacronym{CNF}{CNF}{Conjunctive Normal Form}
\newacronym{CNN}{CNN}{Convolutional Neural Network}
\newacronym{CWA}{CWA}{Closed World Assumption}

\newacronym{EqNet}{EqNet}{Neural Equivalence Network}

\newacronym{FOIL}{FOIL}{First Order Inductive Learner}

\newacronym{GPCA}{GPCA}{Generalized Principal Component Analysis}
\newacronym{GPU}{GPU}{Graphics Processing Unit}

\newacronym{ILP}{ILP}{Inductive Logic Programming}

\newacronym{KB}{KB}{Knowledge Base}

\newacronym[longplural={Long Short-Term Memories}]{LSTM}{LSTM}{long short-term memory}

\newacronym{NLP}{NLP}{Natural Language Processing}
\newacronym{NLI}{NLI}{Natural Language Inference}
\newacronym{NTP}{NTP}{Neural Theorem Prover}
\newacronym{NTN}{NTN}{Neural Tensor Network}

\newacronym{MAP}{MAP}{Mean Average Precision}
\newacronym{MRR}{MRR}{Mean Reciprocal Rank}
\newacronym{MLP}{MLP}{Multi-layer Perceptron}

\newacronym{OpenIE}{OpenIE}{Open Information Extraction}

\newacronym{PCA}{PCA}{Principal Component Analysis}
\newacronym{PRA}{PRA}{Path Ranking Algorithm}
\newacronym{ProPPR}{ProPPR}{Programming with Personalized PageRank}

\newacronym{RBF}{RBF}{Radial Basis Function}
\newacronym{ROC}{ROC}{Receiver Operating Characteristic}
\newacronym{RNN}{RNN}{Recurrent Neural Network}
\newacronym{RTE}{RTE}{Recognizing Textual Entailment}

\newacronym{SGD}{SGD}{Stochastic Gradient Descent}
\newacronym{SLD}{SLD}{Selective Linear Definite clause resolution}
\newacronym{SNLI}{SNLI}{Stanford Natural Language Inference}

\usepackage[utf8]{inputenc}

\makeglossaries

\title{Combining Representation Learning\\[1ex] with Logic for Language Processing} 
\author{Tim Rocktäschel}
\department{Department of Computer Science}

\begin{document}
\nobibliography*

\pagenumbering{arabic}

\maketitle

\chapter*{}
\begin{center}
\emph{To Paula, Emily, Sabine, and Lutz.}
\end{center}

\chapter*{Acknowledgements}      
I am deeply grateful to my supervisor and mentor Sebastian Riedel.
He always has been a great source of inspiration and supportive and encouraging in all matters. 
I am particularly amazed by the trust he put into me from the first time we met, the freedom he gave me to pursue ambitious ideas, his contagious optimistic attitude, and the many opportunities he presented to me.
There is absolutely no way I could have wished for a better supervisor.

I would like to thank my second supervisors Thore Graepel and Daniel Tarlow for their feedback, as well as Sameer Singh whose collaboration and guidance made my start into the Ph.D. very smooth, motivating, and fun. 
Furthermore, I thank Edward Grefenstette, Karl Moritz Hermann, Thomas Kociský and Phil Blunsom, who I was fortunate to work with during my internship at DeepMind in Summer 2015.
I am thankful for the guidance by Thomas Demeester, Andreas Vlachos, Pasquale Minervini, Pontus Stenetorp, Isabelle Augenstein, and Jason Naradowsky during their time at the University College London Machine Reading lab.
In addition, thanks to Dirk Weissenborn for many fruitful discussions.
I would also like to thank Ulf Leser, Philippe Thomas, and Roman Klinger for preparing me well for the Ph.D. during my studies at Humboldt-Universität zu Berlin.

I had the pleasure to work with brilliant students at University College London. Thanks to Michal Daniluk, Luke Hewitt, Ben Eisner, Vladyslav Kolesnyk, Avishkar Bhoopchand, and Nayen Pankhania for their hard work and trust.
Ph.D. life would not have been much fun without my lab mates Matko Bosnjak, George Spithourakis, Johannes Welbl, Marzieh Saeidi, and Ivan Sanchez.
Thanks to Matko, Dirk, Pasquale, Thomas, Johannes, and Sebastian for feedback on this thesis.
Furthermore, I thank my examiners Luke Dickens and Charles Sutton for their extremely helpful in-depth comments and corrections of this thesis.
Thank you, Federation Coffee in Brixton, for making the best coffee in the world.

Many thanks to Microsoft Research for supporting my work through its Ph.D. Scholarship Programme, and to Google Research for a Ph.D. Fellowship in Natural Language Processing.
Thanks to their generous support, as well as the funding from the University College London Computer Science Department, I was able to travel to all conferences that I wanted to attend.

I am greatly indebted and thankful to my parents Sabine and Lutz. 
They sparked my interest in science, but more importantly, they ensured that I had a fantastic childhood.
I always felt loved, supported, and well protected from the many troubles in life.
Furthermore, I would like to thank Christa, Ulrike, Tillmann, Hella, Hans, Gretel, and Walter for their unconditional support over the last years.

Lastly, thanks to the two wonderful women in my life, Paula and Emily.
Thank you, Paula, for keeping up with my ups and downs during the Ph.D., your love, motivation, and support.
Thank you for giving me a family, and for making us feel at home wherever we are.
Emily, you are the greatest wonder and joy in my life.

\chapter*{Declaration}

I, Tim Rocktäschel confirm that the work presented in this thesis is my own. Where information has been derived from other sources, I confirm that this has been indicated in the thesis.

\vspace{2cm}
\hfill\textsc{Tim Rocktäschel}

\clearpage

\begin{abstract}

The current state-of-the-art in many natural language processing and automated knowledge base completion tasks is held by representation learning methods which learn distributed vector representations of symbols via gradient-based optimization. They require little or no hand-crafted features, thus avoiding the need for most preprocessing steps and task-specific assumptions. However, in many cases representation learning requires a large amount of annotated training data to generalize well to unseen data. Such labeled training data is provided by human annotators who often use formal logic as the language for specifying annotations.

This thesis investigates different combinations of representation learning methods with logic for reducing the need for annotated training data, and for improving generalization. We introduce a mapping of function-free first-order logic rules to loss functions that we combine with neural link prediction models. Using this method, logical prior knowledge is directly embedded in vector representations of predicates and constants. We find that this method learns accurate predicate representations for which no or little training data is available, while at the same time generalizing to other predicates not explicitly stated in rules. However, this method relies on grounding first-order logic rules, which does not scale to large rule sets. To overcome this limitation, we propose a scalable method for embedding implications in a vector space by only regularizing predicate representations. Subsequently, we explore a tighter integration of representation learning and logical deduction. We introduce an end-to-end differentiable prover -- a neural network that is recursively constructed from Prolog's backward chaining algorithm. The constructed network allows us to calculate the gradient of proofs with respect to symbol representations and to learn these representations from proving facts in a knowledge base. In addition to incorporating complex first-order rules, it induces interpretable logic programs via gradient descent. Lastly, we propose recurrent neural networks with conditional encoding and a neural attention mechanism for determining the logical relationship between two natural language sentences.

\end{abstract}

\clearpage

\chapter*{Impact Statement}
Machine learning, and representation learning in particular, is ubiquitous in many applications nowadays. 
Representation learning requires little or no hand-crafted features, thus avoiding the need for task-specific assumptions.
At the same time, it requires a large amount of annotated training data.
Many important domains lack such large training sets, for instance, because annotation is too costly or domain expert knowledge is generally hard to obtain.

The combination of neural and symbolic approaches investigated in this thesis has only recently regained significant attention due to advances of representation learning research in certain domains and, more importantly, their lack of success in other domains.
The research conducted under this Ph.D. investigated ways of training representation learning models using explanations in form of function-free first-order logic rules in addition to individual training facts. 
This opens up the possibility of taking advantage of the strong generalization of representation learning models, while still being able to express domain expert knowledge.
We hope that this work will be particularly useful for applying representation learning in domains where annotated training data is scarce, and that it will empower domain experts to train machine learning models by providing explanations.

\clearpage

\setcounter{tocdepth}{2} 

{
\tableofcontents
\listoffigures
\listoftables
}

\chapter{Introduction}
\glsresetall

``\emph{We are attempting to replace symbols by vectors so we can replace logic by algebra.}''
\begin{flushright}
--- Yann LeCun
\end{flushright}

The vast majority of knowledge produced by mankind is nowadays available in a digital but unstructured form such as images or text. 
It is hard for algorithms to extract meaningful information from such data resources, let alone to reason with it. 
This issue is becoming more severe as the amount of unstructured data is growing very rapidly.

In recent years, remarkable successes in processing unstructured data have been achieved by representation learning methods which automatically learn abstractions from large collections of training data. 
This is achieved by processing input data using artificial neural networks whose weights are adapted during training.
Representation learning lead to breakthroughs in applications such as automated \gls{KB} completion \citep{nickel2012factorizing,riedel2013relation,socher2013reasoning,chang2014typed,yang2014embedding,neelakantan2015compositional,toutanova2015representing,trouillon2016complex}, as well as \gls{NLP} applications like paraphrase detection \citep{socher2011dynamic, hu2014convolutional, yin2015convolutional}, machine translation \citep{bahdanau2014neural}, image captioning \citep{xu2015show}, speech recognition \citep{chorowski2015attention} and sentence summarization \citep{rush2015neural}, to name just a few.

Representation learning methods achieve remarkable results, but they usually rely on a large amount of annotated training data.
Moreover, since representation learning operates on a subsymbolic level (for instance by replacing words with lists of real numbers -- so-called \emph{vector representations} or \emph{embeddings}), it is hard to determine why we obtain a certain prediction, let alone how to correct systematic errors or how to incorporate domain and commonsense knowledge.
In fact, a recent General Data Protection Regulation by the European Union introduces the “right to explanation” of decisions by algorithms and machine learning models that affect users \citep{council2016position}, to be enacted in 2018. 
This has profound implications for the future development and research of machine learning algorithms \citep{goodman2016eu}, especially for nowadays commonly used representation learning methods. 
Moreover, for many domains of interests there is not enough annotated training data, which renders applying recent representation learning methods difficult.

Many of these issues do not exist with purely symbolic approaches.
For instance, given a \gls{KB} of facts and first-order logic rules, we can use Prolog to obtain an answer as well as a proof for a query to this \gls{KB}.
Furthermore, we can easily incorporate domain knowledge by adding more rules.
However, rule-based system do not generalize to new questions.
For instance, given that an apple is a fruit and apples are similar to oranges, we would like to infer that oranges are likely also fruits.

To summarize, symbolic rule-based systems are interpretable and easy to modify. They do not need large amounts of training data and we can easily incorporate domain knowledge. On the other hand, learning subsymbolic representations requires a lot of training data. The trained models are generally opaque and it is hard to incorporate domain knowledge. Consequently, we would like to develop methods that take the best of both worlds.

\section{Aims}
In this thesis, we are investigating the combination of representation learning with first-order logic rules and reasoning.  
Representation learning methods achieve strong generalization by learning subsymbolic vector representations that can capture similarity and even logical relationships directly in a vector space \citep{mikolov2013distributed}. 
Symbolic representations, on the other hand, allow us to formalize domain and commonsense knowledge using rules. 
For instance, we can state that every \rel{human} is \rel{mortal}, or that every \rel{grandfather} is a \rel{father} of a \rel{parent}.
Such rules are often worth many training facts.
Furthermore, by using symbolic representations we can take advantage of algorithms for multi-hop reasoning like Prolog's backward chaining algorithm \citep{colmerauer1990introduction}. 
Backward chaining is not only widely used for multi-hop question answering in \glspl{KB}, but it also provides us with proofs in addition to the answer for a question.
However, such symbolic reasoning is relying on a complete specification of background and commonsense knowledge in logical form.
As an example, let us assume we are asking for the \rel{grandpa} of a person, but only know the \rel{grandfather} of that person.
If there is no explicit rule connecting \rel{grandpa} to \rel{grandfather}, we will not find an answer.
However, given a large \gls{KB}, we can use representation learning to learn that \rel{grandpa} and \rel{grandfather} mean the same thing, or that a \rel{lecturer} is similar to a \rel{professor} \citep{nickel2012factorizing,riedel2013relation,socher2013reasoning,chang2014typed,yang2014embedding,neelakantan2015compositional,toutanova2015representing,trouillon2016complex}.
This becomes more relevant once we do not only want to reason with structured relations but also use textual patterns as relations \citep{riedel2013relation}.

The problem that we seek to address in this thesis is how symbolic logical knowledge can be combined with representation learning to make use of the best of both worlds.
Specifically, we investigate the following research questions.
\begin{itemize}
\item Can we efficiently incorporate domain and commonsense knowledge in form of rules into representation learning methods? 
\item Can we use rules to alleviate the need for large amounts of training data while still generalizing beyond what is explicitly stated in these rules?
\item Can we synthesize representation learning with symbolic multi-hop reasoning as used for automated theorem proving?
\item Can we learn rules directly from data using representation learning?
\item Can we determine the logical relationship between natural language sentences using representation learning?
\end{itemize}

\section{Contributions}

This thesis makes the following core contributions.

\paragraph{Regularizing Representations by First-order Logic Rules}
We introduce a method for incorporating function-free first-order logic rules directly into vector representations of symbols, which avoids the need for symbolic inference. 
Instead of symbolic inference, we regularize symbol representations by given rules such that logical relationships hold implicitly in the vector space (\cref{log}).
This is achieved by mapping propositional logical rules to differentiable loss terms so that we can calculate the gradient of a given rule with respect to symbol representations.
Given a first-order logic rule, we stochastically ground free variables using constants in the domain, and add the resulting loss term for the propositional rule to the training objective of a neural link prediction model for automated \gls{KB} completion.
This allows us to infer relations with little or no training facts in a \gls{KB}.
While mapping logical rules to soft rules using algebraic operations has a long tradition (\eg{} in Fuzzy logic), our contribution is the connection to representation learning, \ie{}, using such rules to directly learning better vector representations of symbols that can be used to improve performance on a downstream task such as automated \gls{KB} completion.
Content in this chapter first appeared in the following two publications:
\vspace{0.5ex}
\begin{adjustwidth}{2em}{}
  Tim Rocktäschel, Matko Bosnjak, Sameer Singh and Sebastian Riedel. 
  2014.  
  Low-Dimensional Embeddings of Logic.
  In {\em Proceedings of Association for Computational Linguistics Workshop on Semantic Parsing (SP'14)}.
  \nocite{rocktaschel2015injecting}
\end{adjustwidth}
\vspace{1ex}
\begin{adjustwidth}{2em}{}
  Tim Rocktäschel, Sameer Singh and Sebastian Riedel. 
  2015.  
  Injecting Logical Background Knowledge into Embeddings for Relation Extraction. 
  In {\em Proceedings of North American Chapter of the Association for Computational Linguistics -- Human Language Technologies (NAACL HLT 2015)}.
  \nocite{rocktaschel2015injecting}
\end{adjustwidth}
\vspace{1ex}

\paragraph{Lifted Regularization of Predicate Representations by Implications}
For the subclass of first-order logic implication rules, we present a scalable method that is independent of the size of the domain of constants, that generalizes to unseen constants, and that can be used with a broader class of training objectives (\cref{foil}).
Instead of relying on stochastic grounding, we use implication rules directly as regularizers for predicate representations.
Compared to the method in \cref{log}, this method is independent of the number of constants and ensures that a given implication between two predicates holds for any possible pair of constants at test time.
Our method is based on Order Embeddings \citep{vendrov2016order} and our contribution is the extension to the task of automated \gls{KB} completion which requires constraining entity representations to be non-negative.
This chapter is based on the following two publications:
\vspace{0.5ex}
\begin{adjustwidth}{2em}{}
  Thomas Demeester, Tim Rocktäschel and Sebastian Riedel. 
  2016.
  Regularizing Relation Representations by First-order Implications.
  In {\em Proceedings of North American Chapter of the Association for Computational Linguistics (NAACL) Workshop on Automated Knowledge Base Construction (AKBC)}.
  \nocite{demeester2016lifted}
\end{adjustwidth}
\vspace{1ex}
\begin{adjustwidth}{2em}{}
  Thomas Demeester, Tim Rocktäschel and Sebastian Riedel. 
  2016.
  Lifted Rule Injection for Relation Embeddings.
  In {\em Proceedings of Empirical Methods in Natural Language Processing (EMNLP)}.
  \nocite{demeester2016lifted}
\end{adjustwidth}
My contribution to this work is the conceptualization of the model, the design of experiments, and the extraction of commonsense rules from WordNet. 

\paragraph{End-to-end Differentiable Proving} 
Current representation learning and neural link prediction models have deficits when it comes to complex multi-hop inferences such as transitive reasoning \citep{bouchard2015approximate,nickel2015holographic}.
Automated theorem provers, on the other hand, have a long tradition in computer science and provide elegant ways to reason with symbolic knowledge. 
In \cref{ntp}, we propose \glspl{NTP}: end-to-end differentiable theorem provers for automated \gls{KB} completion based on neural networks that are recursively constructed and inspired by Prolog's backward chaining algorithm.
By doing so, we can calculate the gradient of a proof success with respect to symbol representations in a \gls{KB}.
This allows us to learn symbol representations directly from facts in a \gls{KB}, and to make use of the similarities of symbol representations and provided rules in proofs.
In addition, we demonstrate that we can induce interpretable rules of predefined structure.
On three out of four benchmark \glspl{KB}, our method outperforms ComplEx \citep{trouillon2016complex}, a state-of-the-art neural link prediction model.
Work in this chapter appeared in:
\vspace{0.5ex}
\begin{adjustwidth}{2em}{}
  Tim Rocktäschel and Sebastian Riedel. 
  2016.  
  Learning Knowledge Base Inference with Neural Theorem Provers.
  In {\em Proceedings of North American Chapter of the Association for Computational Linguistics (NAACL) Workshop on Automated Knowledge Base Construction (AKBC)}.
  \nocite{rocktaschel2016learning}
\end{adjustwidth}
\vspace{1ex}
\begin{adjustwidth}{2em}{}
  Tim Rocktäschel and Sebastian Riedel. 
  2017.  
  End-to-End Differentiable Proving.
  In {\em Advances in Neural Information Processing Systems 31: Annual Conference on Neural Information Processing Systems (NIPS)}.
  \nocite{rocktaschel2017end}
\end{adjustwidth}

\paragraph{Recognizing Textual Entailment with Recurrent Neural Networks}
Representation learning models such as \glspl{RNN} can be used to map natural language sentences to fixed-length vector representations, which has been successfully applied for various downstream \gls{NLP} tasks including \gls{RTE}.
In \gls{RTE}, the task is to determine the logical relationship between two natural language sentences.
This has so far been either approached by \gls{NLP} pipelines with hand-crafted features, or neural network architectures that independently map the two sentences  to fixed-length vector representations.
Instead of encoding the two sentences independently, we propose a model that encodes the second sentence conditioned on an encoding of the first sentence.
Furthermore, we apply a neural attention mechanism to bridge the hidden state bottleneck of the \gls{RNN} (\cref{rte}).
Work in this chapter first appeared in:
\vspace{0.5ex}
\begin{adjustwidth}{2em}{}
  Tim Rocktäschel, Edward Grefenstette, Karl Moritz Hermann, Tomas Kocisky and Phil Blunsom.
  2016.
  Reasoning about Entailment with Neural Attention.
  In {\em Proceedings of International Conference on Learning Representations (ICLR)}.
  \nocite{rocktaschel2016reasoning}
\end{adjustwidth}
\vspace{1ex}

\section{Thesis Structure}
In \cref{back}, we provide background on representation learning, computation graphs, first-order logic, and the notation used throughout the thesis.
Furthermore, we explain the task of automated \gls{KB} completion and describe neural link prediction and path-based approaches that have been proposed for this task.
\Cref{log} introduces a method for regularizing symbol representations by first-order logic rules.
In \cref{foil}, we subsequently focus on direct implications between predicates, a subset of first-order logic rules.
For this class of rules, we provide an efficient method by directly regularizing predicate representations.
In \cref{ntp}, we introduce a recursive construction of a neural network for automated \gls{KB} completion based on Prolog's backward chaining algorithm.
\cref{rte} presents a \gls{RNN} for \gls{RTE} based on conditional encoding and a neural attention mechanism. 
Finally, \cref{conclusions} concludes the thesis with a discussion of limitations, open issues, and future research avenues.

\chapter{Background}
\label{back}
\glsresetall

This chapter introduces core methods used in the thesis. 
\Cref{sec:function_approximation} explains function approximation with neural networks and backpropagation.
Subsequently, \cref{sec:fol} introduces function-free first-order logic, the backward chaining algorithm, and inductive logic programming.
Finally, \cref{sec:akbc} discusses prior work on automated knowledge base completion, linking the first two sections together.

\section{Function Approximation with Neural Networks}
\label{sec:function_approximation}
In this thesis, we consider models that can be formulated as differentiable functions~$f_\params:~\dom{X}~\to~\dom{Y}$ parameterized by~$\params\in\Theta$. 
Our task is to find such functions, \ie{}, to learn parameters $\params$ from a set of training examples $\set{T}=\{(x_i,y_i)\}$ where $x_i\in\dom{X}$ is the input and $y_i\in\dom{Y}$ some desired output of the $i$th training example.
Both, $x_i$ and $y_i$, can be structured objects.
For instance, $x_i$ could be a fact about the world, like $\rel{directedBy}(\ent{Interstellar}, \ent{Nolan})$, and $y_i$ a corresponding target truth score (\eg{} $1.0$ for \textsc{True}).

We define a loss function $\loss:~\dom{Y}\times\dom{Y}\times\Theta \to \R$ that measures the discrepancy between a provided output $y$ and a predicted output $f_{\params}(x)$ on an input $x$, given a current setting of parameters $\params$.
We seek to find those parameters $\params^*$ that minimize this discrepancy on a training set.
We denote the global loss over the entire training data as $\globalloss$.
Our learning problem can thus be written as
\begin{equation}
  \label{eq:1}
  \params^* = \argmin_\params \globalloss = \argmin_\params \frac{1}{|\set{T}|}\sum_{(x,y)\,\in\,\set{T}}\loss(f_{\params}(x), y, \params).
\end{equation}
Note that $\loss$ is also a function of $\params$, since we might not only want to measure the discrepancy between given and predicted outputs, but also use a regularizer on the parameters to improve generalization.
Sometimes, we omit $\params$ in $\loss$ for brevity.\maybe{Seb: usually regularizer is not defined on a per-instance basis}
As $\loss$ and $f_{\params}$ are differentiable functions, we can use gradient-based optimization methods, such as \gls{SGD} \citep{nemirovski1978cezari}, for iteratively updating $\params$ based on mini-batches $\set{B}\subseteq\set{T}$ of the training data\footnote{Note that there are many alternative methods for minimizing the loss in \cref{eq:1}, but all models in this thesis are optimized by variants of \gls{SGD}.}
\begin{equation}
  \label{eq:2}
  \params_{t+1} = \params_t - \eta\,\grad_{\params_t}\,\frac{1}{|\set{B}|}\sum_{(x,y)\,\in\,\set{B}}\loss(f_{\params}(x), y, \params_t)
\end{equation}
where $\eta$ denotes a learning rate, and $\grad_\params$ denotes the differentiation operation of the loss with respect to parameters, given the current batch at time step $t$.
\maybe{more generally: minimize expected loss over full data distribution $\E_{(x,y)\sim p_\text{data}}$}

\subsection{Computation Graphs}
\label{sec:computation_graphs}
A useful abstraction for defining models as differentiable functions are \emph{computation graphs} which illustrate the computations carried out by a model more precisely \citep{goodfellow2016deep}.
In such a directed acyclic graph, nodes represent variables and directed edges from one or multiple nodes to another node correspond to a differentiable operation.
As variables we consider scalars, vectors, matrices, and, more generally, higher-order tensors.\footnote{For implementation purpose we will also consider structured objects over such tensors, like tuples and lists. Current deep learning libraries such as Theano \citep{al2016theano}, Torch \citep{collobert2011torch7} and TensorFlow \citep{abadi2015tensorflow} come with support for tuples and lists. However, for brevity we leave them out of the description here.}
We denote scalars by lower case letters $x$, vectors by bold lower case letters $\vec{x}$, matrices by bold capital letters $\vec{X}$, and higher-order tensors by Euler script letters $\ten{X}$.
Variables can either be inputs, outputs, or parameters of a model.

\begin{figure}[t!]
  \centering
  \begin{subfigure}[t]{0.5\textwidth}
    \centering
    \includegraphics{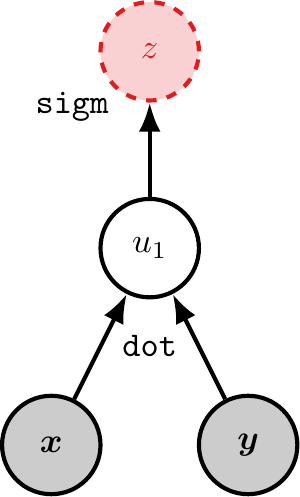}
    \caption{$z=\sigma(\vec{x}^\top\vec{y})$}
    \label{fig:sigm_xy}
  \end{subfigure}\hfill
  \begin{subfigure}[t]{0.5\textwidth}
    \centering
    \includegraphics{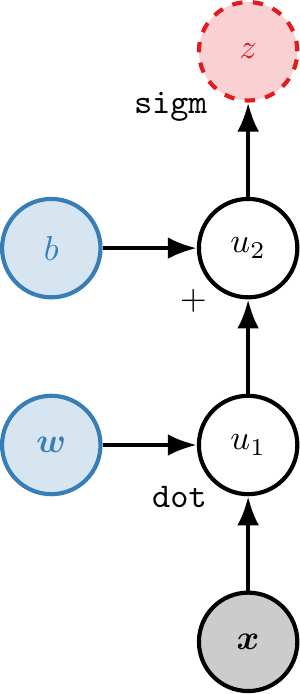}
    \caption{$z=\sigma(\vec{w}^\top\vec{x}+b)$}
    \label{fig:log_regression}
  \end{subfigure}

  \caption{Two simple computation graphs with inputs (gray), parameters (blue), intermediate variables $u_i$, and outputs (dashed). Operations are shown next to the nodes.}
  \label{fig:computation_graphs}
\end{figure}
\Cref{fig:sigm_xy} shows a simple computation graph that calculates $z=\sigma(\vec{x}^\top\vec{y})$.
Here, $\sigma$ and $\sigm$ refer to the element-wise (or scalar) sigmoid operation
\begin{equation}
\sigma(\vec{x}) = \frac{1}{1+e^{-\vec{x}}}
\end{equation} 
and $\ndot$ and $\vec{x}^\top\vec{y}$ denote the dot product between two vectors.
Furthermore, we name the $i$th intermediate expression as $u_i$.
\Cref{fig:log_regression} shows a slightly more complex computation graph with two parameters $\vec{w}$ and $b$. This computation graph in fact represents logistic regression $f(\vec{x}) = \sigma(\vec{w}^\top\vec{x} + b)$.

\subsection{From Symbols to Subsymbolic Representations}
\label{sec:sym2vec}

\begin{figure}[t!]
  \centering
  \includegraphics{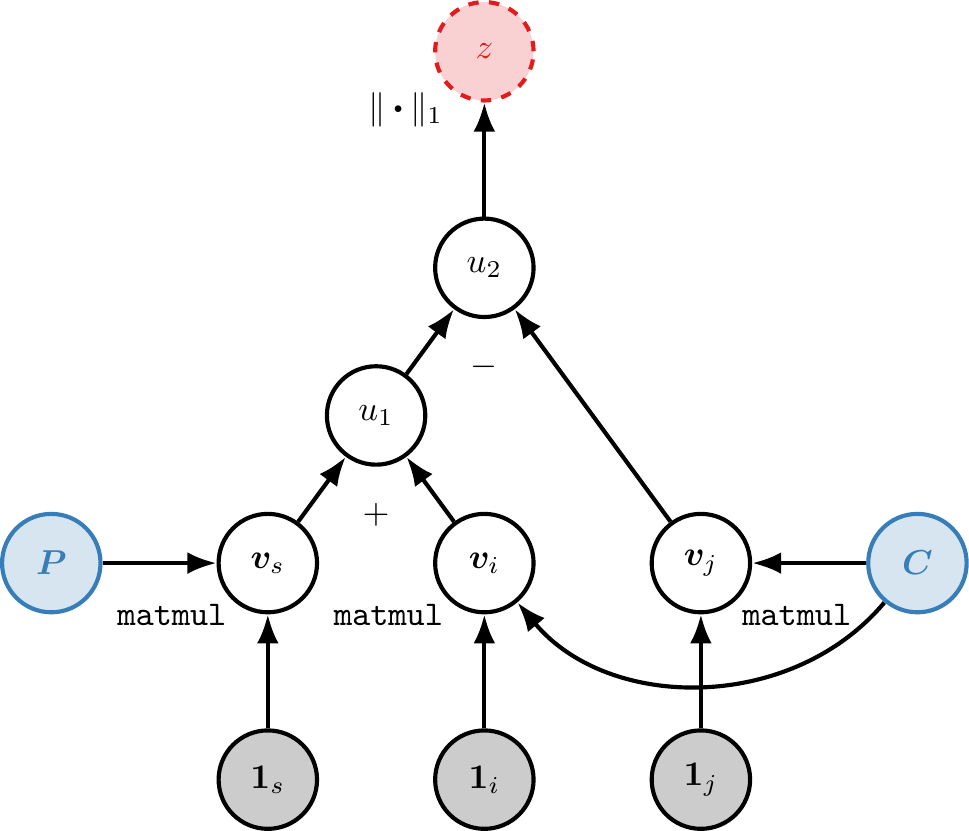}
  \caption{Computation graph for $z=\|\mat{C}\vec{1}_i + \mat{P}\vec{1}_s - \mat{C}\vec{1}_j\|_1$.}
  \label{fig:transe}
\end{figure}

In this thesis, we will use neural networks to learn representations of symbols.
For instance, such symbols can be words, constants, or predicate names.
When we say we learn a subsymbolic representation for symbols, we mean that we map symbols to fixed-dimensional vectors or more generally tensors.
This can be done by first enumerating all symbols and then assigning the number $i$ to the $i$th symbol.
Let $\set{S}$ be the set of all symbols.
We denote the one-hot vector for the $i$th symbol as $\vec{1}_i \in\{0,1\}^{|\set{S}|}$, which is $1$ at index $i$ and $0$ everywhere else.
\Cref{fig:transe} shows a computation graph whose inputs are one-hot vectors of some symbols with indices $s$, $i$, and $j$.
In the first layer, these one-hot vectors are mapped to dense vector representations via a matrix multiplication with so-called embedding lookup matrices ($\mat{P}$ and $\mat{C}$ in this case).

This computation graph corresponds to a neural link prediction model that we explain in more detail in \cref{sec:score}.
Here it serves only as an illustration for how symbols can be mapped to vector representations. 
In the remainder of this thesis, we will often omit the embedding layer for clarity.
The goal is to learn symbol representations such as $\vec{v}_s$, $\vec{v}_i$ and $\vec{v}_j$ automatically from data.
To this end, we need to be able to calculate the gradient of the output of the computation graph with respect to its parameters (the embedding lookup matrices in this case).

\subsection{Backpropagation}
For learning from data, we need to be able to calculate the gradient of a loss with respect to all model parameters.
As we assume all operations in the computation graph are differentiable, we can recursively apply the chain rule of calculus.

\paragraph{Chain Rule of Calculus}
Assume we are given a composite function $\vec{z} = f(\vec{y}) = f(g(\vec{x}))$ with $f: \R^n \to \R^m$ and $g: \R^l \to \R^n$.
The chain rule allows us to decompose the calculation of $\grad_\vec{x} \vec{z}$, \ie{}, the gradient of the entire computation $\vec{z}$ with respect to $\vec{x}$, as follows \citep{goodfellow2016deep}
\begin{equation}
  \label{eq:chain}
  \grad_\vec{x} \vec{z}  = \left(\frac{\partial\vec{y}}{\partial\vec{x}}\right)^\top \grad_\vec{y}\vec{z}.
\end{equation}
Here, $\frac{\partial\vec{y}}{\partial\vec{x}}$ is the Jacobian matrix of $g$, \ie{}, the matrix of partial derivatives, and $\grad_\vec{y}\vec{z}$ is the gradient of $\vec{z}$ with respect to $\vec{y}$.
Note that this approach generalizes to matrices and higher-order tensors by reshaping them to vectors before the gradient calculation (vectorization) and back to their original shape afterwards.

Backpropagation uses the chain rule to recursively define the efficient calculation of gradients of parameters (and inputs) in the computation graph by avoiding recalculation of previously calculated expressions. 
This is achieved via dynamic programming, \ie{}, storing previously calculated gradient expressions and reusing them for later gradient calculations.
We refer the reader to \cite{goodfellow2016deep} for details.
In order to run backpropagation with a differentiable operation $f$ that we want to use in a computation graph, all we need to ensure is that this function is differentiable with respect to each one of its inputs.

\paragraph{Example}
Let us take the computation graph depicted in \cref{fig:sigm_xy} as an example. 
Assume we are given an upstream gradient $\grad_z$ and want to compute the gradient of $z$ with respect to the inputs $\vec{x}$ and $\vec{y}$. 
\begin{figure}[t!]
  \centering
  \includegraphics{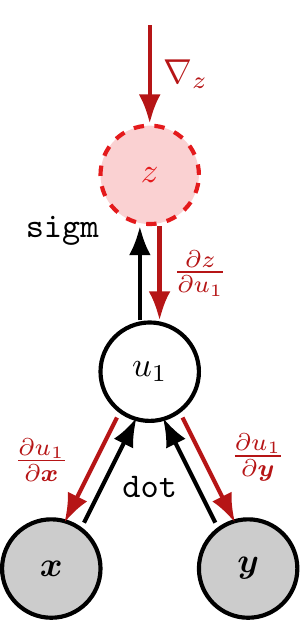}
  \caption{Backward pass for the computation graph shown in \cref{fig:sigm_xy}.}
  \label{fig:sigm_xy_backprop}
\end{figure}
The computations carried out by backpropagation are depicted in \cref{fig:sigm_xy_backprop}.
For instance, by recursively applying the chain rule we can calculate $\grad_\vec{x} z$ as follows:
\begin{equation}
  \label{eq:3}
  \grad_\vec{x} z = \frac{\partial z}{\partial \vec{x}} = \frac{\partial z}{\partial u_1} \frac{\partial u_1}{\partial \vec{x}} = \sigma(u_1) (1-\sigma(u_1)) \vec{y}.
\end{equation}
Note that the computation of $\frac{\partial z}{\partial u_1}$ can be reused for calculating $\grad_\vec{y} z$.
We get the gradient of the entire computation graph (including upstream nodes) with respect to $\vec{x}$ via $\grad_z\grad_\vec{x}z$.
Later we will use computation graphs where nodes are used in multiple downstream computations.
Such nodes receive multiple gradients from downstream nodes during backpropagation, which are summed up to calculate the gradient of the computation graph with respect to the variable represented by the node.

In \cref{log}, we will use backpropagation for computing the gradient of differentiable propositional logic rules  with respect to vector representations of symbols to develop models that combine representation learning with first-order logic.
In \cref{ntp}, we take this further and construct a computation graph for all possible proofs in a \gls{KB} using the backward chaining algorithm. 
This will allow us to calculate the gradient of proofs with respect to symbol representations and to induce rules using gradient descent.
Finally, in \cref{rte}, we will use \glspl{RNN}, \ie{}, computation graphs that are dynamically constructed for input varying-length input sequences of word representations.

\section{Function-free First-order Logic}
\label{sec:fol}

We now turn to a brief introduction of function-free first-order logic to the extent it is used in subsequent chapters.
This section follows the syntax of Prolog and Datalog \citep{gallaire1978logic}, and is based on \cite{lloyd1987foundations,nilsson1990logic}, and \cite{dzeroski2007inductive}.

\subsection{Syntax}

\begin{table}[t!]
  \centering
  \resizebox{\textwidth}{!}{
    \begin{tabular}{llll}
      \toprule
      & Prolog/Datalog Syntax & List Representation\\
      \midrule
      1 & $\rel{fatherOf}(\ent{abe}, \ent{homer}).$ &  $[[\rel{fatherOf}, \ent{abe}, \ent{homer}]]$\\
      2 & $\rel{parentOf}(\ent{homer}, \ent{lisa}).$ & $[[\rel{parentOf}, \ent{homer}, \ent{lisa}]]$\\
      3 & $\rel{parentOf}(\ent{homer}, \ent{bart}).$ & $[[\rel{parentOf}, \ent{homer}, \ent{bart}]]$\\
      4 & $\rel{grandpaOf}(\ent{abe}, \ent{lisa}).$ & $[[\rel{grandpaOf}, \ent{abe}, \ent{lisa}]]$\\
      5 & $\rel{grandfatherOf}(\ent{abe}, \ent{maggie}).$ & $[[\rel{grandfatherOf}, \ent{abe}, \ent{maggie}]]$\\
      6 & $\rel{grandfatherOf}(\var{X}_1, \var{Y}_1) \lif$ & $[[\rel{grandfatherOf}, \var{X}_1, \var{Y}_1], $\\
      & $\qquad\rel{fatherOf}(\var{X}_1, \var{Z}_1),$ & $\qquad[\rel{fatherOf}, \var{X}_1, \var{Z}_1],$\\
      & $\qquad\rel{parentOf}(\var{Z}_1, \var{Y}_1).$ & $\qquad[\rel{parentOf}, \var{Z}_1, \var{Y}_1]]$\\
      7 & $\rel{grandparentOf}(\var{X}_2, \var{Y}_2) \lif$ & $[[\rel{grandparentOf}, \var{X}_2, \var{Y}_2], $\\ 
      & $\qquad\rel{grandfatherOf}(\var{X}_2, \var{Y}_2).$ & $\qquad[\rel{grandfatherOf}, \var{X}_2, \var{Y}_2]]$\\
      \midrule
      $\set{P}$ & \multicolumn{2}{l}{$\{\rel{fatherOf}, \rel{parentOf}, \rel{grandpaOf}, \rel{grandfatherOf}, \rel{grandparentOf}\}$}\\
      $\set{C}$ & \multicolumn{2}{l}{$\{\ent{abe}, \ent{homer}, \ent{lisa}, \ent{bart}, \ent{maggie}\}$}\\
      $\set{V}$ & $\{\var{X}_1, \var{Y}_1, \var{Z}_1, \var{X}_2, \var{Y}_2\}$\\
      \bottomrule
    \end{tabular}
  }
  \caption{Example knowledge base using Prolog syntax (left) and as list representation as used in the backward chaining algorithm (right).}
  \label{tab:folkb}
\end{table}

We start by defining an \emph{atom} as a \emph{predicate}\footnote{We will use predicate and relation synonymously throughout this thesis.} symbol and a list of terms.
We will use lowercase names to refer to predicate and constant symbols (\eg{} \rel{fatherOf} and \ent{bart}), and uppercase names for variables (\eg{} \var{X}, \var{Y}, \var{Z}).
In Prolog, one also considers function terms and defines constants as function terms with zero arity. 
However, in this thesis we will work only with  function-free first-order logic rules, the subset of logic that Datalog supports.
Hence, for us a \emph{term} can be a constant or a variable.
For instance, $\rel{grandfatherOf}(\var{Q},\ent{bart})$ is an atom with the predicate $\rel{grandfatherOf}$, and two terms, the variable $\var{Q}$ and the constant $\ent{bart}$, respectively.
We define the \emph{arity} of a predicate to be the number of terms it takes as arguments. 
Thus, $\rel{grandfatherOf}$ is a binary predicate.
A \emph{literal} is defined as a negated or non-negated atom.
A \emph{ground literal} is a literal with no variables (see rules $1$ to $5$ in \cref{tab:folkb}).
Furthermore, we consider first-order logic \emph{rules}\footnote{We will use rule, clause and formula synonymously.} of the form $\ls{H} \lif \lss{B}$, where the body $\lss{B}$ (also called condition or premise) is a possibly empty conjunction of atoms represented as a list, and the head $\ls{H}$ (also called conclusion, consequent or hypothesis) is an atom.
Examples are rules $6$ and $7$ in \cref{tab:folkb}.
Such rules with only one atom as the head are called \emph{definite rules}.
In this thesis we only consider definite rules.
Variables are universally quantified (\eg{} $\forall\var{X}_1,\var{Y}_1,\var{Z}_1$ in rule $6$).
A rule is a ground rule if all its literals are ground.
We call a ground rule with an empty body a \emph{fact}, hence the rules $1$ to $5$ in \cref{tab:folkb} are facts.\footnote{We sometimes only call a rule a rule if it has a non-empty body. This will be clear from the context.}
We define $\set{S} = \set{C} \union \set{P} \union \set{V}$ to be the set of symbols, containing constant symbols $\set{C}$, predicate symbols $\set{P}$, and variable symbols $\set{V}$. 
We call a set of definite rules like the one in \cref{tab:folkb} a \emph{knowledge base} or \emph{logic program}.
A \emph{substitution} $\subs=\{\var{X}_1/t_1,\ldots,\var{X}_N/t_N\}$ is an assignment of variable symbols $\var{X}_i$ to terms $t_i$, and applying a substitution to an atom replaces all occurrences of variables $\var{X}_i$ by their respective term $t_i$.

What we have defined so far is the syntax of logic used in this thesis.
To assign meaning to this language (semantics), we need to be able to derive the truth value for facts.
We focus on proof theory, \ie{}, deriving the truth of a fact from other facts and rules in a \gls{KB}.\footnote{See \cite{dzeroski2007inductive} for other methods for semantics.}
In the next subsection we explain backward chaining, an algorithm for deductive reasoning. 
It is used to derive atoms from other atoms by applying rules.

\subsection{Deduction with Backward Chaining}
\label{sec:backchain}
Representing knowledge (facts and rules) in symbolic form has the appeal that one can use automated deduction systems to infer new facts.
For instance, given the logic program in \cref{tab:folkb}, we can automatically deduce that $\rel{grandfatherOf}(\ent{abe},\ent{lisa})$ is a true fact by applying rule 6 using facts 1 and 2.

\begin{figure}[t!]
  \centering
\begin{flalign*}
1. &\ \ \fun{or}(\ls{G}, \state) = \xs{\state' \ |\ \state' \in \fun{and}(\lss{B}, \fun{unify}(\ls{H}, \ls{G}, \state)) \text{ for } \ls{H} \lif \lss{B} \in \kb}&\\[0.75em]
2. &\ \ \fun{and}(\_,\fail) = \fail&\\
3. &\ \ \fun{and}(\emptylist,\state) = \state&\\
4. &\ \ \fun{and}(\ls{G}:\lss{G},\state) = \xs{\state''\ |\ \state''\in\fun{and}(\lss{G},\state') \text{ for } \state' \in \fun{or}(\fun{substitute}(\ls{G}, \state),\state)}&\\[0.75em]
5. & \ \ \fun{unify}(\_, \_, \fail) = \fail&\\
6. & \ \ \fun{unify}(\emptylist, \emptylist, \state) = \state&\\
7. & \ \ \fun{unify}(\emptylist, \_, \_) = \fail&\\
8. & \ \ \fun{unify}(\_, \emptylist, \_) = \fail&\\
9. & \ \ \fun{unify}(h:\lst{H}, g:\lst{G}, \state) = \fun{unify}\left(\lst{H},\lst{G},
\left\{\begin{array}{ll}
\state\union\{h/g\}         & \text{if } h\in \set{V}\\
\state\union\{g/h\}         & \text{if } g\in \set{V}, h\not\in \set{V}\\
\state & \text{if } g = h \\
\fail & \text{otherwise}
\end{array}\right\}\right)\\[0.75em]
10. &\ \ \fun{substitute}(\emptylist, \_) = \emptylist&\\
11. &\ \ \fun{substitute}(g : \ls{G}, \state) = 
\left\{\begin{array}{ll}
x & \text{if } g/x \in \state\\
g & \text{otherwise}
\end{array}\right\}
: \fun{substitute}(\ls{G}, \state)
&\\
\end{flalign*}
  \caption{Simplified pseudocode for symbolic backward chaining (cycle detection omitted for brevity, see \cite{russell2010artificial,gelder1987efficient,gallaire1978logic} for details).}
  \label{alg:backchain}
\end{figure}

Backward chaining is a common method for automated theorem proving, and we refer the reader to \cite{russell2010artificial,gelder1987efficient,gallaire1978logic} for details and to \cref{alg:backchain} for an excerpt of the pseudocode in style of a functional programming language.
Particularly, we are making use of pattern matching to check for properties of arguments passed to a module.
Note that "$\_$" matches every argument and that the order matters, \ie{}, if arguments match a line, subsequent lines are not evaluated.
We denote sets by Euler script letters (\eg{} $\set{E}$), lists by small capital letters (\eg{} $\lst{E}$), lists of lists by blackboard bold letters (\eg{} $\mathbb{E}$) and we use $:$ to refer to prepending an element to a list (\eg{} $e:\lst{E}$ or $\ls{E}:\lss{E}$). While an atom is a list of a predicate symbol and terms, a rule can be seen as a list of atoms and thus a list of lists where the head of the list is the rule head.\footnote{For example, $[[\rel{grandfatherOf},\var{X}, \var{Y}], [\rel{fatherOf},\var{X},\var{Z}],[\rel{parentOf},\var{Z},\var{Y}]]$.}

Given a goal such as $\rel{grandparentOf}(\var{Q}_1, \var{Q}_2)$, backward chaining finds substitutions of free variables with constants in facts in a \gls{KB} (\eg{} $\{\var{Q}_1/\const{abraham},\var{Q}_2/\const{bart}\}$). 
This is achieved by recursively iterating through rules that translate a goal into subgoals which we attempt to prove, thereby exploring possible proofs. 
For example, the \gls{KB} could contain the following rule that can be applied to find answers for the above goal:\\ 
\[
\rel{grandfatherOf}(\var{X},\var{Y}) \lif \rel{fatherOf}(\var{X},\var{Z}), \rel{parentOf}(\var{Z},\var{Y}).
\]

\begin{figure}[t!]
  \centering
  \includegraphics[width=\textwidth]{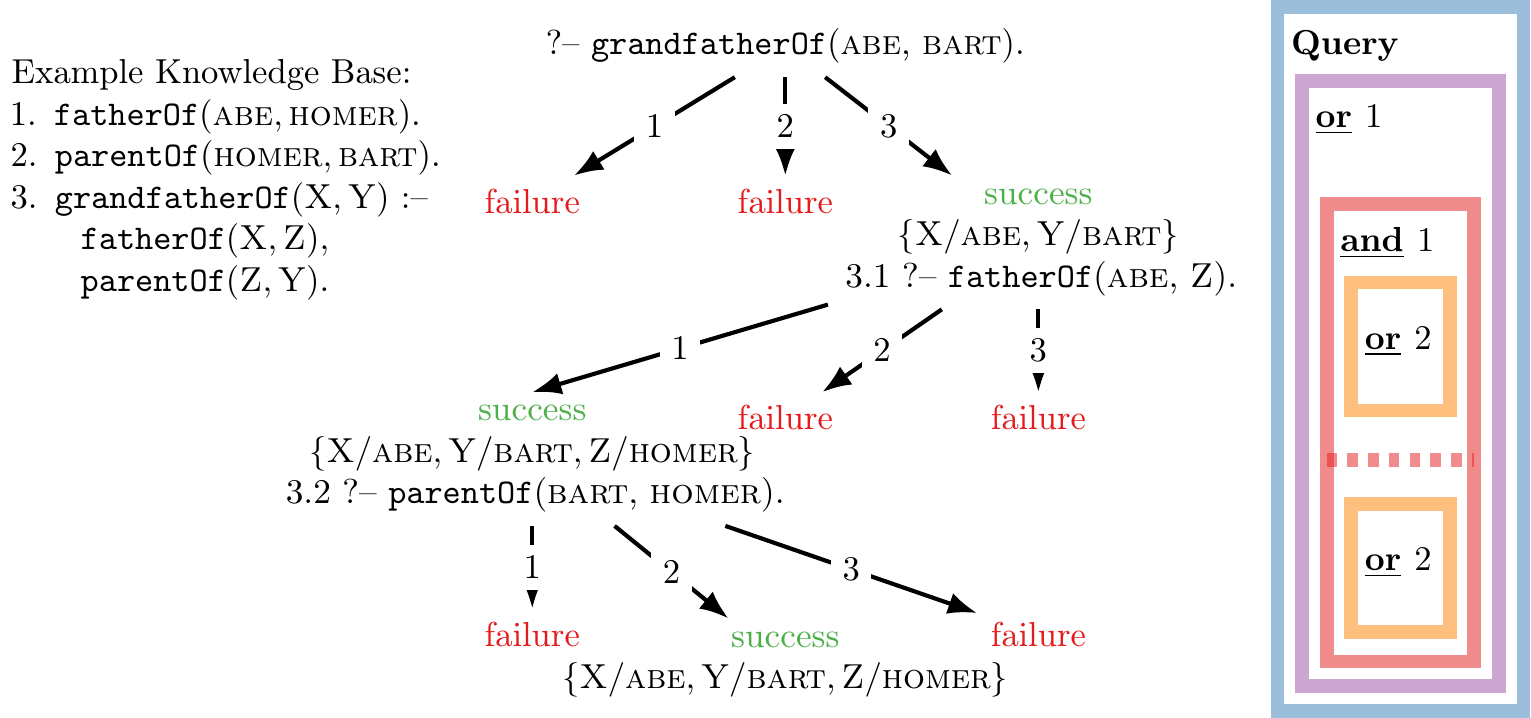}
  \caption{Full proof tree for a small knowledge base.}
  \label{fig:proof_bc}
\end{figure}

The proof exploration in backward-chaining is divided into two functions called \underline{or} and \underline{and} that perform a depth-first search through the space of possible proofs.
The function \underline{or} (line 1) attempts to prove a goal by unifying it with the head of every rule in a KB, yielding intermediate substitutions.
Unification (lines 5-9) iterates through pairs of symbols in the two lists corresponding to the atoms that we want to unify and updates the substitution set if one of the two symbols is a variable. It returns a failure if two non-variable symbols are not identical or the two atoms have different arity.
For rules where unification with the head of the rule succeeds, the body and substitution are passed to \underline{and} (lines 2-4), which then attempts to prove every atom in the body sequentially by first applying substitutions and subsequently calling \underline{or}.
This is repeated recursively until unification fails, atoms are proven by unification via grounding with facts in the KB, or a certain proof-depth is exceeded.
\Cref{fig:exampleproof} shows a proof for the query $\rel{grandparentOf}(\var{Q}_1, \var{Q}_2)$ given the \gls{KB} in \cref{tab:folkb} using \cref{alg:backchain}.
The method \underline{substitute} (lines 10-11) replaces variables in an atom by the corresponding symbol if there exists a substitution for that variable in the substitution list.
\Cref{fig:proof_bc} shows the full proof tree for a small knowledge base and the query ?-- $\rel{grandfatherOf}(\ent{abe}, \ent{bart})$.\footnote{We denote queries by the prefix ``?--''.}
The numbers on the arrows correspond to the application of the respective rules.
We visualize the recursive calls to \underline{or} and \underline{and} together with the proof depth on the right side.
Note how most proofs can be aborted early due to unification failure.

\begin{table}[t!]
  \centering
\resizebox{\textwidth}{!}{
  \begin{tabular}{lll}
\toprule
  Rule & Remaining Goals & $S$\\
  \midrule
    & $[[\rel{grandparentOf}, \var{Q}_1, \var{Q}_2]]$ & $\{\ \}$\\
  7 & $[[\rel{grandfatherOf}, \var{Q}_1, \var{Q}_2]]$ & $\{\var{X}_2/\var{Q}_1, \var{Y}_2/\var{Q}_2\}$ \\
  6 & $[[\rel{fatherOf}, \var{Q}_1, \var{Z}_1], [\rel{parentOf}, \var{Z}_1,\var{Q}_2]]$ & $\{\var{X}_2/\var{Q}_1, \var{Y}_2/\var{Q}_2, \var{X}_1/\var{Q}_1, \var{Y}_1/\var{Q}_2\}$ \\
  1 & $[[\rel{parentOf}, \const{homer},\var{Q}_2]]$ & $\{\var{X}_2/\var{Q}_1, \var{Y}_2/\var{Q}_2, \var{X}_1/\var{Q}_1, \var{Y}_1/\var{Q}_2,\var{Q}_1/\const{abe}, \var{Z}_1/\const{homer}\}$\\
  2 & $[\ ]$ & $\{\var{X}_2/\var{Q}_1, \var{Y}_2/\var{Q}_2, \var{X}_1/\var{Q}_1, \var{Y}_1/\var{Q}_2,\var{Q}_1/\const{abe}, \var{Z}_1/\const{homer}, \var{Q}_2/\const{lisa}\}$\\
\bottomrule
  \end{tabular}
}
  \caption{Example proof using backward chaining.}
  \label{fig:exampleproof}
\end{table}

Though first-order logic can be used for complex multi-hop reasoning, a drawback is that for such symbolic inference there is no generalization beyond what we explicitly specify in the facts and rules.
For instance, given a large \gls{KB} we would like to learn automatically that in many cases where we observe $\rel{fatherOf}$ we also observe $\rel{parentOf}$. 
This can be approached by statistical relational learning, inductive logic programming, and particularly neural networks for \gls{KB} completion which we discuss in the remainder of this chapter.

\subsection{Inductive Logic Programming}
\label{sec:ilp}
While backward chaining is used for \emph{deduction}, \ie{}, for inferring facts given rules and other facts in a \gls{KB},
\gls{ILP} \citep{muggleton1991inductive} combines logic programming with inductive learning to learn logic programs from training data.
This training data can include facts, but also a provided logic program that the \gls{ILP} system is supposed to extend.
Specifically, given facts and rules, the task of an \gls{ILP} system is to find further regularities and form hypotheses for unseen facts  \citep{dzeroski2007inductive}.\todo{example using Simpsons kinship}
Crucially, these hypothesis are again formulated using first-order logic.

There are different variants of \gls{ILP} learning tasks \citep{raedt1997logical} and 
we focus on \emph{learning from entailment} \citep{muggleton1991inductive}.
That is, given examples of positive facts and negative facts, the \gls{ILP} system is supposed to find rules such that positive facts can be deduced, but negative facts cannot.
There are many variants of \gls{ILP} systems, and we refer the reader to \cite{muggleton1994inductive} and \cite{dzeroski2007inductive} for an overview.
One of the most prominent systems is the \gls{FOIL} \citep{quinlan1990learning}, which is a greedy algorithm that induces one rule at a time by constructing a body that satisfies the maximum number of positive facts and the minimum number of negative facts.

In \cref{ntp}, we will construct neural networks for proving facts in a \gls{KB} and introduce a method for inducing logic programs using gradient descent while learning vector representations of symbols.
\section{Automated Knowledge Base Completion}
\label{sec:akbc}
Automated \gls{KB} completion is the task of inferring facts from information contained in a \gls{KB} and other resources such as text.
This is an important task as real-world \glspl{KB} are usually incomplete.
For instance, the \rel{placeOfBirth} predicate is missing for $71\%$ of people in Freebase \citep{dong2014knowledge}.
Prominent recent approaches to automated \gls{KB} completion learn vector representations of symbols via neural link prediction models.
The appeal of learning such \emph{subsymbolic} representations lies in their ability to capture similarity and even implicature directly in a vector space.
Compared to \gls{ILP} systems, neural link prediction models do not rely on a combinatorial search over the space of logic programs, but instead learn a local scoring function based on subsymbolic representations using continuous optimization.
However, this comes at the cost of uninterpretable models and no straightforward ways of incorporating logical background knowledge -- drawbacks that we seek to address in this thesis.
Another benefit of neural link prediction models over \gls{ILP} is that inferring whether a fact is true or not often  amounts to efficient algebraic operations (feed-forward passes in shallow neural networks), which makes test-time inference very scalable.
In addition, representations of symbols can be compositional.
For instance, we can compose a representation of a natural language predicate from a sequence of word representations \citep{verga2015multilingual}.

In recent years, many models for automated \gls{KB} completion have been proposed.
In the next sections, we discuss prominent approaches.
On a high level, these methods can be categorized into (i) neural link prediction models which define a local scoring function for the truth of a fact based on estimated symbol representations, and (ii) models that use paths between two entities in a \gls{KB} for predicting new relations between them.
\todo{LOG and FOIL are related to score-based methods, NTP is related to path-based methods}

\subsection{Matrix Factorization}
\label{sec:mf}
In this section, we describe the matrix factorization relation extraction model by \newcite{riedel2013relation}, which is an instance of a simple neural link prediction model.
We discuss this model in detail, as it the basis on which we develop rule injection methods in \cref{log} and \ref{foil}. 

Assume a set of observed entity pair symbols $\set{C}$ and a set of predicate symbols $\set{P}$, which can either represent structured binary relations from Freebase, a large collaborative knowledge base \citep{bollacker2008freebase}, or unstructured \gls{OpenIE} \citep{etzioni2008open} textual surface patterns collected from news articles.
Examples for structured and unstructured relations are $\rel{company/founders}$ and $\rel{\#2-co-founder-of-\#1}$, respectively.
Here, $\rel{\#2-co-founder-of-\#1}$ is a textual pattern where $\rel{\#1}$ and $\rel{\#2}$ are placeholders for entities.
For instance, the relationship between Elon Musk and Tesla in the sentence ``Elon Musk, the co-founder of Tesla and the CEO of SpaceX, cites The Foundation Trilogy by Isaac Asimov as a major influence on his thinking''\footnote{\url{http://www.dailymail.co.uk/news/article-4045816}} could be expressed by the ground atom \rel{\#2-co-founder-of-\#1}(\ent{tesla}, \ent{elon musk}). 
In this example, $\ent{elon musk}$ appeared first in the textual pattern, but $\rel{\#2}$ indicates that this constant will be used as the second argument in the predicate corresponding to the pattern. 
That way we can later introduce a rule $\forall \var{X},\var{Y}: \rel{company/founders}(\var{X},\var{Y}) \lif \rel{\#2-co-founder-of-\#1}(\var{X}, \var{Y})$ without changing the order of the variables in the body of the rule.

Let $\set{O} = \{r_s(e_i,e_j))\}$ be the set of observed ground atoms.
Model F by \newcite{riedel2013relation} maps all symbols in a knowledge base to subsymbolic representations, \ie{}, it learns a dense $k$-dimensional vector representation for every relation and entity pair.
Thus, a training fact $r_s(e_i, e_j)$ is represented by two vectors, $\vec{v}_s\in\R^k$ and $\vec{v}_{ij}\in\R^k$, respectively.
We will refer to these as \emph{embeddings}, \emph{subsymbolic representations}, \emph{vector representations}, \emph{neural representations}, or simply (symbol) \emph{representations} when it is clear from the context.
The truth estimate of a fact is modeled via the sigmoid of the dot product of the two symbol representations:
\begin{equation}
  \label{eq:mf}
  p_{sij} = \sigma(\vec{v}^\top_s\vec{v}_{ij}).
\end{equation}
In fact, this expression corresponds to the computation graph shown in \cref{fig:sigm_xy} with $\vec{x} = \vec{v}_s$ and $\vec{y} = \vec{v}_{ij}$.
The score $p_{sij}$ is measuring the compatibility between the relation and entity pair representation and can be interpreted as the probability of the fact being true conditioned on parameters of the model.
We would like to train symbol representations such that true ground atoms get a score close to one and false ground atoms a score close to zero.
This results in a low-rank matrix factorization corresponding to \gls{GPCA} \citep{collins2001generalization}
\begin{align}
  \label{eq:5}
  \mat{K} \approx \sigma(\mat{P}\mat{C}^\top) & \qquad\in\R^{|\set{P}|\times|\set{C}|}
\end{align}
where $\mat{P}\in\R^{|\set{P}|\times k}$ and $\mat{C}\in\R^{|\set{C}|\times k}$ are matrices of all relation and entity pair representations, respectively. 
\begin{figure}[t!]
  \centering
  \includegraphics{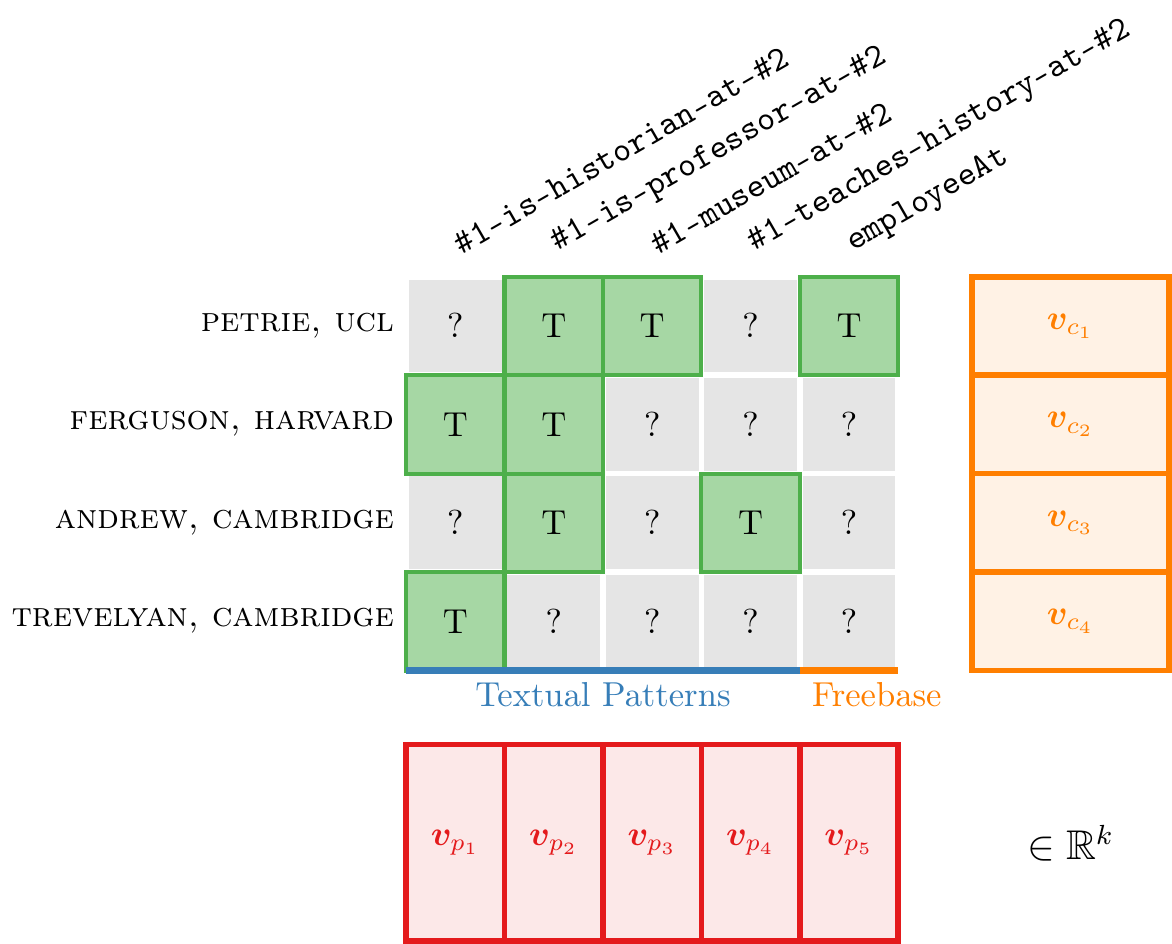}
  \caption{Knowledge base inference as matrix completion with true training facts (green), unobserved facts (question mark), relation representations (red), and entity pair representations (orange).}
  \label{fig:mf_example}
\end{figure}
This factorization is depicted in \cref{fig:mf_example} where known facts are shown in green and the task is to complete this matrix for cells with a question mark.
\Cref{eq:5} leads to generalization with respect to unseen facts as every relation and entity pair is represented in a low-dimensional space, and this information bottleneck will lead to similar entity pairs being represented close in distance in the vector space (likewise for similar predicates).

The distributional hypothesis states that ``one shall know a word by the company it keeps'' \citep{firth1957synopsis}.
It has been used for learning word meaning from large collections of text \citep{lowe2000direct, pado2007dependency}.
Applied to automated \gls{KB} completion, one could say that the meaning of a relation or entity pair can be estimated from the entity pairs and relations that respectively appear together in facts in a \gls{KB}.
That is, if for many entity pairs we observe that both \rel{fatherOf} and \rel{dadOf} is true, we can assume that both relations are similar.

\subsubsection{Bayesian Personalized Ranking}
\label{sec:bpr}
A common problem in automated \gls{KB} completion is that we do not observe any negative facts. 
In the recommendation literature, this problem is called implicit feedback \citep{rendle2009bpr}.
Applied to \gls{KB} completion, we would like to infer (or recommend) for a target relation some (unobserved) facts that we do not know from facts that we do know.
Facts that we do not know can be unknown either because they are not true or because they are true but missing in the \gls{KB}.

\begin{figure}[t!]
  \centering
  \includegraphics[scale=0.85]{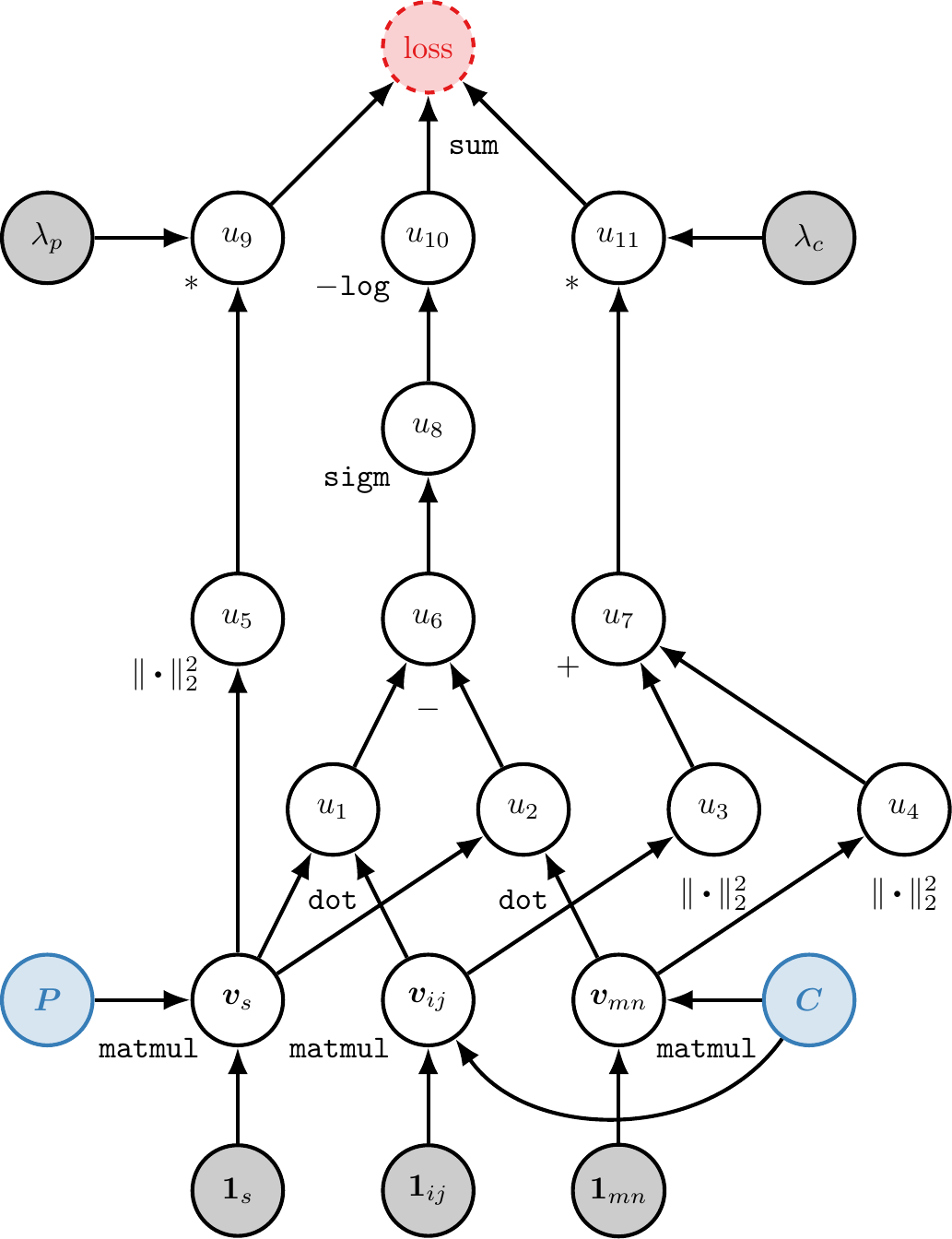}
  \caption{A complete computation graph of a single training example for Bayesian personalized ranking with $\ell_2$-regularization.}
  \label{fig:mf_bpr}
\end{figure}

One method to address this issue is to formulate the problem in terms of a ranking loss, and sample unobserved facts as negative facts during training.
Given a known fact $r_s(e_i, e_j)\in\set{O}$, \gls{BPR} \citep{rendle2009bpr} samples another entity pair $(e_m,e_n) \in\set{C}$ such that $r_s(e_m,e_n)\not\in\set{O}$ and adds the soft-constraint
\begin{equation}
\vec{v}_s^\top\vec{v}_{ij} \geq \vec{v}_s^\top\vec{v}_{mn}. 
\end{equation}
This follows a relaxation of the local \gls{CWA} \citep{galarraga2013mining,dong2014knowledge}.
In the \gls{CWA}, the knowledge about true facts for relation $r_s$ is assumed to be complete and any sampled unobserved fact can consequently be assumed to be negative. 
In \gls{BPR}, the assumption is that unseen facts are not necessarily false but their probability of being true should be less than for known facts.
Thus, sampled unobserved facts should have a lower score than known facts.
Instead of working with a fixed set of samples, we resample negative facts for every known fact in every epoch, where an epoch is a full iteration through all known facts in a \gls{KB}.
We denote a sample from the set of constant pairs as $(e_m,e_n)\sim\set{C}$.\footnote{Note that $\set{C}$ is the set of constant-pairs.}
This leads to the overall approximate loss
\begin{equation}
  \label{eq:mf_loss}
  \globalloss = \sum_{\substack{r_s(e_i,e_j)\ \in\ \set{O},\\(e_m,e_n)\ \sim\ \set{C},\\r_s(e_m,e_n)\ \not\in\ \set{O}}} - w_s \log\sigma(\vec{v}_s^\top\vec{v}_{ij} - \vec{v}_s^\top\vec{v}_{mn}) + \lambda_p\|\vec{v}_s\|_2^2+\lambda_c(\|\vec{v}_{ij}\|_2^2+\|\vec{v}_{mn}\|_2^2)
\end{equation}
where $\lambda_p$ and $\lambda_c$ are the strength of the $\ell_2$ regularization of the relation and entity pair representations, respectively.
Furthermore, $w_s$ is a relation-dependent implicit weight accounting for how often $(e_m,e_n)$ is sampled during training time.
As we resample an unobserved fact every time we visit an observed fact during training, unobserved facts for relations with many observed facts are more likely to be sampled as negative facts than for relations with few observed facts.
A complete computation graph of a single training example for matrix factorization using \gls{BPR} with $\ell_2$ regularization is shown in \cref{fig:mf_bpr}.

\subsection{Other Neural Link Prediction Models}
\label{sec:score}
An example for a simple neural link prediction model is the matrix factorization approach from the previous section. 
Alternative methods define the score $p_{sij}$ in \cref{eq:mf} in different ways, use different loss functions, and parameterize relation and entity (or entity pair) representations differently.
For instance, \cite{bordes2011learning} train two projection matrices per relation, one for the left and one for the right-hand argument position, respectively.
Subsequently, the score of a fact is defined as the $\ell_1$ norm of the difference of the two projected entity argument embeddings 
\begin{equation}
  \label{eq:4}
  p_{sij} = \|\mat{M}_s^\text{left}\vec{v}_i-\mat{M}_s^\text{right}\vec{v}_j\|_1.
\end{equation}
Note that compared to the matrix factorization model by \cite{riedel2013relation} which embeds entity-pairs, here we learn individual entity embeddings.
Similarly, TransE \citep{bordes2013translating} models the score as the $\ell_1$ or $\ell_2$ norm of the difference between the right entity embedding and a translation of the left entity embedding via the relation embedding
\maybe{SE, SME Bordes}
\begin{equation}
  \label{eq:transe}
  p_{sij} = \|\vec{v}_i+\vec{v}_s-\vec{v}_j\|_1
\end{equation}
where $\vec{v}_i$ and $\vec{v}_j$ are constrained to be unit-length.
The computation graph for this model is shown in \cref{fig:transe} in \cref{sec:sym2vec}.\maybe{omitting the unit-length constraint}

RESCAL \citep{nickel2012factorizing} represents relations as matrices and defines the score of a fact as 
\begin{equation}
  \label{eq:rescal}
  p_{sij} = \vec{v}_i^\top\mat{M}_s\vec{v}_j.
\end{equation}
In contrast to the other models mentioned in this section, RESCAL is not optimized with \gls{SGD} but using alternating least squares~\citep{nickel2011three}.
TRESCAL \citep{chang2014typed} extends RESCAL with entity type constraints for Freebase relations.
Neural Tensor Networks \citep{socher2013reasoning} add a relation-dependent compatibility score to RESCAL
\begin{equation}
  \label{eq:ntn}
  p_{sij}= \vec{v}_i^\top\mat{M}_s\vec{v}_j+\vec{v}_i^\top\vec{v}_s^\text{left}+\vec{v}_j^\top\vec{v}_s^\text{right}
\end{equation}
and are optimized with L-BFGS \citep{byrd1995limited}.
DistMult by \cite{yang2014embedding} is modeling the score as the trilinear dot product
\begin{equation}
  \label{eq:distmul}
  p_{sij} = \vec{v}_s^\top(\vec{v}_i\odot\vec{v}_j) = \sum_k \vec{v}_{sk}\vec{v}_{ik}\vec{v}_{jk}
\end{equation}
where $\odot$ is the element-wise multiplication. This model is a special case of RESCAL where $\mat{M}_s$ is constrained to be diagonal.
ComplEx by \cite{trouillon2016complex} uses complex vectors $\vec{v}_s, \vec{v}_i, \vec{v}_j\in\C^k$ for representing relations and entities.
Let $\real(\vec{v})$ denote the real part and $\imag(\vec{v})$ the imaginary part of a complex vector $\vec{v}$.
The scoring function defined by ComplEx is 
\begin{align}
  \label{eq:complex}
  p_{sij} &= \real(\vec{v}_s)^\top(\real(\vec{v}_i)\odot\real(\vec{v}_j))\nonumber\\
  & \qquad + \real(\vec{v}_s)^\top(\imag(\vec{v}_i)\odot\imag(\vec{v}_j))\nonumber\\
  & \qquad + \imag(\vec{v}_s)^\top(\real(\vec{v}_i)\odot\imag(\vec{v}_j))\nonumber\\
  & \qquad - \imag(\vec{v}_s)^\top(\imag(\vec{v}_i)\odot\real(\vec{v}_j)).
\end{align}
The benefit of ComplEx over RESCAL and DistMult is that by using complex vectors it can capture symmetric as well as asymmetric relations.

Building upon \cite{riedel2013relation}, \cite{verga2015multilingual} developed a column-less factorization approach by encoding surface form patterns using \glspl{LSTM} \citep{hochreiter1997long} instead of learning a non-compositional representation.
Similarly, \cite{toutanova2015representing} uses \glspl{CNN} to encode surface form patterns.
In a follow-up study, \cite{verga2016generalizing} propose a row-less method where entity pair representations are not learned but instead computed from observed relations, thereby generalizing to new entity pairs at test time.

\subsection{Path-based Models}
\label{sec:path}
While all methods presented in the previous section model the truth of a fact as a local scoring function of the representations of the relation and entities (or entity pairs), path-based models score facts based either on random walks over the \glspl{KB} (path ranking) or by encoding entire paths in a vector space (path encoding).

\subsubsection{Path Ranking}
The \gls{PRA} \citep{lao2010relational,lao2011random} learns to predict a relation between two entities based on logistic regression over features collected from random walks between these entities in the \gls{KB} up to some predefined length.
\cite{lao2012reading} extend \gls{PRA} inference to \gls{OpenIE} surface patterns in addition to structured relations contained in \glspl{KB}. 

A related approach to \gls{PRA} is \gls{ProPPR} \citep{wang2013programming}, which is a first-order probabilistic logic programming language.
It uses Prolog's \gls{SLD} \citep{kowalski1971linear}, a depth-first search strategy for theorem proving, to construct a graph of proofs.
Instead of returning deterministic proofs for a given query, \gls{ProPPR} defines a stochastic process on the graph of proofs using PageRank \citep{page1999pagerank}.
Furthermore, in \gls{ProPPR} one can use features on the head of rules whose weights are learned from data to guide stochastic proofs.
Experiments with ProPPR were conducted on comparably small \glspl{KB} and contrary to neural link prediction models and the extensions to \gls{PRA} below, it has not yet been scaled to large real-world \glspl{KB} \citep{gardner2014incorporating}.

A shortcoming of \gls{PRA} and \gls{ProPPR} is that they are operating on symbols instead of vector representations of symbols.
This limits generalization as it results in an explosion in the number of paths to consider when increasing the path length.
To overcome this limitation, \cite{gardner2013improving} extend \gls{PRA} to include vector representations of verbs.
These verb representations are obtained from pre-training via \acrshort{PCA} on a matrix of co-occurrences of verbs and subject-object tuples collected from a large dependency-parsing corpus. 
Subsequently, these representations are used for clustering relations, thus avoiding an explosion of path features in prior \gls{PRA} work while improving generalization.
\cite{gardner2014incorporating} take this approach further by introducing vector space similarity into random walk inference, thus dealing with paths containing unseen surface forms by measuring the similarity to surface forms seen during training, and following relations proportionally to this similarity.

\subsubsection{Path Encoding}
While \citeauthor{gardner2013improving} introduced vector representations into \gls{PRA}, these representations are not trained end-to-end from task data but instead pretrained on an external corpus. 
This means that relation representations cannot be adapted during training on a \gls{KB}.

\cite{neelakantan2015compositional} propose \glspl{RNN} for learning embeddings of entire paths.
The input to these \glspl{RNN} are trainable relation representations.
Given a known relation between two entities and a path connecting the two entities in the \gls{KB}, an \gls{RNN} for the target relation is trained to output an encoding of the path such that the dot product of that encoding and the relation representation is maximal.

\cite{das2016chains} note three limitations of the work by \cite{neelakantan2015compositional}. 
First, there is no parameter sharing of \glspl{RNN} that encode different paths for different target relations.
Second, there is no aggregation of information from multiple path encodings.
Lastly, there is no use of entity information along the path as only relation representations are fed to the \gls{RNN}.
\citeauthor{das2016chains} address the first issue by using a single \gls{RNN} whose parameters are shared across all  paths.
To address the second issue, \citeauthor{das2016chains} train an aggregation function over the encodings of multiple paths connecting two entities.
Finally, to obtain entity representations that are fed into the \gls{RNN} alongside relation representations they sum learned vector representations of the entity's annotated Freebase types.
\maybe{\cite{neelakantan2015neural}}
\maybe{\cite{toutanova2016compositional}}
\maybe{\cite{neelakantan2016learning}}
\todo{Conclusion? How does it relate to this thesis? Can these methods make efficient use of prior logical knowledge?}

\chapter{Regularizing Representations by First-order Logic Rules}
\label{log}
\glsresetall

In this chapter, we introduce a paradigm for combining neural link prediction models for automated \gls{KB} completion (\cref{sec:score}) with background knowledge in the form of first-order logic rules.
We investigate simple baselines that enforce rules through symbolic inference before and after matrix factorization.
Our main contribution is a novel joint model that learns vector representations of relations and entity pairs using both distant supervision and first-order logic rules, such that these rules are captured directly in the vector space of symbol representations.
To this end, we map symbolic rules to differentiable computation graphs representing real-valued losses that can be added to the training objective of existing neural link prediction models.
At test time, inference is still efficient as only a local scoring function over symbol representations is used and no logical inference is needed.
We present an empirical evaluation where we incorporate automatically mined rules into a matrix factorization neural link prediction model. 
Our experiments demonstrate the benefits of incorporating logical knowledge for Freebase relation extraction. 
Specifically, we find that joint factorization of distant and logic supervision is efficient, accurate, and robust to noise~(\cref{sec:log_results}).
By incorporating logical rules, we were able to train relation extractors for which no or only few training facts are observed.

\section{Matrix Factorization Embeds Ground Literals}
\label{sec:log_mf}
In \cref{sec:mf}, we introduced matrix factorization as a method for learning representations of predicates and constant pairs for automated \gls{KB} completion.
In this section, we elaborate on how matrix factorization indeed embeds ground atoms in a vector space and lay out the foundation for developing a method that embeds first-order logic rules.

Let $\lrule{f} \in \set{F}$ denote a rule in a \gls{KB}. 
For instance, $\lrule{f}$ could be a ground rule without a body (\ie{} a fact) like $\rel{parentOf}(\const{homer}, \const{bart})$.
Furthermore, let $\ldiff{\lrule{f}}$ denote the probability of this rule being true conditioned on the parameters of the model.
For now, we restrict $\lrule{f}$ to ground atoms and discuss ground literals, propositional rules, and first-order rules later.
With slight abuse of notation, let  $\ldiff{\bdot}$ also denote the mapping of predicate or constant symbols (or a pair of constant symbols) to their subsymbolic representation as assigned by the model.
Note that this mapping depends on the neural link prediction model.
For matrix factorization, $\ldiff{\bdot}$ is a function $\set{S} \to \R^k$ from symbols (constant pairs and predicates) to $k$-dimensional dense vector representations. 
For RESCAL (see \cref{sec:score}), $\ldiff{\bdot}$ maps constants to $\R^k$ and predicates to $\R^{k\times k}$.\maybe{mention more (e.g. TransE); they also differ in the decomposition of constant pairs}
Using this notation, matrix factorization decomposes the probability of a fact $r_s(e_i,e_j)$ as
\begin{equation}
  \label{eq:6}
  \ldiff{r_s(e_i,e_j)} = \sigmoid(\ldiff{r_s}^\top\ldiff{e_i,e_j}) = \sigmoid(\vec{v}_s^\top\vec{v}_{ij}).
\end{equation}

\paragraph{Training Objective}
\cite{riedel2013relation} used \gls{BPR} \citep{rendle2009bpr} as training objective, \ie{}, they encouraged the score of known true facts to be higher than unknown facts (\cref{sec:bpr}).
However, as we will later model the probability of a rule from the probability of ground atoms scored by a neural link prediction model, we need to ensure that all scores are in the interval $[0,1]$. 
Instead of \gls{BPR}, we thus use the negative log-likelihood loss to directly maximize the probability of all rules, \emph{including ground atoms}, in a \gls{KB} (we omit $\ell_2$ regularization for brevity):
\begin{equation}
  \label{eq:log_loss}
  \globalloss = \sum_{\lrule{f}\ \in\ \set{F}} -\log(\ldiff{\lrule{f}}).
\end{equation}
Therefore, instead of learning to rank facts, we optimize representations to assign a score close to $1.0$ to rules (including facts).
Our model can thus be seen as generalization of a neural link prediction model to rules beyond ground atoms. 

For matrix factorization as neural link prediction model, we are embedding ground atoms in a vector space of predicate and constant pair representations.
Next, we will extend this to ground literals and afterwards to propositional and then first-order logic rules.

\paragraph{Negation}
Let $\lrule{f} = \lneg\lrule{g}$ be the negation of a ground atom $\lrule{g}$.
We can model the probability of $\lrule{f}$ as follows:
\begin{equation}
  \label{eq:9}
  \ldiff{\lrule{f}} = \ldiff{\lneg\lrule{g}} = 1-\ldiff{\lrule{g}}.
\end{equation}
By using ground literals in the training objective in \cref{eq:log_loss}, we can say that matrix factorization is embedding ground literals via learning predicate and constant pair representations.
In other words, given known ground literals (negated and non-negated facts), matrix factorization embeds symbols in a low-dimensional vector space such that a scoring function assigns a high probability to these ground literals.
As symbols are embedded in a low-dimensional vector space, this method can generalize to unknown facts and predict a probability for these at test time by placing similar symbols close in distance in the embedding space.

Note that so far we have not gained anything over matrix factorization as explained in \cref{sec:mf}. 
\Cref{eq:log_loss,eq:9} are only introducing notation that will make it easier to embed more complex rules later.
Now that we can embed negated and non-negated facts, we can ask the question whether we can also embed propositional and first-order logic rules.

\section{Embedding Propositional Logic}
\label{sec:prop}
We know from propositional logic that with the negation and conjunction operators we can model any other Boolean operator and propositional rule.
In \cref{eq:9}, we effectively turned a symbolic logical operation (negation) into a differentiable operation that can be used to learn subsymbolic representations for automated \gls{KB} completion.
If we can find such a differentiable operation for conjunction, then we could backpropagate through any propositional logical expression, and learn vector representations of symbols that encode given background knowledge in propositional logic.

\paragraph{Conjunction}
In Product Fuzzy Logic, conjunction is modeled using a Product t-Norm \citep{lukasiewicz1920logice}.
Let $\lrule{f} = \lrule{a} \land \lrule{b}$ be the conjunction of two propositional expressions $\lrule{a}$ and $\lrule{b}$.
The probability of $\lrule{h}$ is then defined as follows:
\begin{equation}
  \label{eq:10}
  \ldiff{\lrule{f}} = \ldiff{\lrule{a} \land \lrule{b}} = \ldiff{\lrule{a}}\ldiff{\lrule{b}}.
\end{equation}
In other words, we replaced conjunction, a symbolic logical operation, with multiplication, a differentiable operation.
Note that alternatives for modeling conjunction exist. For instance, one could take the min of $\ldiff{\lrule{a}}$ and $\ldiff{\lrule{b}}$ (Gödel t-Norm \citep{godel1932intuitionistischen}).

Given the probability of ground atoms, we can use Product Fuzzy Logic to calculate the probability of the conjunction of these atoms.
However, we will go a step further and assume that we know the ground truth probability of the conjunction of two atoms.
We can then use the negative log-likelihood loss to measure the discrepancy between the predicted probability of the conjunction and the ground truth.
Our contribution is backpropagating this discrepancy through the propositional rule and a neural link prediction model that scores ground atoms to calculate a gradient with respect to vector representations of symbols.
Subsequently, we update these representations using gradient descent, thereby encoding the ground truth of a propositional rule directly in the vector representations of symbols.
At test time, predicting a score for any unobserved ground atom $r_s(e_i,e_j)$ is done efficiently by calculating $\ldiff{r_s(e_i,e_j)}$. 

\paragraph{Disjunction}
Let $\lrule{f} = \lrule{a} \lor \lrule{b}$ be the disjunction of two propositional expressions $\lrule{a}$ and $\lrule{b}$.
Using De Morgan's law and \cref{eq:9,eq:10}, we can model the probability of $\lrule{f}$ as follows:
\begin{align}
  \label{eq:11}
  \ldiff{\lrule{f}} 
&= \ldiff{\lrule{a} \lor \lrule{b}} \nonumber\\ 
&= \ldiff{\lneg(\lneg(\lrule{a} \lor \lrule{b}))} \nonumber\\
&= \ldiff{\lneg(\lneg\lrule{a} \land \lneg\lrule{b})} \nonumber\\
&= 1-(1-\ldiff{\lrule{a}})(1-\ldiff{\lrule{b}}) \nonumber\\
&= \ldiff{\lrule{a}}+\ldiff{\lrule{b}}-\ldiff{\lrule{a}}\ldiff{\lrule{b}}. 
\end{align}
Note that \cref{eq:9} not only holds for ground atoms, but any propositional logical expression.
Furthermore, any propositional logical expression can be normalized to \acrlong{CNF}.
Thus, with \crefrange{eq:9}{eq:11} we now have a way to construct a differentiable computation graph, and thus a real-valued loss term, for any symbolic expression in propositional logic.

\begin{figure}[t!]
  \centering
  \includegraphics[scale=0.8]{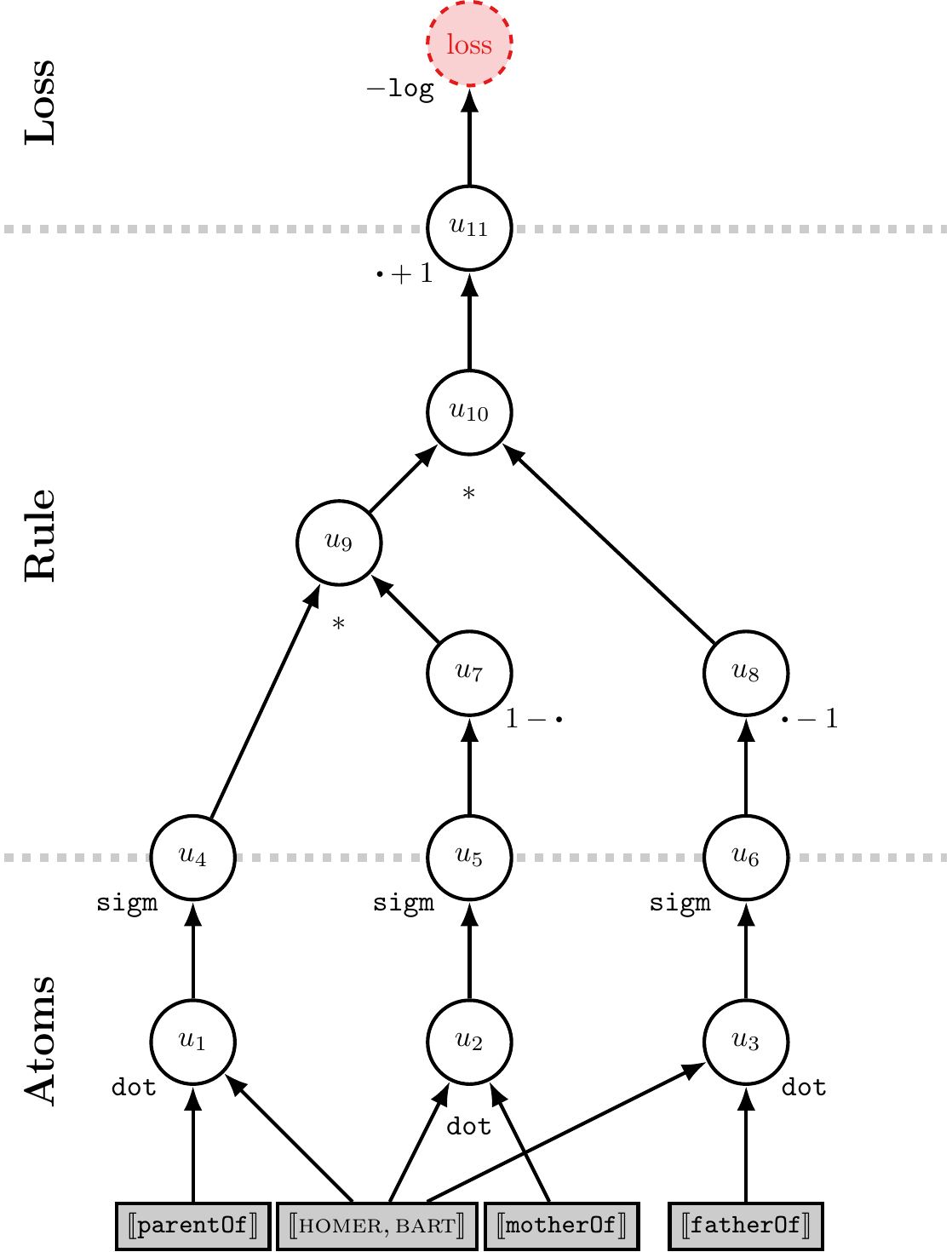}
  \caption{Computation graph for rule in \cref{eq:13} where $\cdot$ denotes a placeholder for the output of the connected node.}
  \label{fig:rule_ex}
\end{figure}

\paragraph{Implication}
A particular class of logical expressions that we care about in practice are propositional implication rules of the form $\ls{H} \lif \lss{B}$, where the body $\lss{B}$ is a possibly empty conjunction of atoms represented as a list, and the head $\ls{H}$ is an atom.
Let $\lrule{f} = \lrule{h} \lif \lrule{b}$. 
The probability of $\lrule{f}$ is then modeled as:
\begin{align}
  \label{eq:12}
  \ldiff{\lrule{f}} &= \ldiff{\lrule{h} \lif \lrule{b}} \nonumber\\
                    &= \ldiff{\lneg\lrule{b} \lor \lrule{h}} \nonumber\\
                    &= \ldiff{\lneg\lrule{b}} + \ldiff{\lrule{h}} - \ldiff{\lneg\lrule{b}}\ldiff{\lrule{h}} \nonumber\\
                    &= 1 - \ldiff{\lrule{b}} +  \ldiff{\lrule{h}} - (1-\ldiff{\lrule{b}})\ldiff{\lrule{h}} \nonumber\\
                    &= 1 - \ldiff{\lrule{b}} +  \ldiff{\lrule{h}} - \ldiff{\lrule{h}} + \ldiff{\lrule{b}}\ldiff{\lrule{h}} \nonumber\\
                    &= 1 - \ldiff{\lrule{b}} +  \ldiff{\lrule{b}}\ldiff{\lrule{h}} \nonumber\\
                    &= \ldiff{\lrule{b}}(\ldiff{\lrule{h}}-1) +1.
\end{align}
Say we want to ensure that
\begin{align}
  \label{eq:13}
&\rel{fatherOf}(\ent{homer},\ent{bart}) \lif \nonumber\\ 
&\qquad\rel{parentOf}(\ent{homer},\ent{bart})\, \nonumber\\
&\qquad\lneg\rel{motherOf}(\ent{homer},\ent{bart}).
\end{align}
We now have a way to map this rule to a differentiable expression that we can use alongside facts in \cref{eq:log_loss} and optimize the symbol representations using gradient descent as we did previously for matrix factorization. 
The computation graph that allows us to calculate the gradient of this rule with respect to symbol representations is shown in \cref{fig:rule_ex}.
While the structure of the bottom part (Atoms) of this computation graph is determined by the neural link prediction model, the middle part (Rule) is determined by the propositional rule.
Note that we can use any neural link predictor (see \cref{sec:score}) instead of matrix factorization for obtaining a probability of ground atoms. 
The only requirement is that ground atom scores need to lie in the interval $[0,1]$. 
However, for models where this is not the case, we can always apply a transformation such as the sigmoid.
\maybe{add figures showing the loss shapes}
\info{we let the data speak for itself: if the is strong evidence against a rule, then the fact and rule objective will compete and what we get is determined by the low-dimensional space constraint}

\paragraph{Independence Assumption}
\Cref{eq:10} underlies a strong assumption, namely that the probability of the arguments of the conjunction are conditionally independent given symbol embeddings.\todo{Seb: formalize?} 
We already get a violation of this assumption for the simple case $\ldiff{\lrule{f}\land\lrule{f}}$ with $0 < \ldiff{\lrule{f}} < 1$, which results in $\ldiff{\lrule{f}\land\lrule{f}} = \ldiff{\lrule{f}}\ldiff{\lrule{f}} < \ldiff{\lrule{f}}$.
However, for dependent arguments we get an approximation to the probability of the conjunction that can still be used for gradient updates of the symbol representations, and we demonstrate empirically that conjunction as modeled in \cref{eq:10} is useful for improving automated \gls{KB} completion.
In \cref{foil}, we will present a way to avoid this independence assumption for implications.
\maybe{$\min$ avoids this independence assumption}
\maybe{Seb: where do we get such rules?}

\section{Embedding First-order Logic via Grounding}
\label{sec:method}

\begin{figure*}[tb]
  \centering
  \includegraphics[width=1.0\textwidth]{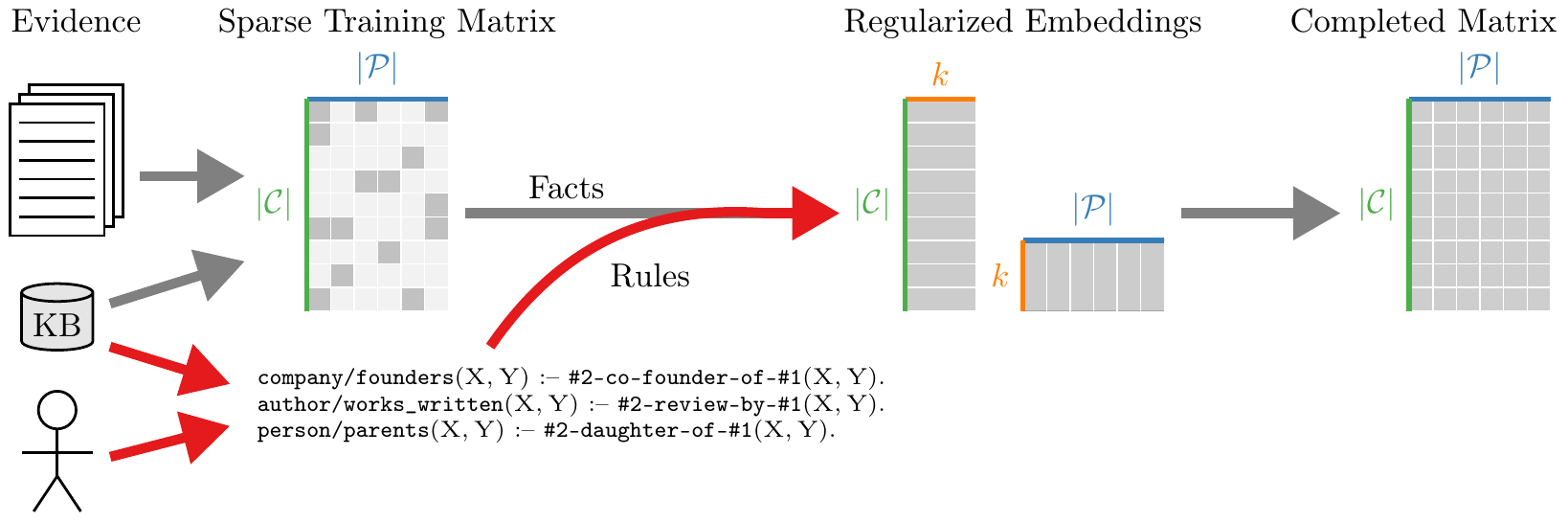}
  \caption{Given a set of training ground atoms, matrix factorization learns $k$-dimensional predicate and constant pair representations. Here, we also consider additional first-order logic rules (red) and seek to learn symbol representations such that the predictions (completed matrix) comply with these given rules.}
  \label{fig:mf}
\end{figure*}

Now that we can backpropagate through propositional Boolean expressions, we turn to embedding first-order logic rules in the vector space of symbol representations.
Note that in this process, we do not chain rules, \ie{}, we do not perform logical inference at training or test time.
Instead, a provided rule is used to construct a loss term for optimizing vector representations of symbols.
When training the model using gradient descent, we attempt to find a minimum of the global loss \cref{eq:log_loss} such that the probability of all rules (including facts) is high.
This minimum might be attained where the prediction of the neural link prediction model for scoring ground atoms not only agrees with a single but all rules, thereby predicting ground atoms as if we chained rules.
A high-level overview of this process for matrix factorization as neural link prediction model of ground atoms is shown in \cref{fig:mf}.

Assuming a finite set of constants, we can ground any first-order rule by replacing free variables with constants in the domain.
Let $\lrule{f}(\var{X},\var{Y})$ denote a universally quantified rule with two free variables $\var{X}$ and $\var{Y}$, then
\begin{equation}
  \label{eq:14}
  \ldiff{\forall\var{X},\var{Y}: \lrule{f}(\var{X},\var{Y})} = \ldiff{\bigwedge_{(e_i,e_j)\in\set{C}}\lrule{f}(e_i,e_j)}
\end{equation}
where $\set{C}$ is the set of all entity pairs.
With \cref{eq:10,eq:log_loss} we obtain the following ground loss for $\lrule{f}$:
\begin{equation}\small
  \label{eq:15}
  -\log\left(\ldiff{\bigwedge_{(e_i,e_j)\in\set{C}}\lrule{f}(e_i,e_j)}\right) = -\log\left(\prod_{(e_i,e_j)\in\set{C}}\ldiff{\lrule{f}(e_i,e_j)}\right) = -\sum_{(e_i,e_j)\in\set{C}}\log{\ldiff{\lrule{f}(e_i,e_j)}}.
\end{equation}
\maybe{discuss more}

\subsection{Stochastic Grounding}
\label{sec:grounding}
For large domains of constants, \cref{eq:15} becomes very costly to optimize as it would result in many expressions that are added to the training objective of the underlying neural link prediction model.
For rules over pairs of constants, we can reduce the number of terms drastically by only considering pairs of constants $\set{C}^\text{train}$ that appeared together in training facts.
Still, $\set{C}^\text{train}$ might be a prohibitively large set of constant pairs. 
Thus, we resort to a heuristic similar to \gls{BPR} for sampling constant pairs.
Given a rule, we obtain a ground propositional rule for every constant pair for which at least one atom in the rule is a known training fact when substituting free variables with these constants.
In addition, we sample as many constant pairs that appeared together in training facts but for which all atoms in the rule are unknown when substituting free variables with the sampled constant pairs.
\maybe{add example}
\maybe{write entire training objective (together with sampled groundings and implicit weight like \cref{eq:mf_loss})}
\check{Jo: does this introduce a bias? rules with more pairs are harder to assign a high score to than rules with few pairs?}

\subsection{Baseline}
\label{sec:pre-inference}
Background knowledge in form of first-order rules can be seen as \emph{hints} that can be used to generate additional training data~\citep{abu1990learning}.
For \emph{pre-factorization inference} we first perform symbolic logical inference on the training data using provided rules and add inferred facts as additional training data. 
For example, for a rule $r_t(\var{X}, \var{Y}) \lif r_s(\var{X},\var{Y})$, we add an additional observed training facts $r_t(e_i,e_j)$ for any pair of constants $(e_i,e_j)$ for which $r_s(e_i,e_j)$ is a true fact in the training data.
This is repeated until no further facts can be inferred. 
Subsequently, we run matrix factorization on the extended set of observed facts.
The intuition is that the additional training data generated by rules provide evidence of the logical dependencies between relations to the matrix factorization model, while at the same time allowing the factorization to generalize to unobserved facts and to deal with ambiguity and noise in the data. 
No further logical inference is performed during or after training of the factorization model as we expect that the learned embeddings encode the provided rules.

One drawback of pre-factorization inference is that the rules are enforced only on observed atoms, \ie{}, first-order dependencies on \emph{predicted} facts are ignored.
In contrast, with the loss in \cref{eq:15} we add terms for the rule directly to the matrix factorization objective,
thus jointly optimizing embeddings to reconstruct known facts, as well as to obey to provided first-order logical background knowledge.
However, as we stochastically ground first-order rules, we have no guarantee that a given rule will indeed hold for all possible entity pairs at test time. 
While we next demonstrate that this approach is still useful for \gls{KB} completion despite this limitation, in \cref{foil} we will introduce a method that overcomes this limitation for simple first-order implication rules.

\section{Experiments}
\label{sec:exps}

There are two orthogonal questions when evaluating the method above.
First, does regularizing symbol embeddings by first-order logic rules indeed capture such logical knowledge in a vector space and can it be used to improve \gls{KB} completion?
Second, where can we obtain background knowledge in form of rules that is useful for a particular \gls{KB} completion task?
The latter is a well-studied problem~\citep{hipp2000algorithms,schoenmackers10:learning,volker2011statistical}. 
Thus, we focus the evaluation on the ability of various approaches to benefit from rules that we directly extract from the training data using a simple method.

\paragraph{Distant Supervision Evaluation}
We follow the procedure of \citet{riedel2013relation} for evaluating knowledge base completion of Freebase \citep{bollacker2008freebase} with textual data from the NYT corpus~\citep{sandhaus08:the-new-york}. 
The training matrix consists of $4\,111$ columns, representing $151$ Freebase relations and $3\,960$ textual patterns, $41\,913$ rows (constant pairs) and $118\,781$ training facts of which $7\,293$ belong to Freebase relations. 
The constant pairs are divided into train and test, and we remove all Freebase facts for these test pairs from the training data.
Our primary evaluation measure is (weighted) \gls{MAP} \citep{manning2008introduction,riedel2013relation}.
Let $\set{R}$ be the set of test relations and let $\{f_{1j}, \ldots, f_{mj}\}$ be the set of test facts for relation $r_j\in\set{R}$. 
Furthermore, let $R_{kj}$ be the ranked list of facts for relation $r_j$ scored by a model up until fact $f_{kj}$ is reached.
\gls{MAP} is then defined as
\begin{align}
  \text{MAP}(\set{R}) = \frac{1}{|\set{R}|}\sum_{j=1}^{|\set{R}|}\frac{1}{m_j}\sum_{k=1}^{m_j}\text{precision}(R_{kj})
\end{align}
where precision calculates the fraction of correctly predicted test facts of all predicted facts.
For weighted \gls{MAP}, the average precision for every relation is weighted by the number of true facts for the respective relation \citep{riedel2013relation}.
Note that the \gls{MAP} metric operates only on the ranking of facts as predicted by the model and does not take the absolute predicted score into account.

\paragraph{Rule Extraction and Annotation}
\label{sec:setup:formulae}
We use a simple technique for extracting rules from a matrix factorization model based on \cite{sanchez2015towards}.
We first run matrix factorization over the complete training data to learn symbol representations.
After training, we iterate over all pairs of relations $(r_s,r_t)$ where $r_t$ is a Freebase relation.
For every relation pair we iterate over all training atoms $r_s(e_i,e_j)$, evaluate the score $\ldiff{r_t(e_i,e_j) \lif r_s(e_i,e_j)}$ using \cref{eq:12}, and calculate the average to arrive at a score as the proxy for the coverage of the rule. \check{is this similar to AIMEE?}
Finally, we rank all rules by their score and manually inspect and filter the top $100$, 
which resulted in $36$ annotated high-quality rules (see Table \ref{tab:formulae} for the top rules for five different Freebase target relations and \cref{appendixA} for the list of all annotated rules).
Note that our rule extraction approach does not observe the relations for test constant pairs and that all extracted rules are simple first-order logic expressions.
All models in our experiments have access to these rules, except for the matrix factorization baseline.

\begin{table}[tb]
  \centering
  \begin{tabular}{lll}
    \toprule
    Rule & Score\\
    \midrule

$\rel{org/parent/child}(\var{X},\var{Y}) \lif \rel{\#2-unit-of-\#1}(\var{X},\var{Y}).$ & $0.97$\\
$\rel{location/containedby}(\var{X},\var{Y}) \lif \rel{\#2-city-of-\#1}(\var{X},\var{Y}). $ & $0.97$\\
$\rel{person/nationality}(\var{X},\var{Y}) \lif \rel{\#2-minister-\#1}(\var{X},\var{Y}). $ & $0.97$\\
$\rel{person/company}(\var{X},\var{Y}) \lif \rel{\#2-executive-\#1}(\var{X},\var{Y}).$ & $0.96$\\
$\rel{company/founders}(\var{X},\var{Y}) \lif \rel{\#2-co-founder-of-\#1}(\var{X},\var{Y}).$ & $0.96$\\
    \bottomrule
  \end{tabular}
  \caption{Top rules for five different Freebase target relations. These implications were extracted from the matrix factorization model and manually annotated. The premises of these implications are shortest  paths between entity arguments in dependency tree, but we present a simplified version to make these patterns more readable. See \cref{appendixA} for the list of all annotated rules.}
  \label{tab:formulae}
\end{table}

\paragraph{Methods}
Our proposed methods for injecting logic into symbol embeddings are \emph{pre-factorization inference} (\textbf{Pre}; \cref{sec:pre-inference}) which is a baseline method that performs regular matrix factorization after propagating the logic rules in a deterministic manner, and \emph{joint optimization} (\textbf{Joint}; \cref{sec:method}) which maximizes an objective that combines terms from facts and first-order logic rules.
Additionally, we evaluate three baselines.
The \emph{matrix factorization} (\textbf{MF}; \cref{sec:mf}) model uses only ground atoms to learn relation and constant representations (\ie{} it has no access to any rules).
Furthermore, we consider pure \emph{symbolic logical inference} (\textbf{Inf}).
Since we restrict ourselves to a set of consistent, simple rules, this inference can be performed efficiently.
Our final approach, \emph{post-factorization inference} (\textbf{Post}), first runs matrix factorization and then performs logical inference on the known and predicted facts.
Post-inference is computationally expensive since for all premises of rules we have to iterate over \emph{all} rows (constant pairs) in the matrix to assess whether the premise is predicted to be true or not.
\maybe{Seb: this should be quite trivial once you index the predictions per relation, no?}

\subsection{Training Details}
\label{sec:learning}
Since we have no negative training facts, we follow \citet{riedel2013relation} by sampling unobserved ground atoms that we assume to be false. 
For rules, we use stochastic grounding as described in \cref{sec:grounding}.
Thus, in addition to a loss over the score of training facts, we have a loss over sampled unobserved ground atoms that we assume to be negative, as well as loss terms for ground rules. 
In other words, we learn symbol embeddings by minimizing \cref{eq:log_loss} where $\set{F}$ includes known and unobserved atoms, as well as ground propositional rules sampled using stochastic grounding.
In addition, we use $\ell_2$-regularization on all symbol representations.
For minimizing the training loss we use AdaGrad \citep{duchi2011adaptive}.

Remember that at test time, predicting a score for any unobserved ground atom $r_s(e_i,e_j)$ is done efficiently by calculating $\ldiff{r_s(e_i,e_j)}$. 
Note that this does not involve any explicit logical inference.
Instead, we expect the vector space of symbol embeddings to incorporate all given rules.

\paragraph{Hyperparameters}
Based on \cite{riedel2013relation}, we use $k = 100$ as the dimension for symbol representations in all models, $\lambda = 0.01$ as parameter for $\ell_2$-regularization, and $\alpha = 0.1$ as the initial learning rate for AdaGrad, which we run for $200$ epochs.\maybe{Seb: how determined?}

\paragraph{Runtime}
Each AdaGrad update is defined over a single cell of the matrix, and thus training data can be provided one ground atom at a time.\check{were we using batches? I don't think so!}
For matrix factorization, each AdaGrad epoch touches all the observed ground atoms once per epoch, and as many sampled negative ground atoms. 
With provided rules, it additionally revisits all the observed ground atoms that appear as an atom in the rules (and as many sampled negative ground atoms), and thus more general rules will be more expensive. 
However, the updates on ground atoms are performed independently and thus not all the data needs to be stored in memory. All presented models take less than 15 minutes to train on a 2.8 GHz Intel Core i7 machine.

\section{Results and Discussion}
\label{sec:log_results}

To asses the utility of injecting logic rules into symbol representations, we present a comparison on a variety of benchmarks.
First, we study the scenario of learning extractors for relations for which we do not have any Freebase alignments, \ie{}, the number of entity pairs that appear both in textual patterns and structured Freebase relations is zero.
This measures how well the different models are able to generalize only from logic rules and textual patterns (\cref{sec:results:zero}).
In \cref{sec:results:subsample}, we then describe an experiment where the number of Freebase alignments is varied in order to assess the effect of combining distant supervision and background knowledge on the accuracy of predictions.
Although the methods presented here target relations with insufficient alignments, we also provide a comparison on the complete distant supervision dataset in \cref{sec:results:complete}.

\subsection{Zero-shot Relation Learning}
\label{sec:results:zero}

\begin{figure}[t!]
  \centering
  \includegraphics[width=0.8\textwidth]{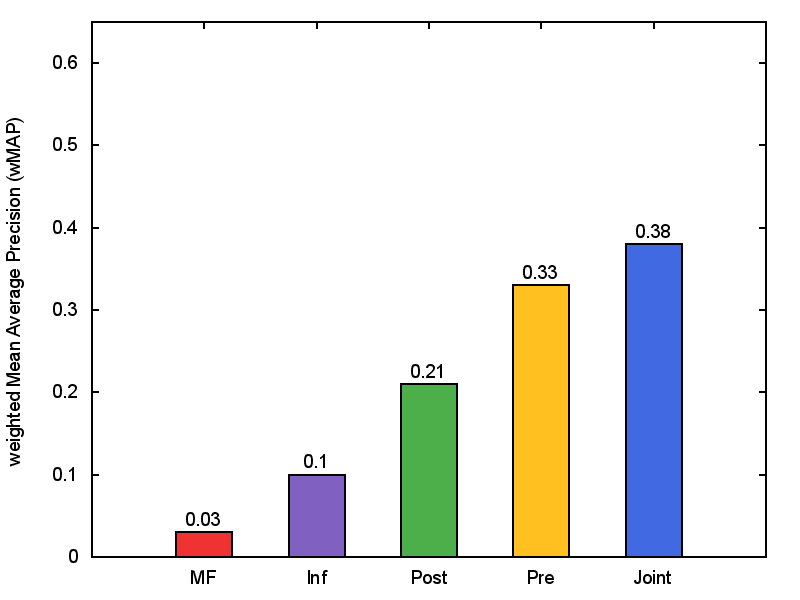}
  \caption{Weighted MAP scores for zero-shot relation learning.}
  \label{fig:zero_shot}
\end{figure}

\begin{table}[t!]
  \centering
\resizebox{\textwidth}{!}{
\begin{tabular}{ll|rrrrrr}
\toprule
              Relation & \# &                \textbf{MF} &  \textbf{Inf} & \textbf{Post} & \textbf{Pre} & \textbf{Joint}\\
\midrule
        \rel{person/company} &  102 &            0.07 & 0.03 & 0.15 & 0.31 & \bf{0.35} \\
  \rel{location/containedby} &   72 &            0.03 & 0.06 & 0.14 & 0.22 & \bf{0.31} \\
  \rel{author/works\_written} &   27 &           0.02 & 0.05 & 0.18 & \bf{0.31} & 0.27 \\
    \rel{person/nationality} &   25 &            0.01 & \em{0.19} & 0.09 & 0.15 & \em{0.19} \\
          \rel{parent/child} &   19 &            0.01 & 0.01 & 0.48 & 0.66 & \bf{0.75} \\
  \rel{person/place\_of\_birth} &   18 &         0.01 & 0.43 & 0.40 & 0.56 & \bf{0.59} \\
  \rel{person/place\_of\_death} &   18 &         0.01 & 0.24 & 0.23 & \bf{0.27} & 0.23 \\
  \rel{neighborhood/neighborhood\_of} &   11 &   0.00 & 0.00 & 0.60 & 0.63 & \bf{0.65} \\
        \rel{person/parents} &    6 &            0.00 & 0.17 & 0.19 & 0.37 & \bf{0.65} \\
      \rel{company/founders} &    4 &            0.00 & 0.25 & 0.13 & 0.37 & \bf{0.77} \\
     \rel{film/directed\_by} &    2 &            0.00 & 0.50 & 0.50 & 0.36 & \bf{0.51} \\
     \rel{film/produced\_by} &    1 &            0.00 & \em{1.00} & \em{1.00} & \em{1.00} & \em{1.00} \\
\midrule
                   \gls{MAP} &      &            0.01 & 0.23 & 0.34 & 0.43 & \bf{0.52} \\
          Weighted \gls{MAP} &      &            0.03 & 0.10 & 0.21 & 0.33 & \bf{0.38} \\
\bottomrule
\end{tabular}  
}
  \caption{(Weighted) MAP with relations that do not appear in any of the annotated rules omitted from the evaluation. 
  The difference between \textbf{Pre} and \textbf{Joint} is significant according to the sign-test ($p < 0.05$).}
  \label{tab:subsample-results}

\end{table}

We start with the scenario of learning extractors for relations that do not appear in the \gls{KB}, \ie{}, those that do not have any textual alignments.
\maybe{explain how logic injection allows for much stronger generalization (Pre and Joint vs rest)}
Such a scenario occurs in practice when a new relation is added to a \gls{KB} for which there are no facts that connect the new relation to existing relations or textual surface forms. 
For accurate extraction of such relations, we can only rely on background domain knowledge, \eg{}, in form of rules, to identify relevant textual alignments.
However, at the same time, there are correlations between textual patterns that can be utilized for improved generalization.
To simulate this setup, we remove all alignments between all entity pairs and Freebase relations from the distant supervision data, use the extracted logic rules (\cref{sec:setup:formulae}) as background knowledge, and assess the ability of the different methods to recover the lost alignments.

\Cref{fig:zero_shot} provides detailed results.
Unsurprisingly, matrix factorization (\textbf{MF}) performs poorly since predicate representations cannot be learned for the Freebase relations without any observed facts. 
The fact that we see a non-zero \gls{MAP} score for matrix factorization is due to random predictions.
Symbolic logical inference (\textbf{Inf}) is limited by the number of known facts that appear as a premise in one of the implications, and thus performs poorly too. 
Although post-factorization inference (\textbf{Post}) is able to achieve a large improvement over logical inference, explicitly injecting logic rules into symbol representations using pre-factorization inference (\textbf{Pre}) or joint optimization (\textbf{Joint}) gives superior results.
Finally, we observe that jointly optimizing the probability of facts and rules is able to best combine logic rules and textual patterns for accurate, zero-shot learning of relation extractors.

\Cref{tab:subsample-results} shows detailed results for each of the Freebase test relations. Except for \rel{author/works\_written} and \rel{person/place\_of\_death}, jointly optimizing the probability of facts and rules yields superior results.

\subsection{Relations with Few Distant Labels}
\label{sec:results:subsample}

\begin{figure}[t!]
  \centering

 \includegraphics[width=\textwidth]{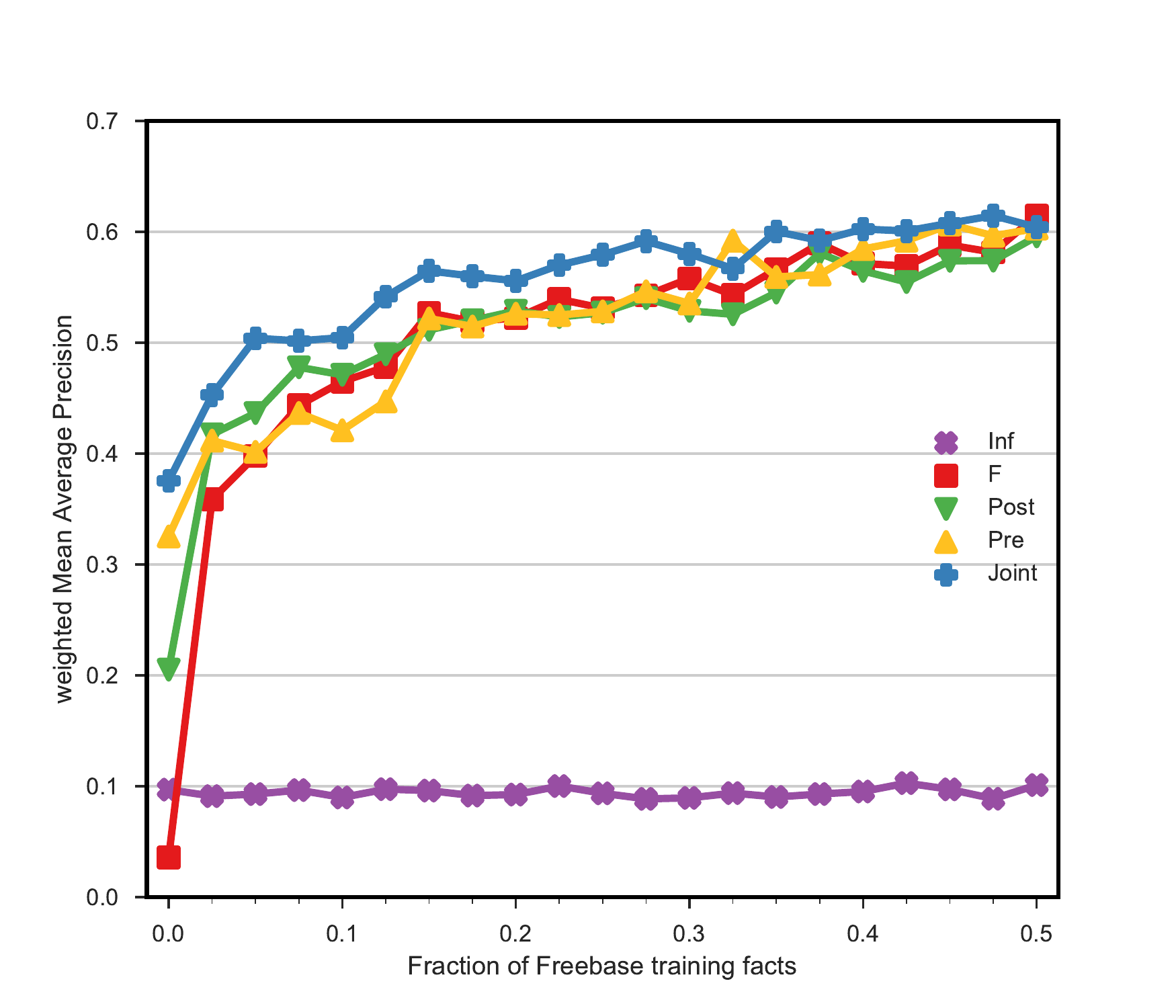}
  \caption{Weighted MAP of the various models as the fraction of Freebase training facts is varied. 
For $0\%$ Freebase training facts we get the zero-shot relation learning results presented in Table \ref{tab:subsample-results}.
  }
  \label{fig:subsample}

\end{figure}

In this section, we study the scenario of learning relations that have only a few distant supervision alignments, \ie{}, structured Freebase relations where only few textual patterns are observed for entity-pairs.
In particular, we observe the behavior of the various methods as the amount of distant supervision is varied.
We run all methods on training data that contains different fractions of Freebase training facts (and therefore different degrees of relation/text pattern alignment), but keep all textual patterns in addition to the set of annotated rules.

\Cref{fig:subsample} summarizes the results. 
The performance of symbolic logical inference does not depend on the amount of distant supervision data since it does not take advantage of the correlations in this data.
Matrix factorization does not make use of logical rules, and thus is the baseline performance when only using distant supervision.
For the factorization based methods, only a small fraction ($15\%$) of the training data is needed to achieve around $0.50$ weighted \gls{MAP} performance, thus demonstrating that they are efficiently exploiting correlations between predicates, and generalizing to unobserved facts.

Pre-factorization inference, however, does not outperform post-factorization inference and is on par with matrix factorization for most of the curve.
This suggests that it is not an effective way of injecting logic into symbol representations when ground atoms are also available.
In contrast, the joint model learns symbol representations that outperform all other methods in the $0$ to $30\%$ Freebase training data interval.
Beyond $30\%$, there seem to be sufficient Freebase facts for matrix factorization to encode these rules, thus yielding diminishing returns.

\subsection{Comparison on Complete Data}
\label{sec:results:complete}

\begin{figure}[t!]
  \centering
 \includegraphics[width=\textwidth]{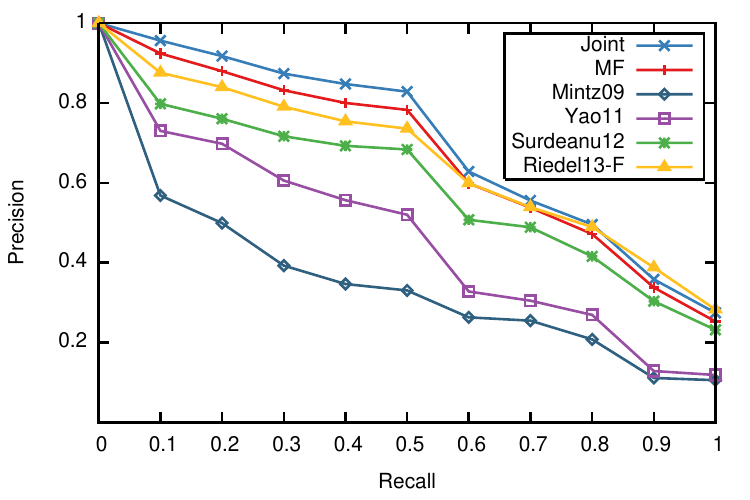}
  \caption{Precision-recall curve demonstrating that the \textbf{Joint} method, which incorporates annotated rules derived from the data, outperforms existing factorization approaches (\textbf{MF} and \textbf{Riedel13-F}).}
  \label{fig:complete}

\end{figure}

Although the focus of our work is on injecting logical rules for relations without sufficient alignments to the knowledge base, we also present an evaluation on the complete distant supervision data by \citet{riedel2013relation}.
Compared to \citeauthor{riedel2013relation}'s matrix factorization model \textbf{Riedel13-F}, our reimplementation (\textbf{MF}) achieves a lower wMAP ($64\%$ vs $68\%$) and a higher MAP ($66\%$ vs $64\%$).
We attribute this difference to the different loss function (\gls{BPR} vs. negative log-likelihood).
We show the precision-recall curve in \cref{fig:complete}, demonstrating that joint optimization provides benefits over the existing factorization and distant supervision techniques even on the complete dataset, and obtains $66\%$ weighted \gls{MAP} and $69\%$ \gls{MAP}, respectively. 
This improvement over the matrix factorization model can be explained by noting that the joint model reinforces high-quality annotated rules. 
\maybe{add overview table of results on complete dataset}

\section{Related Work}

\paragraph{Embeddings for Knowledge Base Completion}
Many methods for embedding predicates and constants (or pairs of constants) based on training facts for knowledge base completion have been proposed in the past (see \cref{sec:score}).
Our work goes further in that we learn embeddings that follow not only factual but also first-order logic knowledge.
Note that the method of regularizing symbol embeddings by rules described in this chapter are generally compatible with any existing neural link prediction model that provides per-atom scores between $0.0$ and $1.0$.
In our experiments we only worked with matrix factorization as neural link prediction model but based on our work \cite{guo2016jointly} were able to incorporate transitivity rules into TransE \citep{bordes2013translating} which models entities separately instead of learning a representation for every entity pair.

\paragraph{Logical Inference}
A common alternative, where adding first-order logic knowledge is trivial, is to perform symbolic logical inference~\citep{bos2005recognising,baader2007completing,bos2008semantics}.
However, such purely symbolic approaches cannot deal with the uncertainty inherent to natural language and generalize poorly.
\todo{need to justify this better}

\paragraph{Probabilistic Inference}
To ameliorate some of the drawbacks of symbolic logical inference, probabilistic logic based approaches have been proposed \citep{schoenmackers2008scaling,garrette2011integrating,beltagy2013montague,beltagy2014probabilistic}.
Since logical connections between relations are modeled explicitly, such approaches are generally hard to scale to large \glspl{KB}.
Specifically, approaches based on Markov Logic Networks (MLNs)~\citep{richardson2006markov} encode logical knowledge in dense, loopy graphical models, making structure learning, parameter estimation, and inference hard for the scale of our data.
In contrast, in our model the logical knowledge is captured directly in symbol representations, leading to efficient inference at test time as we only have to calculate the forward pass of a neural link prediction model.
Furthermore, as symbols are embedded in a low-dimensional vector space, we have a natural way of dealing with linguistic ambiguities and label errors that appear once OpenIE textual patterns are included as predicates for automated \gls{KB} completion \citep{riedel2013relation}.

Stochastic grounding is related to locally grounding a query in \gls{ProPPR} \citep{wang2013programming}. One difference is that we use stochastically grounded rules as differentiable terms in a representation learning training objective, whereas in \gls{ProPPR} such grounded rules are used for stochastic inference without learning symbol representations.

\paragraph{Weakly Supervised Learning}
Our work is also inspired by weakly supervised approaches~\citep{ganchev2010posterior} that use structural constraints as a source of indirect supervision.
These methods have been used for several NLP tasks~\citep{chang2007guiding,mann2008generalized,druck2009semi,singh2010minimally}.
The semi-supervised information extraction work by \cite{carlson2010coupled} is in spirit similar to our goal as they are using commonsense constraints to jointly train multiple information extractors.
A main difference is that we are learning symbol representations and allow for arbitrarily complex logical rules to be used as regularizers for these representations.

\paragraph{Combining Symbolic and Distributed Representations}
There have been a number of recent approaches that combine trainable subsymbolic representations with symbolic knowledge. 
\cite{grefenstette2013towards} describes an isomorphism between first-order logic and tensor calculus, using full-rank matrices to exactly \emph{memorize} facts. 
Based on this isomorphism, \cite{rocktaschel2014low} combine logic with matrix factorization for learning low-dimensional symbol embeddings that approximately satisfy given rules and generalize to unobserved facts on toy data. 
Our work extends this workshop paper by proposing a simpler formalism without tensor-based logical connectives, presenting results on a large real-world task, and demonstrating the utility of this approach for learning relations with no or few textual alignments.

\cite{chang2014typed} use Freebase entity types as hard constraints in a tensor factorization objective for universal schema relation extraction. 
In contrast, our approach is imposing soft constraints that are formulated as universally quantified first-order rules. 

\cite{de2013unsupervised} combine first-order logic knowledge
with a topic model to improve surface pattern clustering for relation extraction.
Since these rules only specify which relations can be clustered and which cannot, they do not capture the variety of dependencies embeddings can model, such as asymmetry. 
\cite{lewis2013combined} use distributed representations to cluster predicates before logical inference. 
Again, this approach is not as expressive as learning subsymbolic representations for predicates, as clustering does not deal with asymmetric logical relationships between predicates. 

Several studies have investigated the use of symbolic representations (such as dependency trees) to guide the composition of symbol representations~\citep{clark2007combining,mitchell2008vector,coecke2010mathematical,hermann2013role}.
Instead of guiding composition, we are using first-order logic rules as prior domain knowledge in form of regularizers to directly learn better symbol representations.

Combining symbolic information with neural networks has a long tradition.
\cite{towell1994knowledge} introduce Knowledge-Based Artificial Neural Networks whose topology is isomorphic to a \gls{KB} of facts and inference rules. 
There, facts are input units, intermediate conclusions hidden units, and final conclusions (inferred facts) output units.
Unlike in our work, there are no learned symbol representations.
\cite{holldobler1999approximating} and \cite{hitzler2004logic} prove that for every logic program there exists a recurrent neural network that approximates the semantics of that program.
This is a theoretical insight that unfortunately does not provide a way of constructing such a neural network.
Recently, \cite{bowman2013can} demonstrated that \glspl{NTN} \citep{socher2013reasoning} can accurately learn natural logic reasoning.

The method presented in this chapter is also related to the recently introduced \glspl{EqNet} \citep{allamanis2016learning}. 
\glspl{EqNet} recursively construct neural representations of symbolic expressions to learn about equivalence classes.
In our approach, we recursively construct neural networks for evaluating Boolean expressions and use them as regularizers to learn better symbol representations for automated \gls{KB} completion.

\section{Summary}
In this chapter, we introduced a method for mapping symbolic first-order logic rules to differentiable terms that can be used to regularize symbol representations learned by neural link prediction models for automated \gls{KB} completion.
Specifically, we proposed a joint training objective that maximizes the probability of known training facts as well as propositional rules that we made continuous by replacing logical operators with differentiable functions.
While this is inspired by Fuzzy Logic, our contribution is backpropagating a gradient from a negative log-likelihood loss through the propositional rule and a neural link prediction model that scores ground atoms to calculate a gradient with respect to vector representations of symbols. Subsequently, we update these representations using gradient descent, thereby encoding the ground truth of a propositional rule directly in the vector representations of symbols.
This leads to efficient predictions at test time as we only have to calculate the forward pass of the neural link prediction model.
We described a stochastic grounding process for incorporating first-order logic rules.
Our experiments for automated \gls{KB} completion show that the proposed method can be used to learn extractors for relations with little to no observed textual alignments, while at the same time benefiting from correlations between textual surface form patterns.

\chapter{Lifted Regularization of Predicate Representations by Implications}
\glsresetall

\label{foil}

The method for incorporating first-order logic rules into symbol representations introduced in the previous chapter relies on stochastic grounding. Moreover, not only vector representations of predicates but also representations of pairs of constants are optimized to maximize the probability of provided rules.
This is problematic for the following reasons.
\paragraph{Scalability}
Even with stochastic grounding, incorporating first-order logic rules with the method described so far is dependent on the size of the domain of constants. As an example take the simple rule $\rel{isMortal}(\var{X}) \lif \rel{isHuman}(\var{X})$ and assume we observe $\rel{isHuman}$ for seven billion constants. 
Only for this single rule, we would already add seven billion loss terms to the training objective in \cref{eq:log_loss}.
\paragraph{Generalizability}
Since we backpropagate upstream gradients of a rule not only into predicate representations but also into representations of pairs of constants, there is no theoretical guarantee that the rule will indeed hold for constant pairs not observed during training.
\paragraph{Flexibility of Training Loss}
The previous method is not compatible with rank-based losses such as \gls{BPR}. 
Instead, we had to use the negative log-likelihood loss, which results in lower performance compared to \gls{BPR} for automated \gls{KB} completion.
\paragraph{Independence Assumption}
As explained in \cref{sec:prop}, we had to assume that the probability of ground atoms is conditionally independent given symbol representations.
This assumption is already violated for the simple Boolean expression $\ldiff{\lrule{f}\land\lrule{f}}$ with $0 < \ldiff{\lrule{f}} < 1$, which results in $\ldiff{\lrule{f}\land\lrule{f}} = \ldiff{\lrule{f}}\ldiff{\lrule{f}} < \ldiff{\lrule{f}}$.

Ideally, we would like to have a way to incorporate first-order logic rules into symbol representations in a way that (i) is independent of the size of the domain of constants, (ii) generalizes to unseen constant pairs, (iii) can be used with a broader class of training objectives, and (iv) does not assume that the probability of ground atoms is conditionally independent given symbol representations.

In this chapter, we present a method that satisfies these desiderata, but only for simple implication rules instead of general function-free first-order logic rules, and for matrix factorization as neural link prediction model. 
However, we note that such simple implications are commonly used and can improve automated \gls{KB} completion. 
The method we propose incorporates implications into vector representations of predicates while maintaining the computational efficiency of only modeling training facts.
This is achieved by enforcing a partial order of predicate representations in the vector space, which is entirely independent of the number of constants in the domain.
It only involves representations of the predicates that are mentioned in rules, as well as a general rule-independent constraint on the embedding space of constant pairs. 
In the example given above, if we require that every component of the vector representation $\ldiff{\rel{isHuman}}$ is smaller or equal to the corresponding component of the predicate representation $\ldiff{\rel{isMortal}}$, then we can show that the implication holds for any \emph{non-negative} representation of a constant.
Hence, our method avoids the need for separate loss terms for every ground atom resulting from grounding rules. 
In statistical relational learning, this type of approach is often referred to as \emph{lifted} inference or learning~\citep{poole2003first,Braz:2007:LFP:1369181} because it deals with groups of random variables at a first-order level. 
In this sense, our approach is a lifted form of rule injection. 
This allows us to impose a large number of rules while learning distributed representations of predicates and constant pairs.
Furthermore, once these constraints are satisfied, the injected rules always hold, even for unseen but inferred ground atoms.
In addition, it does not rely on the assumption of conditional independence of ground atoms.

\section{Method}
\label{sec:foil_model}

\begin{figure}[t!]
  \centering
  \includegraphics[]{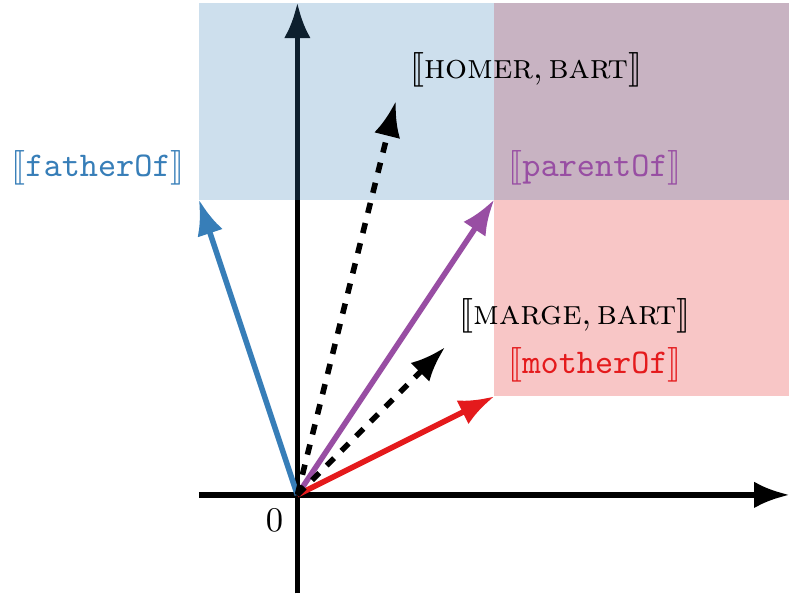}
  \caption{Example for implications that are directly captured in a vector space.}
  \label{fig:foil_overview}
\end{figure}

We want to incorporate implications of the form $\forall\var{X},\var{Y}: \rel{h}(\var{X},\var{Y}) \lif \rel{b}(\var{X}, \var{Y})$ and we consider matrix factorization as the neural link prediction model for scoring atoms, \ie{}, $\ldiff{r(e_i,e_j)} = \sigmoid({\ldiff{r}^\top\ldiff{e_i,e_j}})$ (\cref{sec:mf}).
A necessary condition for the implication to hold is that for every possible assignment of constants $e_i$ to $\var{X}$ and $e_j$ to $\var{Y}$, the score of $\ldiff{\rel{h}(e_i,e_j)}$ needs to be as least as large as $\ldiff{\rel{b}(e_i,e_j)}$. 
In the discrete case this means that if $\ldiff{\rel{b}(e_i,e_j)}$ is $1.0$ (True) within some small epsilon, $\ldiff{\rel{h}(e_i,e_j)}$ needs to be $1.0$ too, but not vice versa.
As sigmoid is a monotonic function, we can rewrite $\forall\var{X},\var{Y}: \rel{h}(\var{X},\var{Y}) \lif \rel{b}(\var{X}, \var{Y})$ in terms of grounded rules as the following condition:
\begin{equation}
  \label{eq:8}
   \forall (e_i,e_j)\in\set{C}: \ldiff{\rel{h}}^\top\ldiff{e_i,e_j} \geq \ldiff{\rel{b}}^\top\ldiff{e_i,e_j}.
\end{equation}
Ordering symbol representations in a vector space such that implications are directly captured is inspired by Order Embeddings \citep{vendrov2016order}.
An example where \cref{eq:8} holds is illustrated in \cref{fig:foil_overview}.
Here, $\rel{fatherOf}$ and $\rel{motherOf}$ both imply $\rel{parentOf}$, since every component of $\ldiff{\rel{parentOf}}$ is larger than the corresponding component in $\ldiff{\rel{fatherOf}}$ or $\ldiff{\rel{motherOf}}$.
Predicate representations in the blue, red and purple area are implied by $\rel{fatherOf}$, $\rel{motherOf}$ and $\rel{parentOf}$, respectively.
Both constant pair representations, $\ldiff{\const{homer},\const{bart}}$ and $\ldiff{\const{marge},\const{bart}}$, are non-negative.
Thus, for any score of $\ldiff{\rel{fatherOf}(\const{homer},\const{bart})}$, the score of $\ldiff{\rel{parentOf}(\const{homer},\const{bart})}$ is larger, but not vice versa.
Note that this holds for any representation of constant pairs $\ldiff{e_i,e_j}$ as long as it is non-negative (\ie{} placed in the upper-right quadrant).

The main insight is that we can make this condition independent of $e_i,e_j$ if we make sure that $\ldiff{e_i,e_j} \in \R_+^k$, \ie{}, all constant representations are non-negative.
Thus, \cref{eq:8} becomes
\begin{equation}
  \label{eq:16}
  \ldiff{\rel{h}} \geq \ldiff{\rel{b}}, \quad\forall (e_i,e_j)\in\set{C}: \ldiff{e_i,e_j} \in \R_+^k
\end{equation}
where $\geq$ is the component-wise comparison.
In other words, for ensuring that $\rel{b}$ implies $\rel{h}$ for any pair of constants, we only need one relation-specific loss term that makes sure all components in $\ldiff{\rel{h}}$ are at least as large as in $\ldiff{\rel{b}}$, and one general restriction on the representations of constant pairs.

\paragraph{Non-negative Representation of Constant Pairs}
There are many choices for ensuring all constant pair representations are positive.
One option is to initialize constant pair representations to non-negative vectors and projecting gradient updates to make sure they stay non-negative.
Another option is to apply a transformation $f: \R^k \to \R_+^k$ before constant pair representations are used in a neural link prediction model for scoring atoms.
For instance, we could use $\exp(x) = e^x$ or $\relu(x) = \max(0, x)$ for $f$.
However, we choose to restrict constant representations even more than required, and decided to use a transformation to approximately Boolean embeddings \citep{kruszewski2015boolean}.
For every constant pair representation $\ldiff{e_i,e_j} \in \R^k$, we obtain a non-negative representation $\ldiff{e_i,e_j}_+ \in [0,1]^k$ by applying the element-wise sigmoid function.\todo{motivate approximately Boolean representations are a good idea}
Thus, the matrix factorization objective with \gls{BPR} for facts in \cref{eq:mf_loss} becomes the following approximate loss 
\begin{equation}
  \label{eq:17}
  \globalloss = \sum_{\substack{r_s(e_i,e_j)\ \in\ \set{O},\\(e_m,e_n)\ \sim\ \set{C},\\r_s(e_m,e_n)\ \not\in\ \set{O}}} - w_r \log\sigma(\vec{v}_r^\top\sigma(\vec{v}_{ij}) - \vec{v}_r^\top\sigma(\vec{v}_{mn})) + \lambda_p\|\vec{v}_s\|_2^2+\lambda_c(\|\vec{v}_{ij}\|_2^2+\|\vec{v}_{mn}\|_2^2)
\end{equation}
where we sample constant pairs as in \cref{sec:bpr}.
We denote the extension with sigmoidal constant pair representations of the matrix factorization model (\textbf{F}) by \citep{riedel2013relation} as \textbf{FS} (S = sigmoidal).

\paragraph{Implication Loss}
There are various ways for modeling $\ldiff{h} \geq \ldiff{b}$ to incorporate the implication $\rel{h} \lif \rel{b}$.
Here, we propose to use a hinge-like loss
\begin{equation}
\label{eq:foil}
\loss(\ldiff{\rel{h} \lif \rel{b}}) = \sum_i^k\max(0, \ldiff{\rel{b}}_i-\ldiff{\rel{h}}_i+\epsilon)
\end{equation}
where $\epsilon$ is a small positive margin to ensure that the gradient does not disappear before the inequality is actually satisfied. 
A nice property of this loss compared to the method presented in the previous chapter is that once the implication holds (\ie{} $\ldiff{\rel{h}}$ is larger than $\ldiff{\rel{b}}$), the gradient is zero and both predicate representations are not further updated for this rule.
For every given implication rule, we add the corresponding loss term to the fact loss in \cref{eq:17}.
The global approximate loss over facts and rules in a set $\set{R}$ is thus
\begin{equation}
  \label{eq:7}
  \globalloss' = \globalloss + \sum_{\rel{h} \lif \rel{b}\ \in\ \set{R}} \loss(\ldiff{\rel{h} \lif \rel{b}}) 
\end{equation} and we denote the resulting model as \textbf{FSL} (L = logic).
Furthermore, we use a margin of $\epsilon=0.01$ in all experiments.
As in \cref{log}, we predict the probability of an atom $r_s(e_i,e_j)$ at test time via $\ldiff{r_s(e_i,e_j)}$, which is efficient as there is no logical inference.

\section{Experiments}
\label{sec:foil_experiments}
We follow the experimental setup from the previous chapter (\cref{sec:exps}) and evaluate on the NYT corpus \citep{riedel2013relation}.
Again, we test how well the presented models can incorporate rules when there is no alignment between textual surface forms and Freebase relations (Zero-shot Relation Learning), and when the number of Freebase training facts is increased (Relations with Few Distant Labels).
In addition, we experiment with rules automatically extracted from WordNet~\citep{miller1995wordnet} to improve automated \gls{KB} completion on the full dataset.

\paragraph{Incorporating Background Knowledge from WordNet}
We use WordNet hypernyms to generate rules for the NYT dataset. 
To this end, we iterate over all surface form patterns in the dataset and attempt to replace words in the pattern by their hypernyms. 
If the resulting surface form is contained in the dataset, we generate the corresponding rule. 
For instance, we generate a rule $\rel{\#1-official-\#2}(\var{X}, \var{Y}) \lif \rel{\#1-diplomat-\#2}(\var{X}, \var{Y})$ since both patterns are contained in the dataset and we know from WordNet that $\rel{official}$ is a hypernym of $\rel{diplomat}$. 
This resulted in $427$ generated rules that we subsequently annotated manually, yielding $36$ high-quality rules listed in \cref{appendixB}.
Note that all of these rules are between surface form patterns. 
Thus, none of these rules has a Freebase relation as the head predicate. 
Although the test relations all originate from Freebase, we still hope to see improvements by transitive effects, such as better surface form representations that in turn help to predict Freebase facts. \maybe{be more precise: we match WordNet rules only if the two patterns also have correct dependency edges (not mentioned here so far for brevity)}

\subsection{Training Details}
All models were implemented in TensorFlow~\citep{abadi2015tensorflow}.
We use the hyperparameters of \cite{riedel2013relation}, with $k=100$ as the size of symbol representations and a weight of $0.01$ for the $\ell_2$ regularization (\cref{eq:17}).  
We use ADAM \citep{kingma2014adam} for optimization with an initial learning rate of $0.005$ and a batch size of $8192$. 
The embeddings are initialized by sampling $k$ values uniformly from $[-0.1,0.1]$.
\maybe{ we use $\tilde{\beta}=0.1$ for the implication loss throughout our experiments???}

\section{Results and Discussion}\label{sec:foil_results}

Before turning to the injection of rules, we compare model \textbf{F} with model \textbf{FS}, and show that restricting the constant embedding space has a regularization effect rather than limiting the expressiveness of the model (\cref{subsec:Frestricted}).
We then demonstrate that model \textbf{FSL} is capable of zero-shot learning (\cref{subsec:zeroshot}), that it can take advantage of alignments between textual surface forms and Freebase relations alongside rules (\cref{subsec:few}), and we show that injecting high-quality WordNet rules leads to improved predictions on the full dataset (\cref{subsec:injectWordNet}). 
Lastly, we provide details on the computational efficiency of the lifted rule injection method (\cref{subsec:time}) and demonstrate that it correctly captures the asymmetry of implication rules (\cref{subsec:asymm}).

\begin{table}[t!]
\centering
\resizebox{1.0\textwidth}{!}{
    \begin{tabular}{ l l | c c c c c }
    \toprule
                  Test relation & \# & \textbf{Riedel13-F} & \textbf{F} & \textbf{FS} & \textbf{FSL} \\
    \midrule
            \rel{person/company} &  106 &     0.75 &   0.73 &   0.74 & {\bf 0.77} \\
       \rel{location/containedby} &   73 &     0.69 &   0.62 &   0.70 & {\bf 0.71} \\
       \rel{person/nationality}   &   28 &     0.19 &   0.20 &   0.20 & {\bf 0.21} \\
      \rel{author/works\_written} &   27 &     0.65 &   {\bf   0.71} &   0.69 &   0.65 \\
       \rel{person/place\_of\_birth} &   21 &  0.72 &  0.69 & {\bf   0.72} &   0.70 \\
               \rel{parent/child} &   19 &   0.76 &   0.77 &   0.81 & {\bf   0.85} \\
       \rel{person/place\_of\_death} &   19 &  0.83 &  {\em 0.85} &   0.83 &   {\em 0.85} \\
       \rel{neighborhood/neighborhood\_of} &   11 &   {\bf 0.70} & 0.67 &   0.63 &   0.62 \\
        \rel{person/parents} &    6 &   0.61 & 0.53 &   {\em 0.66} &   {\em 0.66} \\
        \rel{company/founders} &    4 &   {\bf 0.77} & 0.73 &   0.64 &   0.67 \\
        \rel{sports\_team/league} &    4 &  {\bf 0.59} &  0.44 &   0.43 &   0.56 \\
       \rel{team\_owner/teams\_owned} &    2 &   0.38 &  {\em 0.64} &   {\em 0.64} &   0.61 \\
       \rel{team/arena\_stadium} &    2 &  {\em 0.13} &   {\em 0.13} &   {\em 0.13} &   0.12 \\
        \rel{film/directed\_by} &    2 &  {\bf 0.50} &   0.18 &   0.17 &   0.13 \\
       \rel{broadcast/area\_served} &    2 & 0.58 & 0.83 &   0.83 & {\bf   1.00} \\
       \rel{structure/architect} &    2 & {\em   1.00} & {\em   1.00} & {\em   1.00} & {\em   1.00} \\
       \rel{composer/compositions} &    2 &  {\bf 0.67} &  0.64 &   0.51 &   0.50 \\
        \rel{person/religion} &    1 &  {\em   1.00} & {\em   1.00} & {\em   1.00} & {\em   1.00} \\
       \rel{film/produced\_by} &    1 &  0.50 & {\em   1.00} & {\em   1.00} &   0.33 \\
    \midrule
              Weighted MAP &      &   0.67 &  0.65 &   0.67 &   {\bf 0.69} \\
    \bottomrule          
    \end{tabular}
}
\caption{Weighted MAP for our reimplementation of the matrix factorization model (\textbf{F}), compared to restricting the constant embedding space (\textbf{FS}) and to injecting WordNet rules (\textbf{FSL}). The orginial matrix factorization model by \cite{riedel2013relation} is denoted as \textbf{Riedel13-F}.}
\label{tab:foil_results}
\end{table}

\subsection{Restricted Embedding Space for Constants}\label{subsec:Frestricted}
Before incorporating external commonsense knowledge into relation representations, we were curious about how much we lose by restricting the embedding space of constant symbols to approximately Boolean embeddings.
Surprisingly, we find that the expressiveness of the model does not suffer from this strong restriction. 
From Table~\ref{tab:foil_results} we see that restricting the constant embedding space (\textbf{FS}) yields a $2$ percentage points higher weighted \gls{MAP}  compared to a real-valued constant embedding space (\textbf{F}).
This result suggests that the restriction has a regularization effect that improves generalization. 
We also provide the original results for the matrix factorization model by \newcite{riedel2013relation} denoted as \textbf{Riedel13-F} for comparison. 
Due to a different implementation and optimization procedure, the results for our model \textbf{F} compared to \textbf{Riedel13-F} are slightly worse ($65\%$ vs $67\%$ wMAP).

\subsection{Zero-shot Relation Learning}\label{subsec:zeroshot}
In \cref{sec:results:zero}, we observed that by injecting implications where the head is a Freebase relation for which no training facts are available, we can infer Freebase facts based on rules and correlations between textual surface patterns. 
Here, we repeat this experiment.
The lifted rule injection model (\textbf{FSL}) reaches a weighted \gls{MAP} of $35\%$, comparable to the $38\%$ of the method presented in the last chapter. 
For this experiment, we initialized the predicate representations of Freebase relations implied by the rules with negative random vectors sampled uniformly from $[-7.9, -8.1]$. 
The reason is that without any negative training facts for these relations, their components can only go up due to the lifted implication loss.
Consequently, starting with high values before optimization would impede the freedom with which these representations can be ordered in the embedding space.  
This demonstrates that while \textbf{FSL} performs a bit worse than the \textbf{Joint} model in \cref{log}, it can still be used for zero-shot relation learning.

\subsection{Relations with Few Distant Labels}
\label{subsec:few}

\begin{figure}[t!]
  \centering
  \includegraphics[width=1\textwidth]{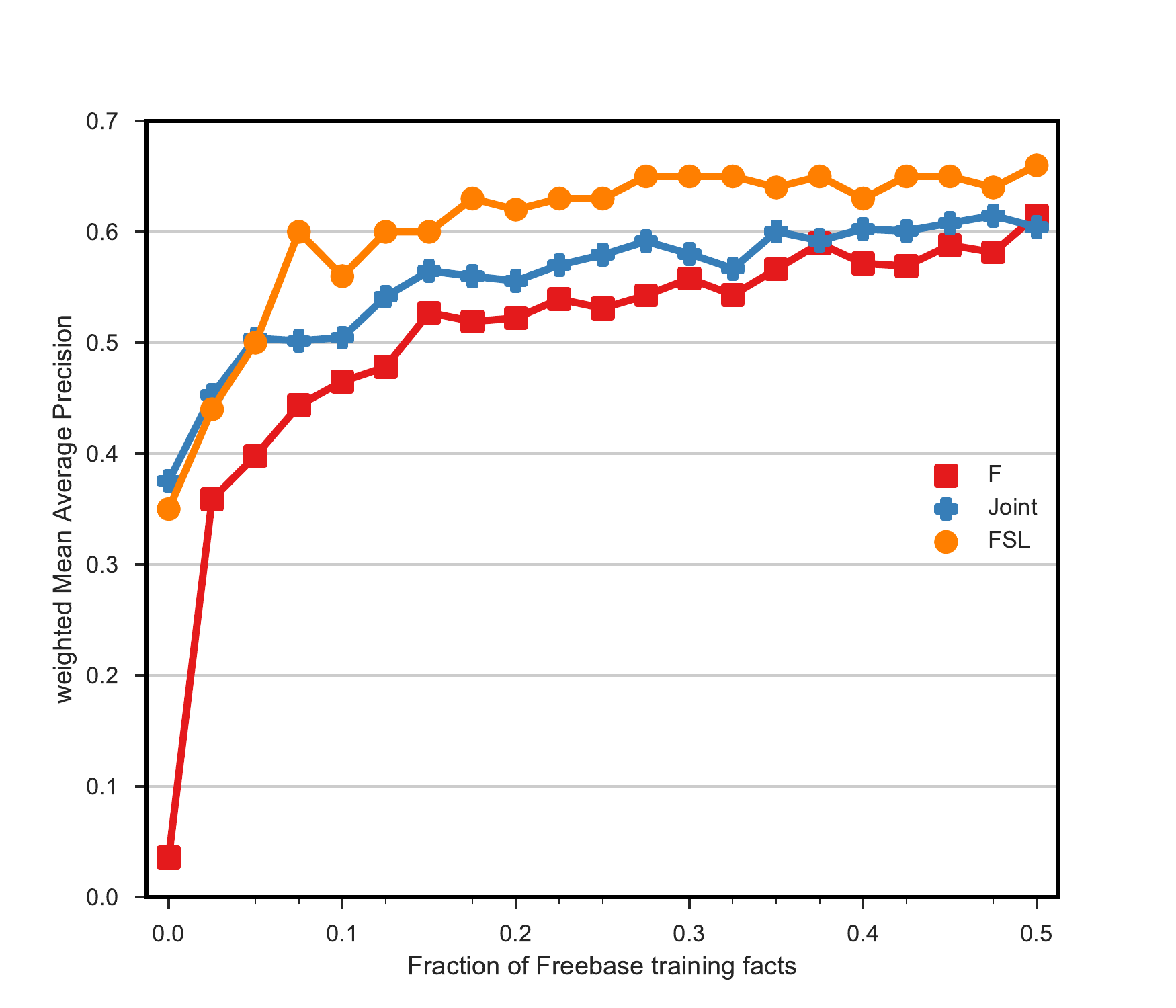}
  \caption{Weighted MAP for injecting rules as a function of the fraction of Freebase facts.}
  \label{fig:zeroshot}
\end{figure}
\Cref{fig:zeroshot} shows how the relation extraction performance improves when more Freebase facts are added.
As in the last chapter, it measures how well the proposed models, matrix factorization (\textbf{F}), propositionalized rule injection (\textbf{Joint}), and our lifted rule injection model (\textbf{FSL}), can make use of provided implication rules, as well as correlations between textual surface form patterns and increasing numbers of Freebase facts.
Although \textbf{FSL} starts at a lower performance than \textbf{Joint} when no Freebase training facts are present, it outperforms \textbf{Joint} and a plain matrix factorization model by a substantial margin when provided with more than $7.5\%$ of Freebase facts.
This indicates that, in addition to being much faster than \textbf{Joint}, it can make better use of provided rules and few training facts. 
We attribute this to being able to use \gls{BPR} as loss for ground atoms and the regularization effect of restricting the embedding space of constants pairs.
The former is compatible with our rule-injection method, but not with the approach of maximizing the expectation of propositional rules presented in the previous chapter.

\subsection{Incorporating Background Knowledge from WordNet}\label{subsec:injectWordNet}
In column \textbf{FSL} in \cref{tab:foil_results}, we show results obtained by injecting  WordNet rules. 
Compared to \textbf{FS}, we obtain an increase of weighted \gls{MAP} by 2\%, as well as 4\% compared to our reimplementation of the matrix factorization model \textbf{F}.
This demonstrates that imposing a partial order based on implication rules can be used to incorporate logical commonsense knowledge and increase the quality of information extraction and automated \gls{KB} completion systems.
Note that our evaluation setting guarantees that only indirect effects of the rules are measured, \ie{}, we do not use any rules directly implying Freebase test relations.
Consequently, our increase of prediction performance is due to an improved predicate embedding space beyond those predicates that are explicitly stated in provided rules. 
For example, injecting the rule $\rel{\#1-parent-\#2}(\var{X}, \var{Y}) \lif \rel{\#1-father-\#2}(\var{X}, \var{Y})$ can contribute to improved predictions for the Freebase test relation $\rel{parent/child}$ via co-occurring entity pairs between $\rel{\#1-parent-\#2}(\var{X}, \var{Y})$ and $\rel{parent/child}$.

\subsection{Computational Efficiency of Lifted Rule Injection}\label{subsec:time}
In order to assess the computational efficiency of the proposed method, we measure the time needed per training epoch when using a single 2.4GHz CPU core. 
We measure on average $6.33s$ per epoch when not using rules (model \textbf{FS}), against $6.76s$ and $6.97s$ when using $36$ filtered and $427$ unfiltered rules (model \textbf{FSL}), respectively.
Increasing the number of rules from $36$ to $427$ leads to an increase of only 3\% in computation time. 
Furthermore, using $427$ rules only adds an overhead of $~10\%$ to the computation time needed for learning ground atoms.
This demonstrates that lifted rule injection scales well with the number of rules.

\subsection{Asymmetry}\label{subsec:asymm}

\begin{table}[t!]
\centering
\resizebox{1.0\textwidth}{!}{
\begin{tabular}{ l c l | c c | c c}
\toprule
& Rule &  & \multicolumn{2}{c|}{\textbf{FS}} & \multicolumn{2}{c}{\textbf{FSL}}\\
\multicolumn{1}{c}{$\rel{head}$} & $\lif$ & \multicolumn{1}{c|}{$\rel{body}$} & $\ldiff{\rel{head}(e_i,e_j)}$ & $\ldiff{\rel{body}(\hat e_i, \hat e_j)}$ & $\ldiff{\rel{head}(e_i,e_j)}$ & $\ldiff{\rel{body}(\hat e_i, \hat e_j)}$ \\
\midrule
\rel{\#1-organization-\#2} & $\lif$ & \rel{\#1-party-\#2}  & 0.70  & 0.86  &  0.99  &  0.22  \\
\rel{\#1-parent-\#2}  & $\lif$ & \rel{\#1-father-\#2}  & 0.72 & 0.89 &  0.96  &  0.00   \\
\rel{\#1-lawyer-\#2}  & $\lif$ & \rel{\#1-prosecutor-\#2}   & 0.87 & 0.80 &  0.99  &  0.01  \\
\rel{\#1-newspaper-\#2}  & $\lif$ & \rel{\#1-daily-\#2} & 0.90 & 0.86  &  0.98  &  0.79  \\
\rel{\#1-diplomat-\#2}  & $\lif$ & \rel{\#1-ambassador-\#2} & 0.93 & 0.84  &  0.31  &  0.05   \\
\midrule
\multicolumn{3}{c|}{Average over all rules}   & 0.74 & 0.70  &  0.95  &  0.28   \\
\bottomrule          
\end{tabular}
}
\caption{Average score of facts with constants that appear in the body of facts ($e_i,e_j$) or in the head ($\hat e_i, \hat e_j$) of a rule.}
\label{tab:asymmetry}
\end{table}

One concern with incorporating implications into a vector space is that the vector representation of the predicates in the head and body are simply moving closer together.
This would violate the asymmetry of implications. 
In the experiments above we might not observe that this is a problem as we are only testing how well the model predicts facts for Freebase relations and not how well we can predict textual surface form patterns.
Thus, we perform the following experiment.
After incorporating WordNet rules of the form $\rel{head} \lif \rel{body}$, we select all constant pairs $(e_i,e_j)$ for which we observe $\rel{body}(e_i,e_j)$ in the training set. 
If the implication holds, $\ldiff{\rel{head}(e_i,e_j)}$ should yield a high score.
If we conversely select $(\hat e_i,\hat e_j)$ based on known facts $\rel{head}(\hat e_i, \hat e_j)$ and assume that $\rel{head}$ and $\rel{body}$ are not equivalent, then we expect a lower score for $\ldiff{\rel{body}(\hat e_i, \hat e_j)}$ than for $\ldiff{\rel{head}(\hat e_i, \hat e_j)}$ as $\rel{body}(\hat e_i, \hat e_j)$ might not be true.

\Cref{tab:asymmetry} lists these scores for five sampled WordNet rules, and the average over all WordNet rules when injecting these rules (model \textbf{FSL}) or not (model \textbf{FS}).
From this table, we can see that the score of the head atom is on average much higher than the score for the body atom once we incorporate the rule in the vector space (\textbf{FSL}).
In contrast, if we only run matrix factorization with the restricted constant embedding space, we often see high predictions for both, the body and head atom.
This suggests that matrix factorization merely captures similarity between predicates.
In contrast, by injecting implications we trained predicate representations that indeed yield asymmetric predictions. 
Given a high score for a \rel{body} ground atom, the model predicts a high score for the \rel{head}, but not vice versa.
The reason that we also get a high score for \rel{body} ground atoms in the fourth rule is that \rel{newspaper} and \rel{daily} are synonymously used in training texts.

\section{Related Work}\label{sec:relatedwork}
Recent research on combining rules with learned vector  representations has been important for new developments in the field of automated \gls{KB} completion.  
\maybe{most of this can go the related work section of LOG?}
\cite{wang2015knowledge} demonstrated how different types of rules can be incorporated using an Integer Linear Programming approach.  
\cite{wang2016learning} learned embeddings for facts and first-order logic rules using matrix factorization.
Yet, all of these approaches, and the method presented in the previous chapter, ground first-order rules with constants in the domain.
This limits their scalability towards large rule sets and \glspl{KB} with large domains of constants. 
It formed an important motivation for our lifted rule injection model, which by construction does not suffer from that limitation. 
\cite{wei2015large} proposed an alternative strategy to tackle the scalability problem by reasoning on a filtered subset of ground atoms.

\cite{wu2015structured} proposed to use the \gls{PRA} (\cref{sec:path}) for capturing long-range interactions between entities in conjunction with modeling pairwise relations.\info{I don't really get this reference} 
Our model differs substantially from their approach, in that we consider pairs of constants instead of separate constants, and that we inject a provided set of rules. 
Yet, by creating a partial order in the relation embeddings as a result of injecting implication rules, model \textbf{FSL} can also capture interactions beyond the predicates directly mentioned in these rules, which we demonstrated in \cref{subsec:injectWordNet} by injecting rules between surface patterns and measuring an improvement on predictions for structured Freebase relations.

Combining logic and distributed representations is also an active field of research outside of automated \gls{KB} completion. 
Recent advances include the work by \cite{faruqui2015retrofitting}, who injected ontological knowledge from WordNet into word embeddings to improve performance on downstream \acrshort{NLP} tasks. 
Furthermore, \cite{vendrov2016order} proposed to enforce a partial order in an embeddings space of images and phrases. 
Our method is related to such Order Embeddings since we define a partial order on relation embeddings. 
We extend this work to automated \gls{KB} completion where we ensure that implications hold for all pairs of constants by introducing a restriction on the embedding space of constant pairs.
Another important contribution is the recent work by \cite{hu2016harnessing}, who proposed a framework for injecting rules into general neural network architectures by jointly training on target outputs and on rule-regularized predictions provided by a so-called teacher network. 
Although quite different at first sight, their work could offer a way to use our model in various neural network architectures by integrating the proposed lifted loss into the teacher network.

\section{Summary}
\label{sec:conclusion}

We presented a fast approach for incorporating first-order implication rules into distributed representations of predicates for automated \gls{KB} completion. 
We termed our approach \emph{lifted rule injection}, as the main contribution over the previous chapter is the fact that it avoids the costly grounding of first-order implication rules and is thus independent of the size of the domain of constants.
By construction, these rules are satisfied for any observed or unobserved ground atom.
The presented approach requires a restriction on the embedding space of constant pairs. 
However, experiments on a large-scale real-world \gls{KB} show that this does not impair the expressiveness of the learned representations. 
On the contrary, it appears to have a beneficial regularization effect. 

By incorporating rules generated from WordNet hypernyms, our model improved over a matrix factorization baseline for \gls{KB} completion. 
Especially for domains where annotation is costly and only small amounts of training facts are available, our approach provides a way to leverage external knowledge sources efficiently for inferring facts. 

On the downside, the lifted rule injection method presented here is only applicable for implication rules and when using matrix factorization as the neural link prediction model. 
Furthermore, it is unclear how far regularizing predicate representations can be pushed without constraining the embedding space too much.
Specifically, it is unclear how more complex rules such as transitivity can be incorporated in a lifted way. 
Hence, we are exploring a more direct synthesis of representation learning and first-order logic inference in the next chapter.

\chapter{End-to-End Differentiable Proving}
\label{ntp} 
\glsresetall

Current state-of-the-art methods for automated \gls{KB} completion use neural link prediction models to learn distributed vector representations of symbols (\ie{} subsymbolic representations) for scoring atoms~\citep{nickel2012factorizing,riedel2013relation,socher2013reasoning,chang2014typed,yang2014embedding,toutanova2015representing,trouillon2016complex}. 
Such subsymbolic representations enable these models to generalize to unseen facts by encoding similarities: If the vector of the predicate symbol $\rel{grandfatherOf}$ is similar to the vector of the symbol $\rel{grandpaOf}$, both predicates likely express a similar relation. 
Likewise, if the vector of the constant symbol $\const{lisa}$ is similar to $\const{maggie}$, similar relations likely hold for both constants (\eg{} they live in the same city, have the same parents etc.).

This simple form of reasoning based on similarities is remarkably effective for automatically completing large \glspl{KB}. 
However, in practice it is often important to capture more complex reasoning patterns that involve several inference steps. 
For example, if $\const{abe}$ is the father of $\const{homer}$ and $\const{homer}$ is a parent of $\const{bart}$, we would like to infer that $\const{abe}$ is a grandfather of $\const{bart}$. 
Such transitive reasoning is inherently hard for neural link prediction models as they only learn to score facts locally.
In contrast, symbolic theorem provers like Prolog \citep{gallaire1978logic} enable exactly this type of multi-hop reasoning. 
Furthermore, \gls{ILP} \citep{muggleton1991inductive} builds upon such provers to learn interpretable rules from data and to exploit them for reasoning in \glspl{KB}.
However, symbolic provers lack the ability to learn subsymbolic representations and similarities between them from large \glspl{KB}, which limits their ability to generalize to queries with similar but not identical symbols.  

While the connection between logic and machine learning has been addressed by statistical relational learning approaches, these models traditionally do not support reasoning with subsymbolic representations (\eg{} \cite{kok2007statistical}), and when using subsymbolic representations they are not trained end-to-end from training data (\eg{} \cite{gardner2013improving,gardner2014incorporating,beltagy2017representing}).
Neural multi-hop reasoning models~\citep{neelakantan2015compositional,peng2015towards,das2016chains,weissenborn2016separating,shen2016reasonet} address the aforementioned limitations to some extent by encoding reasoning chains in a vector space or by iteratively refining subsymbolic representations of a question before comparison with answers.
In many ways, these models operate like basic theorem provers, but they lack two of their most crucial ingredients: interpretability and straightforward ways of incorporating domain-specific knowledge in form of rules.

Our approach to this problem is inspired by recent neural network architectures like Neural Turing Machines~\citep{graves2014neural}, Memory Networks~\citep{weston2014memory}, Neural Stacks/Queues~\citep{grefenstette2015learning,joulin2015inferring}, Neural Programmer~\citep{neelakantan2015neural}, Neural Programmer-Interpreters~\citep{reed2015neural}, Hierarchical Attentive Memory~\citep{andrychowicz2016learning} and the Differentiable Forth Interpreter~\citep{bosnjak2016programming}. 
These architectures replace discrete algorithms and data structures by end-to-end differentiable counterparts that operate on real-valued vectors. 
At the heart of our approach is the idea to translate this concept to basic symbolic theorem provers, and hence combine their advantages (multi-hop reasoning, interpretability, easy integration of domain knowledge) with the ability to reason with vector representations of predicates and constants.   
Specifically, we keep variable binding symbolic but compare symbols using their subsymbolic vector representations.

In this chapter we introduce \glspl{NTP}: End-to-end differentiable provers for basic theorems formulated as queries to a \gls{KB}.
We use Prolog's backward chaining algorithm as a recipe for recursively constructing neural networks that are capable of proving queries to a \gls{KB} using subsymbolic representations.
The success score of such proofs is differentiable with respect to vector representations of symbols, which enables us to learn such representations for predicates and constants in ground atoms, as well as parameters of function-free first-order logic rules of predefined structure.
By doing so, \glspl{NTP} learn to place representations of similar symbols in close proximity in a vector space and to induce rules given prior assumptions about the structure of logical relationships in a \gls{KB} such as transitivity.
Furthermore, \glspl{NTP} can seamlessly reason with provided domain-specific rules. 
As \glspl{NTP} operate on distributed representations of symbols, a single hand-crafted rule can be leveraged for many proofs of queries with symbols that have a similar representation. 
Finally, \glspl{NTP} demonstrate a high degree of interpretability as they induce latent rules that we can decode to human-readable symbolic rules.

Our contributions are threefold: (i) We present the construction of \glspl{NTP} inspired by Prolog's backward chaining algorithm and a differentiable unification operation using subsymbolic representations, (ii) we propose optimizations to this architecture by joint training with a neural link prediction model, batch proving, and approximate gradient calculation, and (iii) we experimentally show that \glspl{NTP} can learn representations of symbols and function-free first-order rules of predefined structure, enabling them to learn to perform multi-hop reasoning on benchmark \glspl{KB} and to outperform ComplEx~\citep{trouillon2016complex}, a state-of-the-art neural link prediction model, on three out of four \glspl{KB}.

\section{Differentiable Prover}
\begin{figure}[t!]
\centering
\resizebox{0.5\textwidth}{!}{
  \begin{tikzpicture}
      \at{-3}{0}{
        \draw[ultra thick] (0,0) rectangle (4,2.5);
        \draw[ultra thick, dashed] (2,0) -- (2,2.5);
        \node[text width=2cm, text centered] at (1,1.25) {
          \var{X}/\var{Q}\\
          \var{Y}/\const{bart}
        };

        \name{c1}{\draw[thick, fill=white] (3,2.25) circle (0.1);}
        \name{c2}{\draw[thick, fill=white] (2.5,2) circle (0.1);}
        
        \name{c3}{\draw[thick, fill=white] (3,1.75) circle (0.1);}
        \name{c4}{\draw[thick, fill=white] (3.5,1.5) circle (0.1);}
        \name{c5}{\draw[thick, fill=white] (3,1.25) circle (0.1);}

        \draw[thick, -Latex] (c1) -- (c3);
        \draw[thick, -Latex] (c2) -- (c3);
        \draw[thick, -Latex] (c3) -- (c5);
        \draw[thick, -Latex] (c4) -- (c5);

        \node[anchor=north] at (1, 0) {$\state_\subs$};
        \node[anchor=north] at (3, 0) {$\state_\success$};
      }
      
      \at{3}{0}{
        \draw[ultra thick] (0,0) rectangle (4,2.5);
        \draw[ultra thick, dashed] (2,0) -- (2,2.5);
        \node[text width=2cm, text centered] at (1,1.25) {
          \var{X}/\var{Q}\\
          \var{Y}/\const{bart}\\
          \var{Z}/\const{homer}
        };

        \name{c1}{\draw[thick, fill=white] (3,2.25) circle (0.1);}
        \name{c2}{\draw[thick, fill=white] (2.5,2) circle (0.1);}
        
        \name{c3}{\draw[thick, fill=white] (3,1.75) circle (0.1);}
        \name{c4}{\draw[thick, fill=white] (3.5,1.5) circle (0.1);}
        \name{c5}{\draw[thick, fill=white] (3,1.25) circle (0.1);}

        \name{c6}{\draw[thick, fill=white] (2.5,1) circle (0.1);}
        \name{c7}{\draw[thick, fill=white] (3,.75) circle (0.1);}
        \name{c8}{\draw[thick, fill=white] (3.5,.5) circle (0.1);}
        \name{c9}{\draw[thick, fill=white] (3,.25) circle (0.1);}

        \draw[thick, -Latex] (c1) -- (c3);
        \draw[thick, -Latex] (c2) -- (c3);
        \draw[thick, -Latex] (c3) -- (c5);
        \draw[thick, -Latex] (c4) -- (c5);
        \draw[thick, -Latex] (c5) -- (c7);
        \draw[thick, -Latex] (c6) -- (c7);
        \draw[thick, -Latex] (c7) -- (c9);
        \draw[thick, -Latex] (c8) -- (c9);

        \node[anchor=north] at (1, 0) {$\state'_\subs$};
        \node[anchor=north] at (3, 0) {$\state'_\success$};
      }

      \draw[line width=3pt, -Latex, dashed] (1,1.25) -- (3,1.25);
      \draw[ultra thick] (3.1, 2.5) -- (3.1, 2.6) -- (7.1, 2.6) -- (7.1, 0.1) -- (7, 0.1);
      \draw[ultra thick] (3.2, 2.6) -- (3.2, 2.7) -- (7.2, 2.7) -- (7.2, 0.2) -- (7.1, 0.2);
  \end{tikzpicture}
}
    \caption{A module is mapping an upstream proof state (left) to a list of new proof states (right), thereby extending the substitution set $\state_\subs$ and adding nodes to the computation graph of the neural network $\state_\success$ representing the proof success.}
    \label{fig:state}
\end{figure}
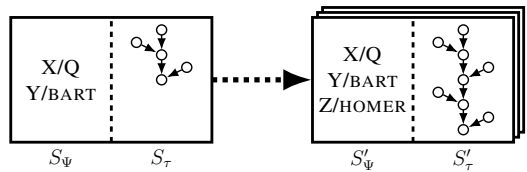

In the following, we describe the recursive construction of \glspl{NTP} -- neural networks for end-to-end differentiable proving that allow us to calculate the gradient of proof successes with respect to vector representations of symbols.
We define the construction of \glspl{NTP} in terms of \emph{modules} similar to dynamic neural module networks \citep{andreas2016learning}. 
Each module takes as inputs \emph{discrete objects} (atoms and rules) and a \emph{proof state}, and returns a list of new proof states (see \Cref{fig:state} for a graphical representation).

A proof state $\state = (\subs, \success)$ is a tuple consisting of the substitution set $\subs$ constructed in the proof so far and a neural network $\success$ that outputs a real-valued success score of a (partial) proof.
While discrete objects and the substitution set are only used during construction of the neural network, once the network is constructed a continuous proof success score can be calculated for many different goals at training and test time.
To summarize, modules are instantiated by discrete objects and the substitution set. 
They construct a neural network representing the (partial) proof success score and recursively instantiate submodules to continue the proof.

The shared signature of modules is
$\datadom{\set{D}} \times \tensordom{\set{S}} \to \tensordom{\set{S}}^N$
where $\datadom{\set{D}}$ is a domain that controls the construction of the network, $\tensordom{\set{S}}$ is the domain of proof states, and $N$ is the number of output proof states. 
Furthermore, let $\state_\subs$ denote the substitution set of the proof state $\state$ and let $\state_\success$ denote the neural network for calculating the proof success.
Akin to the pseudocode of backward chaining in \cref{back}, we use pseudocode in style of a functional programming language to define the behavior of modules and auxiliary functions.

\subsection{Unification Module}
Unification of two atoms, \eg{}, a goal that we want to prove and a rule head, is a central operation in backward chaining.
Two non-variable symbols (predicates or constants) are checked for equality and the proof can be aborted if this check fails. 
However, we want to be able to apply rules even if symbols in the goal and head are not equal but similar in meaning (\eg{} \rel{grandfatherOf} and \rel{grandpaOf}) and thus replace symbolic comparison with a computation that measures the similarity of both symbols in a vector space.

The module \module{unify} updates a substitution set and creates a neural network for comparing the vector representations of non-variable symbols in two sequences of terms. 
The signature of this module is
$\datadom{\set{L} \times \set{L}} \times \tensordom{\set{S}} \to \tensordom{\set{S}}$
where $\set{L}$ is the domain of lists of terms.
\module{unify} takes two atoms represented as lists of terms and an upstream proof state, and maps these to a new proof state (substitution set and proof success).
To this end, \module{unify} iterates through the list of terms of two atoms and compares their symbols. If one of the symbols is a variable, a substitution is added to the substitution set. 
Otherwise, the vector representations of the two non-variable symbols are compared using a \gls{RBF} kernel \citep{broomhead1988radial} where $\mu$ is a hyperparameter that we set to $\frac{1}{\sqrt{2}}$ in our experiments.
The following pseudocode implements \module{unify}. Note that "$\_$" matches every argument and that the order matters, \ie{}, if arguments match a line, subsequent lines are not evaluated.
\begin{flalign*}
1. & \ \ \module{unify}_\params(\emptylist, \emptylist, \state) = \state&\\
2. & \ \ \module{unify}_\params(\emptylist, \_, \_) = \fail&\\
3. & \ \ \module{unify}_\params(\_, \emptylist, \_) = \fail&\\
4. & \ \ \module{unify}_\params(h:\lst{H}, g:\lst{G}, \state) = \module{unify}_\params(\lst{H},\lst{G},\state') = (\state'_\subs, \state'_\success)\quad\text{where}&\\
&\state'_\subs = 
\left\{\begin{array}{ll}
\state_\subs\union\{h/g\}         & \text{if } h\in \set{V}\\
\state_\subs\union\{g/h\}         & \text{if } g\in \set{V}, h\not\in \set{V}\\
\state_\subs   & \text{otherwise}
\end{array}\right\}\\
&\state'_\success =
\min\left(
\state_\success,
\left\{\begin{array}{ll}
\exp\left(\frac{-\|\params_{h:}-\params_{g:}\|_2}{2\mu^2}\right) & \text{if } h,g\not\in \set{V}\\
1 & \text{otherwise}
\end{array}\right\}
\right)
\end{flalign*}
Here, $\state'$ refers to the new proof state, $\set{V}$ refers to the set of variable symbols, $h/g$ is a substitution from the variable symbol $h$ to the symbol $g$, and $\params_{g:}$ denotes the embedding lookup of the non-variable symbol with index $g$.
\module{unify} is parameterized by an embedding matrix $\params\in\R^{|\set{Z}|\times k}$ where $\set{Z}$ is the set of non-variables symbols and $k$ is the dimension of vector representations of symbols.
Furthermore, $\fail$ represents a unification failure due to mismatching arity of two atoms.
Once a failure is reached, we abort the creation of the neural network for this branch of proving.
In addition, we constrain proofs to be cycle-free by checking whether a variable is already bound. 
Note that this is a simple heuristic that prohibits applying the same non-ground rule twice.
There are more sophisticated ways for finding and avoiding cycles in a proof graph such that the same rule can still be applied multiple times (\eg{} \cite{gelder1987efficient}), but we leave this for future work.

\paragraph{Example}
Assume that we are unifying two atoms $[\rel{grandpaOf},\const{abe}, \const{bart}]$ and $[s,\var{Q},i]$ given an upstream proof state $S=(\emptyset, \success)$ where the latter input atom has placeholders for a predicate $s$ and a constant $i$, and the neural network $\success$ would output $0.7$ when evaluated.
Furthermore, assume $\rel{grandpaOf}$, $\const{abe}$ and $\const{bart}$ represent the indices of the respective symbols in a global symbol vocabulary.
Then, the new proof state constructed by \module{unify} is:
{
\begin{align*}
&\module{unify}_\params(\xs{\rel{grandpaOf},\const{abe},\const{bart}}, \xs{s, \var{Q}, i}, (\emptyset, \success)) = (\state'_\subs, \state'_\success) =\\ 
&\qquad\left(\{\var{Q}/\const{abe}\}, \min\left(\success, \exp(-\|\params_{{\scriptsize\rel{grandpaOf}}:}-\params_{s:}\|_2), \exp(-\|\params_{\const{bart}:}-\params_{i:}\|_2)\right)\right)
\end{align*}%
}Thus, the output score of the neural network $\state'_\success$ will be high if the subsymbolic representation of the input $s$ is close to $\rel{grandpaOf}$ and the input $i$ is close to $\const{bart}$.
However, the score cannot be higher than $0.7$ due to the upstream proof success score in the forward pass of the neural network $\success$.
Note that in addition to extending the neural network $\success$ to $\state'_\success$, this module also outputs a substitution set $\{\var{Q}/\const{abe}\}$ at graph creation time that will be used to instantiate submodules.

Furthermore, note that unification is applied multiple times for a proof that involves more than one step, resulting in chained application of the \gls{RBF} kernel and $\min$ operations.
The choice of $\min$ stems from the property that for a successful proof, all unifications should be successful (conjunction).
This could also be realized with a multiplication of unification scores along the proof, but it would likely result in unstable optimization for longer proofs due to exploding gradients.

\subsection{OR Module}
\label{sec:or}
Based on \module{unify}, we now define the \module{or} module which attempts to apply rules in a \gls{KB}.
The signature of \module{or} is 
$\datadom{\set{L} \times \N} \times \tensordom{\set{S}} \to \tensordom{\set{S}}^N$
where $\set{L}$ is the domain of goal atoms and $\N$ is the domain of integers used for specifying the maximum proof depth of the neural network. 
Furthermore, $N$ is the number of possible output proof states for a goal of a given structure and a provided \gls{KB}.\footnote{The creation of the neural network is dependent on the \gls{KB} but also the structure of the goal. For instance, the goal $s(\var{Q},i)$ would result in a different neural network, and hence a different number of output proof states, than $s(i,j)$.}
We implement \module{or} as
{
\begin{flalign*}
1. &\ \ \module{or}^{\kb}_\params(\lst{G}, d, \state) = \xs{\state' \ |\ \state' \in \module{and}^{\kb}_\params(\lss{B}, d, \module{unify}_\params(\lst{H}, \lst{G}, \state)) \text{ for } \ls{H} \lif \lss{B} \in \kb}&
\end{flalign*}%
}where $\ls{H}\lif\lss{B}$ denotes a rule in a given \gls{KB} $\kb$ with a head atom $\ls{H}$ and a list of body atoms $\lss{B}$.
In contrast to the symbolic \por{} method, the \module{or} module is able to use the $\rel{grandfatherOf}$ rule above for a query involving $\rel{grandpaOf}$ provided that the subsymbolic representations of both predicates are similar as measured by the RBF kernel in the \module{unify} module.

\paragraph{Example}
For a goal $[s,\var{Q},i]$, \module{or} would instantiate an \module{and} submodule based on the rule $[\rel{grandfatherOf},\var{X}, \var{Y}]$ \lif $[[\rel{fatherOf},\var{X},\var{Z}],[\rel{parentOf},\var{Z},\var{Y}]]$ as follows
{\small
\begin{align*}
  &\module{or}^{\kb}_\params(\xs{s,\var{Q},i}, d, \state) =\\
  &\qquad\xs{\state'\ |\ \state' \in \module{and}^{\kb}_\params(\xs{\xs{\rel{fatherOf},\var{X},\var{Z}}, \xs{\rel{parentOf},\var{Z},\var{Y}}}, d, \underbrace{(\{\var{X}/\var{Q},\var{Y}/i\}, \hat\state_\success)}_{\text{result of }\module{unify}}), \ldots}
\end{align*}}
\subsection{AND Module}
For implementing \module{and} we first define an auxiliary function called substitute which applies substitutions to variables in an atom if possible.
This is realized via
{
\begin{flalign*}
1. &\ \ \fun{substitute}(\emptylist, \_) = \emptylist&\\
2. &\ \ \fun{substitute}(g : \lst{G}, \subs) = 
\left\{\begin{array}{ll}
x & \text{if } g/x \in \subs\\
g & \text{otherwise}
\end{array}\right\}
: \fun{substitute}(\lst{G}, \subs)
&
\end{flalign*}%
}For example, $\fun{substitute}(\xs{\rel{fatherOf},\var{X},\var{Z}}, \{\var{X}/\var{Q},\var{Y}/i\}) = \xs{\rel{fatherOf},\var{Q},\var{Z}}$.

The signature of \module{and} is $\datadom{\set{L} \times \N} \times \tensordom{\set{S}} \to \tensordom{\set{S}}^N$
where $\set{L}$ is the domain of lists of atoms and $N$ is the number of possible output proof states for a list of atoms with a known structure and a provided \gls{KB}.
This module is implemented as
{
\begin{flalign*}
1. &\ \ \module{and}^{\kb}_\params(\_, \_, \fail) = \fail&\\
2. &\ \ \module{and}^{\kb}_\params(\_, 0, \_) = \fail&\\
3. &\ \ \module{and}^{\kb}_\params(\emptylist, \_, \state) = \state&\\
4. &\ \ \module{and}^{\kb}_\params(\ls{G}:\lss{G}, d, \state) = \\
   & \quad\xs{\state''\ |\ \state''\in\module{and}^{\kb}_\params(\lss{G}, d, \state') \text{ for } \state' \in \module{or}^{\kb}_\params(\fun{substitute}(\ls{G}, \state_\subs), d-1, \state)}&
\end{flalign*}
}where the first two lines define the failure of a proof, either because of an upstream unification failure that has been passed from the \module{or} module (line 1), or because the maximum proof depth has been reached (line 2).
Line 3 specifies a proof success, \ie{}, the list of subgoals is empty before the maximum proof depth has been reached.
Lastly, line 4 defines the recursion: The first subgoal $\ls{G}$ is proven by instantiating an \module{or} module after substitutions are applied, and every resulting proof state $\state'$ is used for proving the remaining subgoals $\lss{G}$ by again instantiating \module{and} modules.

\paragraph{Example}
Continuing the example from \Cref{sec:or}, the \module{and} module would instantiate submodules as follows:
{\small
\begin{flalign*}
&\module{and}^{\kb}_\params(\xs{\xs{\rel{fatherOf},\var{X},\var{Z}}, \xs{\rel{parentOf},\var{Z},\var{Y}}}, d, \underbrace{(\{\var{X}/\var{Q},\var{Y}/i\}, \hat\state_\success)}_{\text{result of }\module{unify}\text{ in }\module{or}}) = &\\
&\qquad\xs{\state''\ |\ \state''\in \module{and}^{\kb}_\params(\xs{\xs{\rel{parentOf},\var{Z},\var{Y}}}, d, \state') \\&\qquad\quad\ \ \ \text{ for } \state' \in \module{or}^{\kb}_\params(\underbrace{\xs{\rel{fatherOf},\var{Q},\var{Z}}}_{\text{result of }\fun{substitute}}, d-1, \underbrace{(\{\var{X}/\var{Q},\var{Y}/i\}, \hat\state_\success)}_{\text{result of }\module{unify}\text{ in }\module{or}})}
\end{flalign*}
}

\subsection{Proof Aggregation}
Finally, we define the overall success score of proving a goal $\lst{G}$ using a \gls{KB} $\kb$ with parameters $\params$ as
{
\begin{align*}
\module{ntp}^{\kb}_\params(\lst{G}, d) &= \argmax_{\substack{\state\ \in\ \module{or}^{\kb}_\params(\lst{G}, d, (\emptyset, 1)) \\ \state \neq\fail}} \state_\success
\end{align*}
}where $d$ is a predefined maximum proof depth and the initial proof state is set to an empty substitution set and a proof success score of $1$.
Hence, the success of proving a goal is a max-pooling operation over the output of neural networks representing all possible proofs up to some depth.

\begin{figure}[t!]
    \centering    
  \includegraphics[width=\textwidth]{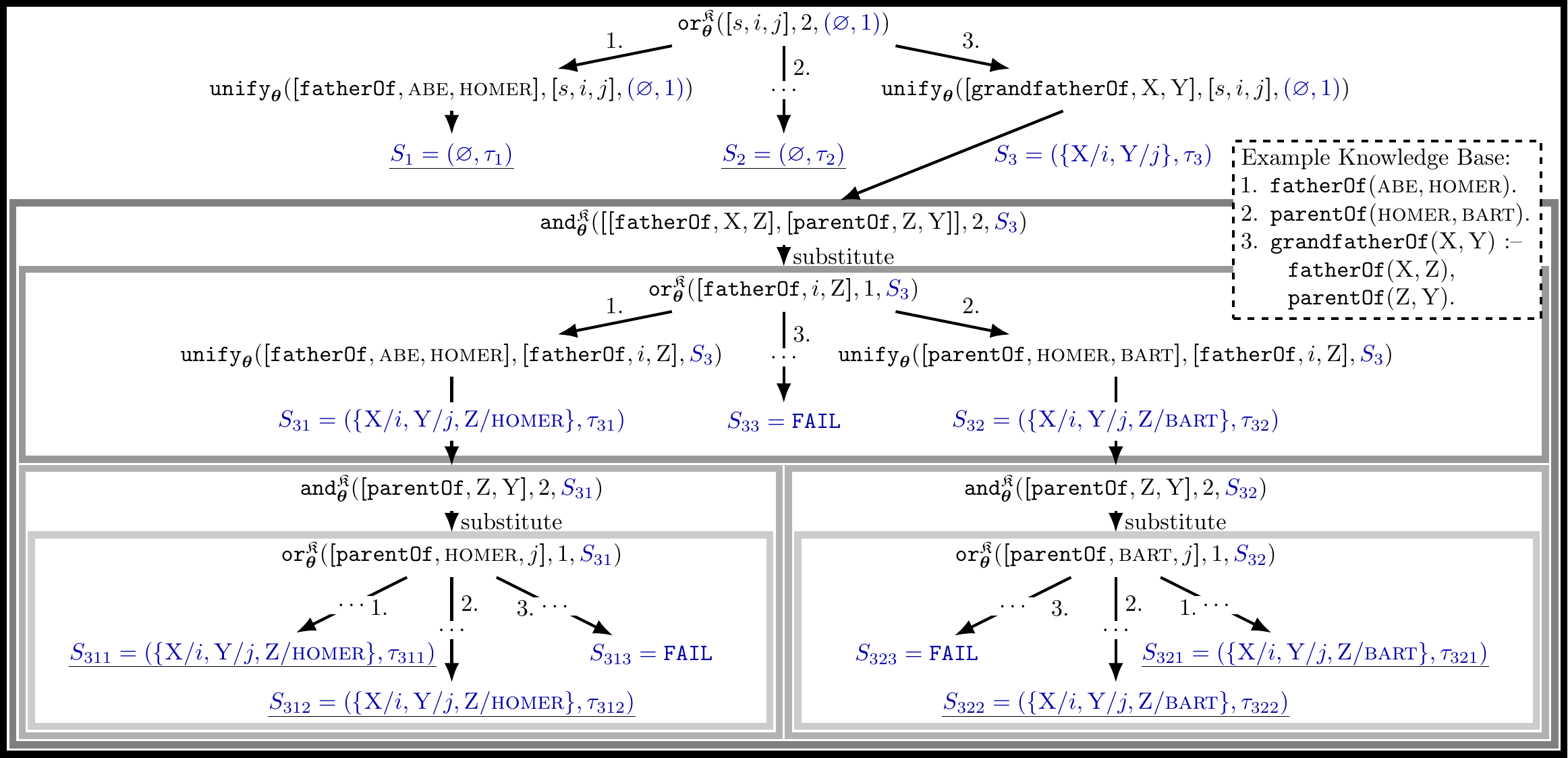}
    \caption{Exemplary construction of an NTP computation graph for a toy knowledge base. Indices on arrows correspond to application of the respective KB rule. Proof states (blue) are subscripted with the sequence of indices of the rules that were applied. Underlined proof states are aggregated to obtain the final proof success. Boxes visualize instantiations of modules (omitted for unify). The proofs $S_{33}, S_{313}$ and $S_{323}$ fail due to cycle-detection (the same rule cannot be applied twice).}
    \label{fig:overview}
\end{figure}

\paragraph{Example} \Cref{fig:overview} illustrates an examplary \gls{NTP} computation graph constructed for a toy \gls{KB}.
Note that such an \gls{NTP} is constructed once before training, and can then be used for proving goals of the structure $[s,i,j]$ at training and test time where $s$ is the index of an input predicate, and $i$ and $j$ are indices of input constants.
Final proof states which are used in proof aggregation are underlined.

\subsection{Neural Inductive Logic Programming}
We can use \glspl{NTP} for \gls{ILP} by gradient descent
instead of a combinatorial search over the space of rules as, for example, done by the \gls{FOIL} \citep{quinlan1990learning}.
Specifically, we are using the concept of learning from entailment \citep{muggleton1991inductive} to induce rules that let us prove known ground atoms, but that do not give high proof success scores to sampled unknown ground atoms. 

Let $\params_{r:},\params_{s:},\params_{t:}\in\R^k$ be representations of some unknown predicates with indices $r,s$ and $t$ respectively.
The prior knowledge of a transitivity between three unknown predicates can be specified via
$r(\var{X}, \var{Y}) \lif s(\var{X}, \var{Z}), t(\var{Z}, \var{Y})$.
We call this a \emph{parameterized rule} as the corresponding predicates are unknown and their representations are learned from data. 
Such a rule can be used for proofs at training and test time in the same way as any other given rule. 
During training, the predicate representations of parameterized rules are optimized jointly with all other subsymbolic representations.
Thus, the model can adapt parameterized rules such that proofs for known facts succeed while proofs for sampled unknown ground atoms fail,
thereby inducing rules of predefined structures like the one above.
Inspired by \cite{wang2015joint}, we use rule templates for conveniently defining the structure of multiple parameterized rules by specifying the number of parameterized rules that should be instantiated for a given rule structure (see \cref{training} for examples).

\subsubsection{Rule Decoding and Implicit Rule Confidence}
For inspection after training, we decode a parameterized rule by searching for the closest representations of known predicates.
Given an induced rule such as $\rparam{1:}(\var{X}, \var{Y}) \lif \rparam{2:}(\var{X}, \var{Z}), \rparam{2:}(\var{Z}, \var{Y})$ where $\rparam{1:}$ and $\rparam{2:}$ have been trained,
we find the closest representation of a known predicate for every parameterized predicate representation in the rule (\eg{}, $\rparam{1:} \to \rparam{{\small\rel{grandparentOf}}:}$, $\rparam{2:}\to\rparam{{\small\rel{parentOf}}:}$).
Formally, we decode $\rparam{i:}$ to a predicate symbol from the set of all predicates $\set{P}$ using
\begin{align}
  \label{ntp:decode}
  \module{decode}(\rparam{i:}) = \arg\max_{r_s\in\set{P}}\exp(-\|\rparam{i:}-\rparam{r_s:}\|_2).
\end{align}

In addition, we provide users with a rule confidence by taking the minimum similarity between unknown and decoded predicate representations using the \gls{RBF} kernel in \module{unify}.
Let $\Theta = [\rparam{i}]$ be the list of predicate representations of a parameterized rule.
The confidence of that rule is then calculated as
\begin{align}
\gamma = \min_{\rparam{i:}\in\Theta}\max_{r_s\in\set{P}}\exp(-\|\rparam{i:}-\rparam{r_s:}\|_2).
\end{align}
This confidence score is an upper bound on the proof success score that can be achieved when the induced rule is used in proofs.

\section{Optimization}
In this section, we present the basic training loss that we use for \glspl{NTP}, a training loss where a neural link prediction models is used as auxiliary task, as well as various computational optimizations.

\subsection{Training Objective}
\label{sec:loss}
Let $\set{K}$ be the set of known facts in a given \gls{KB}.
Usually, we do not observe negative facts and thus resort to sampling corrupted ground atoms as done in previous work~\citep{bordes2013translating}.
Specifically, for every $[s,i,j]\in\set{K}$ we obtain corrupted ground atoms $[s, \hat i, j], [s, i, \hat j], [s, \tilde{i},\tilde{j}]\not\in\set{K}$ by sampling $\hat i, \hat j, \tilde{i}$ and $\tilde{j}$ from the set of constants.
These corrupted ground atoms are resampled in every iteration of training, and we denote the set of known and corrupted ground atoms together with their target score ($1.0$ for known ground atoms and $0.0$ for corrupted ones) as $\set{T}$.
We use the negative log-likelihood of the proof success score as loss function for an \gls{NTP} with parameters $\params$ and a given \gls{KB} $\kb$
{\small
\begin{align*}
\loss_{\module{ntp}^\kb_\params} = \sum_{([s,i,j],y)\ \in\ \set{T}} -y\log(\module{ntp}^{\kb}_\params([s,i,j], d)_\success) - (1-y)\log(1-\module{ntp}^{\kb}_\params([s,i,j], d)_\success)
\end{align*}
} where $[s,i,j]$ is a training ground atom and $y$ its target proof success score.
Note that since in our application all training facts are ground atoms, we only make use of the proof success score $\success$ and not the substitution list of the resulting proof state.
We can prove known facts trivially by a unification with themselves, resulting in no parameter updates during training and hence no generalization.
Therefore, during training we are masking the calculation of the unification success of a known ground atom that we want to prove. Specifically, we set the unification score to $0$ to temporarily hide that training fact and assume it can be proven from other facts and rules in the \gls{KB}.

\subsection{Neural Link Prediction as Auxiliary Loss}
At the beginning of training all subsymbolic representations are initialized randomly.
When unifying a goal with all facts in a \gls{KB} we consequently get very noisy success scores in early stages of training.
Moreover, as only the maximum success score will result in gradient updates for the respective subsymbolic representations along the maximum proof path, it can take a long time until \glspl{NTP} learn to place similar symbols close to each other in the vector space and to make effective use of rules. 

To speed up learning subsymbolic representations, we train \glspl{NTP} jointly with ComplEx \citep{trouillon2016complex} (\cref{eq:complex} in \cref{back}).
ComplEx and the \gls{NTP} share the same subsymbolic representations, which is feasible as the \gls{RBF} kernel in \module{unify} is also defined for complex vectors.
While the \gls{NTP} is responsible for multi-hop reasoning, the neural link prediction model learns to score ground atoms locally.
At test time, only the \gls{NTP} is used for predictions.
Thus, the training loss for ComplEx can be seen as an auxiliary loss for the subsymbolic representations learned by the \gls{NTP}.
We term the resulting model \gls{NTP}$\lambda$.
Based on the loss in \Cref{sec:loss}, the joint training loss is defined as
\begin{align*}
  \loss_{\module{ntp}\lambda^\kb_\params} = \loss_{\module{ntp}^\kb_\params}+\sum_{([s,i,j],y)\ \in\ \set{T}}\big( & -y\log(\module{complex}_\params(s,i,j)) \\
         & - (1-y)\log(1-\module{complex}_\params(s,i,j))\big)
\end{align*}
where $[s,i,j]$ is a training atom, $y$ is its ground truth target, and $\module{complex}(s,i,j)=p_{sij}$ as defined in \cref{eq:complex}.

\subsection{Computational Optimizations}
\glspl{NTP} as described above suffer from severe computational limitations since the neural network is representing all possible proofs up to some predefined depth.
In contrast to symbolic backward chaining where a proof can be aborted as soon as unification fails, in differentiable proving we only get a unification failure for atoms whose arity does not match or when we detect cyclic rule application.
We propose two optimizations to speed up \glspl{NTP}. 
First, we make use of modern GPUs by batch processing many proofs in parallel (\Cref{sec:batch}).
Second, we exploit the sparseness of gradients caused by the $\min$ and $\max$ operations used in the unification and proof aggregation respectively to derive a heuristic for a truncated forward and backward pass that drastically reduces the number of proofs that have to be considered for calculating gradients (\Cref{sec:kmax}).

\subsubsection{Batch Proving}
\label{sec:batch}
Let $\mat{A} \in \R^{N\times k}$ be a matrix of $N$ subsymbolic representations that are to be unified with $M$ other representations $\mat{B}\in\R^{M\times k}$.
We can adapt the unification module to calculate the unification success in a batched way using
{\small
\begin{align*}
\exp\left(- \sqrt{
\left(
\left[
    \begin{array}{l}
      \sum_{i=1}^k \mat{A}^2_{1i}\\
      \qquad\vdots\\
      \sum_{i=1}^k \mat{A}^2_{Ni}
    \end{array}
   \right]\vec{1}^\top_M
\right)
+ 
\left(
\vec{1}_N\left[
    \begin{array}{l}
      \sum_{i=1}^k \mat{B}^2_{1i}\\
      \qquad\vdots\\
      \sum_{i=1}^k \mat{B}^2_{Mi}
    \end{array}
   \right]
^\top\right) - 2\mat{A}\mat{B}^\top}\right)&\in\R^{N\times M}
\end{align*}
}%
where $\vec{1}_N$ and $\vec{1}_M$ are vectors of $N$ and $M$ ones respectively, and the square root is taken element-wise.
In practice, we partition the \gls{KB} into rules that have the same structure and batch-unify goals with all rule heads per partition at the same time on a \gls{GPU}.
Furthermore, substitution sets bind variables to vectors of symbol indices instead of single symbol indices, and min and max operations are taken per goal.

\subsubsection{$K\max$ Gradient Approximation}
\label{sec:kmax}
\glspl{NTP} allow us to calculate the gradient of proof success scores with respect to subsymbolic representations and rule parameters.
While backpropagating through this large computation graph will give us the exact gradient, it is computationally infeasible for any reasonably-sized \gls{KB}.
Consider the parameterized rule $\rparam{1:}(\var{X}, \var{Y}) \lif \rparam{2:}(\var{X}, \var{Z}), \rparam{3:}(\var{Z}, \var{Y})$ and let us assume the given \gls{KB} contains $1\ 000$ facts with binary predicates.
While \var{X} and \var{Y} will be bound to the respective representations in the goal, \var{Z} we will be substituted with every possible second argument of the $1\ 000$ facts in the \gls{KB} when proving the first atom in the body.
Moreover, for each of these $1\ 000$ substitutions, we will again need to compare with all facts in the \gls{KB} when proving the second atom in the body of the rule, resulting in $1\ 000\ 000$ proof success scores.
However, note that since we use the max operator for aggregating the success of different proofs, only subsymbolic representations in one out of $1\ 000\ 000$ proofs will receive gradients.

To overcome this computational limitation, we propose the following heuristic. 
We assume that when unifying the first atom with facts in the \gls{KB}, it is unlikely for any unification successes below the top $K$ successes to attain the maximum proof success when unifying the remaining atoms in the body of a rule with facts in the \gls{KB}.
That is, after the unification of the first atom, we only keep the top $K$ substitutions and their success scores, and continue proving only with these.
This means that all other partial proofs will not contribute to the forward pass at this stage, and consequently not receive any gradients on the backward pass of backpropagation.
We term this the $K\max$ heuristic.
Note that we cannot guarantee anymore that the gradient of the proof success is the exact gradient, but for a large enough $K$ we get a close approximation to the true gradient.

\section{Experiments}
Consistent with previous work, we carry out experiments on four benchmark \glspl{KB} and compare ComplEx with the \gls{NTP} and \gls{NTP}$\lambda$ in terms of area under the Precision-Recall-curve (AUC-PR) on the Countries \gls{KB}, and \gls{MRR} and HITS@$m$ \citep{bordes2013translating} on the other \glspl{KB} described below.
Training details, including hyperparameters and rule templates, can be found in \Cref{training}.

\paragraph{Countries}
\begin{figure}[t!]
  \centering
  \begin{subfigure}[t]{0.47\textwidth}
    \includegraphics[]{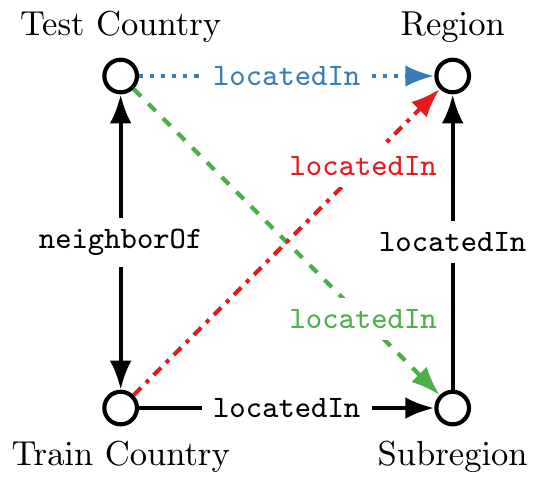}
    \caption{}
    \label{fig:countriesa}
  \end{subfigure}\hfill%
  \begin{subfigure}[t]{0.47\textwidth}
    \includegraphics[]{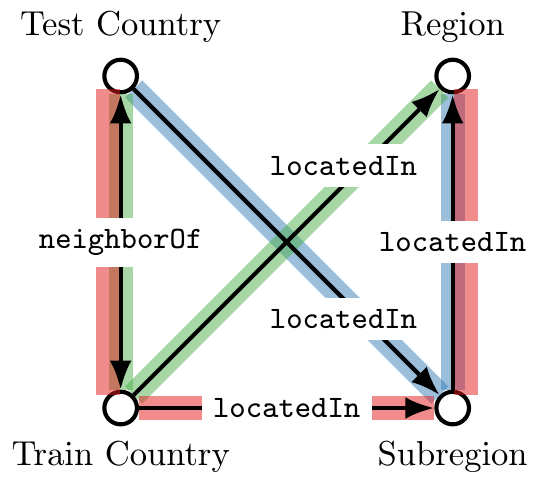}
    \caption{}
    \label{fig:countriesb}
  \end{subfigure}
  \caption{Overview of different tasks in the Contries dataset as visualized by \cite{nickel2015holographic}. The left part (a) shows which atoms are removed for each task (dotted lines), and the right part (b) illustrates the rules that can be used to infer the location of test countries. For task \textbf{S1}, only facts corresponding to the blue dotted line are removed from the training set. For task \textbf{S2}, additionally facts corresponding to the green dashed line are removed. Finally, for task \textbf{S3} also facts for the red dash-dotted line are removed.}
  \label{fig:countries}
\end{figure}

The Countries \gls{KB} is a dataset introduced by \cite{bouchard2015approximate} for testing reasoning capabilities of neural link prediction models.
It consists of $244$ countries, $5$ regions (\eg{} \ent{europe}), $23$ subregions (\eg{} \ent{western europe}, \ent{northern america}), and $1158$ facts about the neighborhood of countries, and the location of countries and subregions.
We follow \cite{nickel2015holographic} and split countries randomly into a training set of $204$ countries (train), a development set of $20$ countries (dev), and a test set of $20$ countries (test), such that every dev and test country has at least one neighbor in the training set.
Subsequently, three different task datasets are created.
For all tasks, the goal is to predict $\rel{locatedIn}(c, r)$ for every test country $c$ and all five regions $r$, but the access to training atoms in the \gls{KB} varies (see \cref{fig:countries}).\\
\textbf{S1:} All ground atoms $\rel{locatedIn}(c, r)$ where $c$ is a test country and $r$ is a region are removed from the \gls{KB}. 
Since information about the subregion of test countries is still contained in the \gls{KB}, this task can be solved by using the transitivity rule
\[\footnotesize
\rel{locatedIn}(\var{X}, \var{Y}) \lif \rel{locatedIn}(\var{X}, \var{Z}), \rel{locatedIn}(\var{Z}, \var{Y}).
\] 
\textbf{S2:} In addition to \textbf{S1}, all ground atoms $\rel{locatedIn}(c, s)$ are removed where $c$ is a test country and $s$ is a subregion.
The location of test countries needs to be inferred from the location of its neighboring countries:
\[\footnotesize
\rel{locatedIn}(\var{X}, \var{Y}) \lif \rel{neighborOf}(\var{X}, \var{Z}), \rel{locatedIn}(\var{Z}, \var{Y}).
\] 
This task is more difficult than \textbf{S1}, as neighboring countries might not be in the same region, so the rule above will not always hold.\\
\textbf{S3:} In addition to \textbf{S2}, all ground atoms $\rel{locatedIn}(c, r)$ where $r$ is a region and $c$ is a training country that has a test or dev country as a neighbor are also removed. 
The location of test countries can for instance be inferred using the three-hop rule
\[\footnotesize
\rel{locatedIn}(\var{X}, \var{Y}) \lif \rel{neighborOf}(\var{X}, \var{Z}), \rel{neighborOf}(\var{Z}, \var{W}), \rel{locatedIn}(\var{W}, \var{Y}).
\]

\paragraph{Kinship, Nations \& UMLS} We use the Nations, Alyawarra kinship (Kinship) and Unified Medical Language System (UMLS) \glspl{KB} from \cite{kok2007statistical}.
We left out the Animals dataset as it only contains unary predicates and can thus not be used for evaluating multi-hop reasoning.
Nations contains $56$ binary predicates, $111$ unary predicates, $14$ constants and $2565$ true facts, Kinship contains $26$ predicates, $104$ constants and $10686$ true facts, and UMLS contains $49$ predicates, $135$ constants and $6529$ true facts.
Since our baseline ComplEx cannot deal with unary predicates, we remove unary atoms from Nations.
We split every \gls{KB} into $80\%$ training facts, $10\%$ development facts and $10\%$ test facts.
For evaluation, we take a test fact and corrupt its first and second argument in all possible ways such that the corrupted fact is not in the original \gls{KB}.
Subsequently, we predict a ranking of every test fact and its corruptions to calculate \gls{MRR} and HITS@$m$.

\subsection{Training Details}
\label{training}
We use ADAM \citep{kingma2014adam} with an initial learning rate of $0.001$ and a mini-batch size of $50$ ($10$ known and $40$ corrupted atoms) for optimization.
We apply an $\ell_2$ regularization of $0.01$ to all model parameters, and clip gradient values at $[-1.0, 1.0]$. 
All subsymbolic representations and rule parameters are initialized using Xavier initialization \citep{glorot2010understanding}.
We train all models for $100$ epochs and repeat every experiment on the Countries corpus ten times.
Statistical significance is tested using the independent $t$-test.
All models are implemented in TensorFlow \citep{abadi2015tensorflow}.
We use a maximum proof depth of $d=2$ and add the following rule templates where the number in front of the rule template indicates how often a parameterized rule of the given structure will be instantiated. Note that a rule template such as $\#1(\var{X}, \var{Y}) \lif \#2(\var{X}, \var{Z}), \#2(\var{Z}, \var{Y})$ specifies that the two predicate representations in the body are shared.\\

\noindent
\textbf{Countries S1}\\
3   $\#1(\var{X}, \var{Y}) \lif \#1(\var{Y}, \var{X}).$\\
3   $\#1(\var{X}, \var{Y}) \lif \#2(\var{X}, \var{Z}), \#2(\var{Z}, \var{Y}).$\\

\noindent
\textbf{Countries S2}\\
3   $\#1(\var{X}, \var{Y}) \lif \#1(\var{Y}, \var{X}).$\\
3   $\#1(\var{X}, \var{Y}) \lif \#2(\var{X}, \var{Z}), \#2(\var{Z}, \var{Y}).$\\
3   $\#1(\var{X}, \var{Y}) \lif \#2(\var{X}, \var{Z}), \#3(\var{Z}, \var{Y}).$\\

\noindent
\textbf{Countries S3}\\
3   $\#1(\var{X}, \var{Y}) \lif \#1(\var{Y}, \var{X}).$\\
3   $\#1(\var{X}, \var{Y}) \lif \#2(\var{X}, \var{Z}), \#2(\var{Z}, \var{Y}).$\\
3   $\#1(\var{X}, \var{Y}) \lif \#2(\var{X}, \var{Z}), \#3(\var{Z}, \var{Y}).$\\
3   $\#1(\var{X}, \var{Y}) \lif \#2(\var{X}, \var{Z}), \#3(\var{Z}, \var{W}), \#4(\var{W}, \var{Y}).$\\

\noindent
\textbf{Kinship, Nations \& UMLS}\\
20   $\#1(\var{X}, \var{Y}) \lif \#2(\var{X}, \var{Y}).$\\
20   $\#1(\var{X}, \var{Y}) \lif \#2(\var{Y}, \var{X}).$\\
20   $\#1(\var{X}, \var{Y}) \lif
        \#2(\var{X}, \var{Z}),
        \#3(\var{Z}, \var{Y}).$

\section{Results and Discussion}
\begin{table}[t!]
    \centering
    \caption{AUC-PR results on Countries and MRR and HITS@$m$ on Kinship, Nations, and UMLS.}
    \label{tab:results}    
\resizebox{\textwidth}{!}{        
    \begin{tabular}{lr|lrrr|l}
        \toprule
        \multicolumn{2}{c|}{Corpus} & \multicolumn{1}{c}{Metric} & \multicolumn{3}{c|}{Model} & \multicolumn{1}{c}{Examples of induced rules and their confidence}\\
        \cmidrule(lr){4-6}
        & & & \multicolumn{1}{c}{\textbf{ComplEx}} & \multicolumn{1}{c}{\textbf{NTP}} & \multicolumn{1}{c|}{\textbf{NTP}$\bm{\lambda}$}\\
        \midrule
        \multirow{3}{*}{Countries}
        & S1 & AUC-PR & $99.37 \pm 0.4$ & $90.83 \pm 15.4$  & $\bm{100.00} \pm\ \ 0.0$ & $0.90$ \rel{locatedIn}(\var{X},\var{Y}) \lif \rel{locatedIn}(\var{X},\var{Z}), \rel{locatedIn}(\var{Z},\var{Y}).\\
        & S2 & AUC-PR & $87.95 \pm 2.8$ & $87.40 \pm 11.7$  & $\bm{93.04}  \pm\ \ 0.4$ & $0.63$ \rel{locatedIn}(\var{X},\var{Y}) \lif \rel{neighborOf}(\var{X},\var{Z}), \rel{locatedIn}(\var{Z},\var{Y}).\\
        & S3 & AUC-PR & $48.44 \pm 6.3$ & $56.68 \pm 17.6$  & $\bm{77.26}  \pm 17.0$ & $0.32$ \rel{locatedIn}(\var{X},\var{Y}) \lif \\ 
        &&&&&& $\qquad\quad$\rel{neighborOf}(\var{X},\var{Z}), \rel{neighborOf}(\var{Z},\var{W}), \rel{locatedIn}(\var{W},\var{Y}).\\
        \midrule
        \multirow{4}{*}{Kinship}
        && MRR & $\bf{0.81}$ & $0.60$& $0.80$ & $0.98$ \rel{term15}(\var{X},\var{Y}) \lif \rel{term5}(\var{Y},\var{X})\\
        && HITS@1 & $0.70$ & $0.48$ & $\bm{0.76}$ & $0.97$ \rel{term18}(\var{X},\var{Y}) \lif \rel{term18}(\var{Y},\var{X})\\
        && HITS@3 & $\bm{0.89}$ & $0.70$ & $0.82$ & $0.86$ \rel{term4}(\var{X},\var{Y}) \lif \rel{term4}(\var{Y},\var{X})\\
        && HITS@10 & $\bm{0.98}$ & $0.78$ & $0.89$ &  $0.73$ \rel{term12}(\var{X},\var{Y}) \lif \rel{term10}(\var{X}, \var{Z}), \rel{term12}(\var{Z}, \var{Y}).\\
        \midrule
        \multirow{4}{*}{Nations}
        && MRR & $\bm{0.75}$ & $\bm{0.75}$ & $0.74$ & $0.68$  \rel{blockpositionindex}(\var{X},\var{Y}) \lif \rel{blockpositionindex}(\var{Y},\var{X}).\\
        && HITS@1 & $\bm{0.62}$ & $\bm{0.62}$ & $0.59$ & $0.46$ \rel{expeldiplomats}(\var{X},\var{Y}) \lif \rel{negativebehavior}(\var{X},\var{Y}).\\
        && HITS@3 & $0.84$ & $0.86$ & $\bm{0.89}$ & $0.38$ \rel{negativecomm}(\var{X},\var{Y}) \lif \rel{commonbloc0}(\var{X},\var{Y}).\\
        && HITS@10 & $\bm{0.99}$ & $\bm{0.99}$ & $\bm{0.99}$ & $0.38$ \rel{intergovorgs3}(\var{X},\var{Y}) \lif \rel{intergovorgs}(\var{Y},\var{X}).\\
        \midrule
        \multirow{4}{*}{UMLS}
        && MRR & $0.89$ & $0.88$ & $\bm{0.93}$ & $0.88$ \rel{interacts\_with}(\var{X},\var{Y}) \lif\\ 
        && HITS@1 & $0.82$ & $0.82$ & $\bm{0.87}$ & $\qquad\quad$\rel{interacts\_with}(\var{X},\var{Z}), \rel{interacts\_with}(\var{Z},\var{Y}).\\ 
        && HITS@3 & $0.96$ & $0.92$ & $\bm{0.98}$ & $0.77$ \rel{isa}(\var{X},\var{Y}) \lif \rel{isa}(\var{X},\var{Z}), \rel{isa}(\var{Z},\var{Y}).\\ 
        && HITS@10 & $\bm{1.00}$ & $0.97$ & $\bm{1.00}$ & $0.71$ \rel{derivative\_of}(\var{X},\var{Y}) \lif\\ 
        &&&&&& $\qquad\quad$\rel{derivative\_of}(\var{X},\var{Z}), \rel{derivative\_of}(\var{Z},\var{Y}).\\
        \bottomrule
    \end{tabular}
}    
\end{table}

Results for the different model variants on the benchmark \glspl{KB} are shown in \Cref{tab:results}.
Another method for inducing rules in a differentiable way for automated \gls{KB} completion has been introduced recently by \cite{yang2017differentiable} and our evaluation setup is equivalent to their Protocol II.
However, our neural link prediction baseline, ComplEx, already achieves much higher HITS@10 results ($1.00$ vs. $0.70$ on UMLS and $0.98$ vs. $0.73$ on Kinship).
We thus focus on the comparison of \glspl{NTP} with ComplEx.

First, we note that vanilla \glspl{NTP} alone do not work particularly well compared to ComplEx. 
They only outperform ComplEx on Countries S3 and Nations, but not on Kinship or UMLS.
This demonstrates the difficulty of learning subsymbolic representations in a differentiable prover from unification alone, and the need for auxiliary losses.
The \gls{NTP}$\lambda$ with ComplEx as auxiliary loss outperforms the other models in the majority of tasks.
The difference in AUC-PR between ComplEx and \gls{NTP}$\lambda$ is significant for all Countries tasks ($p < 0.0001$).

A major advantage of \glspl{NTP} is that we can inspect induced rules which provide us with an interpretable representation of what the model has learned.
The right column in \Cref{tab:results} shows examples of induced rules by NTP$\lambda$ (note that predicates on Kinship are anonymized).
For Countries, the \gls{NTP} recovered those rules that are needed for solving the three different tasks.
On UMLS, the \gls{NTP} induced transitivity rules. 
Those relationships are particularly hard to encode by neural link prediction models like ComplEx, as they are optimized to locally predict the score of a fact.

\section{Related Work}
Combining neural and symbolic approaches to relational learning and reasoning has a long tradition and let to various proposed architectures over the past decades (see \cite{garcez2012neural} for a review). 
Early proposals for neural-symbolic networks are limited to \emph{propositional rules} (\eg{}, EBL-ANN~\citep{shavlik1989approach}, KBANN~\citep{towell1994knowledge} and C-IL$^2$P~\citep{garcez1999connectionist}).
Other neural-symbolic approaches focus on first-order inference, but do not learn subsymbolic vector representations from training facts in a \gls{KB} (\eg{}, SHRUTI~\citep{shastri1992neurally}, Neural Prolog~\citep{ding1995neural}, CLIP++ \citep{franca2014fast}, Lifted Relational Neural Networks \citep{sourek2015lifted}, and TensorLog \citep{cohen2016tensorlog}).
Logic Tensor Networks \citep{serafini2016logic} are in spirit similar to \glspl{NTP}, but need to fully ground first-order logic rules. 
However, they support function terms, whereas \glspl{NTP} currently only support function-free terms.

Recent question-answering architectures such as \citep{peng2015towards,weissenborn2016separating,shen2016reasonet} translate query representations implicitly in a vector space without explicit rule representations and can thus not easily incorporate domain-specific knowledge. 
In addition, \glspl{NTP} are related to random walk \citep{lao2011random,lao2012reading,gardner2013improving,gardner2014incorporating} and path encoding models \citep{neelakantan2015compositional,das2016chains}.
However, instead of aggregating paths from random walks or encoding paths to predict a target predicate, reasoning steps in \glspl{NTP} are explicit and only unification uses subsymbolic representations.
This allows us to induce interpretable rules, as well as to incorporate prior knowledge either in the form of rules or in the form of rule templates which define the structure of logical relationships that we expect to hold in a \gls{KB}.
Another line of work \citep{rocktaschel2014low,rocktaschel2015injecting,vendrov2016order,hu2016harnessing,demeester2016lifted} regularizes distributed representations via domain-specific rules, but these approaches do not learn such rules from data and only support a restricted subset of first-order logic. 
\glspl{NTP} are constructed from Prolog's backward chaining and are thus related to Unification Neural Networks \citep{komendantskaya2011unification,holldobler1990structured}.
However, \glspl{NTP} operate on vector representations of symbols instead of scalar values, which are more expressive.

As \glspl{NTP} can learn rules from data, they are related to \gls{ILP} systems such as \gls{FOIL} \citep{quinlan1990learning}, Sherlock \citep{schoenmackers2010learning} and meta-interpretive learning of higher-order dyadic Datalog (Metagol) \citep{muggleton2015meta}.
While these \gls{ILP} systems operate on symbols and search over the discrete space of logical rules, \glspl{NTP} work with subsymbolic representations and induce rules using gradient descent.
Recently, \cite{yang2017differentiable} introduced a differentiable rule learning system based on TensorLog and a neural network controller similar to LSTMs \citep{hochreiter1997long}. 
Their method is more scalable than the \glspl{NTP} introduced here. 
However, on UMLS and Kinship our baseline already achieved stronger generalization by learning subsymbolic representations.
Still, scaling \glspl{NTP} to larger \glspl{KB} for competing with more scalable relational learning methods is an open problem that we seek to address in future work.

\section{Summary}
We proposed an end-to-end differentiable prover for automated \gls{KB} completion that operates on subsymbolic representations.
To this end, we used Prolog's backward chaining algorithm as a recipe for recursively constructing neural networks that can be used to prove queries to a \gls{KB}. 
Specifically, our contribution is the use of a differentiable unification operation between vector representations of symbols to construct such neural networks.
This allowed us to compute the gradient of proof successes with respect to vector representations of symbols, and thus enabled us to train subsymbolic representations end-to-end from facts in a \gls{KB}, and to induce function-free first-order logic rules using gradient descent.
On benchmark \glspl{KB}, our model outperformed ComplEx, a state-of-the-art neural link prediction model, on three out of four \glspl{KB} while at the same time inducing interpretable rules.

\chapter{Recognizing Textual Entailment with Recurrent Neural Networks}
\label{rte}
\glsresetall

``\emph{You can’t cram the meaning of a whole \%\&!\$\# sentence into a single \$\&!\#* vector!}''
\begin{flushright}
--- Raymond J. Mooney
\end{flushright}

The ability to determine the logical relationship between two natural language sentences is an integral part for machines that are supposed to understand and reason with language.
In previous chapters, we have discussed ways of combining symbolic logical knowledge with subsymbolic representations trained via neural networks. 
As first steps towards models that reason with natural language, we have used textual surface form patterns as predicates for automated \gls{KB} completion.
However, for automated \gls{KB} completion, we assumed surface patterns to be atomic, which does not generalize to unseen patterns as there is no compositional representation.
In this chapter, we are using \glspl{RNN} for learning compositional representations of natural language sentences.
Specifically, we are tackling the task of \gls{RTE}, \ie{}, determining (i) whether two natural language sentences are contradicting each other, (ii) whether they are unrelated, or (iii) whether the first sentence (called the \emph{premise}) entails the second sentence (called the \emph{hypothesis}).
For instance, the sentence ``Two girls and a guy are involved in a pie eating contest'' entails ``Three people are stuffing their faces'', but contradicts ``Three people are drinking beer on a boat''.
This task is important since many \gls{NLP} tasks, such as information extraction, relation extraction, text summarization or machine translation, rely on it explicitly or implicitly and could benefit from more accurate \gls{RTE} systems \citep{dagan2005pascal}.

State-of-the-art systems for \gls{RTE} so far relied heavily on engineered \gls{NLP} pipelines, extensive manual creation of features, as well as various external resources and specialized subcomponents such as negation detection \citep[\emph{e.g.}][]{lai2014illinois, jimenez2014unal, zhao2014ecnu, beltagy2015representing}.
Despite the success of neural networks for paraphrase detection \citep[\emph{e.g.}][]{socher2011dynamic, hu2014convolutional, yin2015convolutional}, end-to-end differentiable neural architectures so far failed to reach acceptable performance for \gls{RTE} due to the lack of large high-quality datasets.
An end-to-end differentiable solution to \gls{RTE} is desirable since it avoids specific assumptions about the underlying language.
In particular, there is no need for language features like part-of-speech tags or dependency parses.
Furthermore, a generic sequence-to-sequence solution allows us to extend the concept of capturing entailment across any sequential data, not only natural language.

Recently, \cite{bowman2015large} published the \gls{SNLI} corpus accompanied by a \gls{LSTM} baseline \citep{hochreiter1997long} which achieves an accuracy of $77.6\%$ for \gls{RTE} on this dataset.
It is the first instance of a generic neural model without hand-crafted features that got close to the accuracy of a simple lexicalized classifier with engineered features for \gls{RTE}.
This can be explained by the high quality and size of \gls{SNLI} compared to the two orders of magnitude smaller and partly synthetic datasets used so far to evaluate \gls{RTE} systems.
\citeauthor{bowman2015large}'s LSTM encodes the premise and hypothesis independently as dense fixed-length vectors whose concatenation is subsequently used in a \gls{MLP} for classification (\cref{sec:independent}).
In contrast, we are proposing a neural network that is capable of fine-grained comparison of pairs of words and phrases by processing the hypothesis \emph{conditioned on the premise} and using a neural attention mechanism.

Our contributions are threefold:
(i) we present a neural model based on two \glspl{LSTM} that read two sentences in one go to determine entailment as opposed to mapping each sentence independently into a vector space (\cref{sec:cond}),
(ii) we extend this model with a neural word-by-word attention mechanism to encourage more fine-grained comparison of pairs of words and phrases (\cref{sec:iatt}), and
(iii) we provide a detailed qualitative analysis of neural attention for \gls{RTE} (\cref{sec:analysis}).
Our benchmark \gls{LSTM} achieves an accuracy of $80.9\%$ on \gls{SNLI}, outperforming a simple lexicalized classifier tailored to \gls{RTE} by $2.7$ percentage points.
An extension with word-by-word neural attention surpasses this strong benchmark \gls{LSTM} result by $2.6$ percentage points, achieving an accuracy of $83.5\%$ for recognizing entailment on \gls{SNLI}.

\section{Background}
\label{sec:background}
In this section, we describe \glspl{RNN} and \glspl{LSTM} for sequence modeling, before explaining how \glspl{LSTM} are used in the independent sentence encoding model for \gls{RTE} by \cite{bowman2015large}.

\subsection{Recurrent Neural Networks}

\begin{figure}[t!]
  \centering
  \includegraphics[scale=0.8]{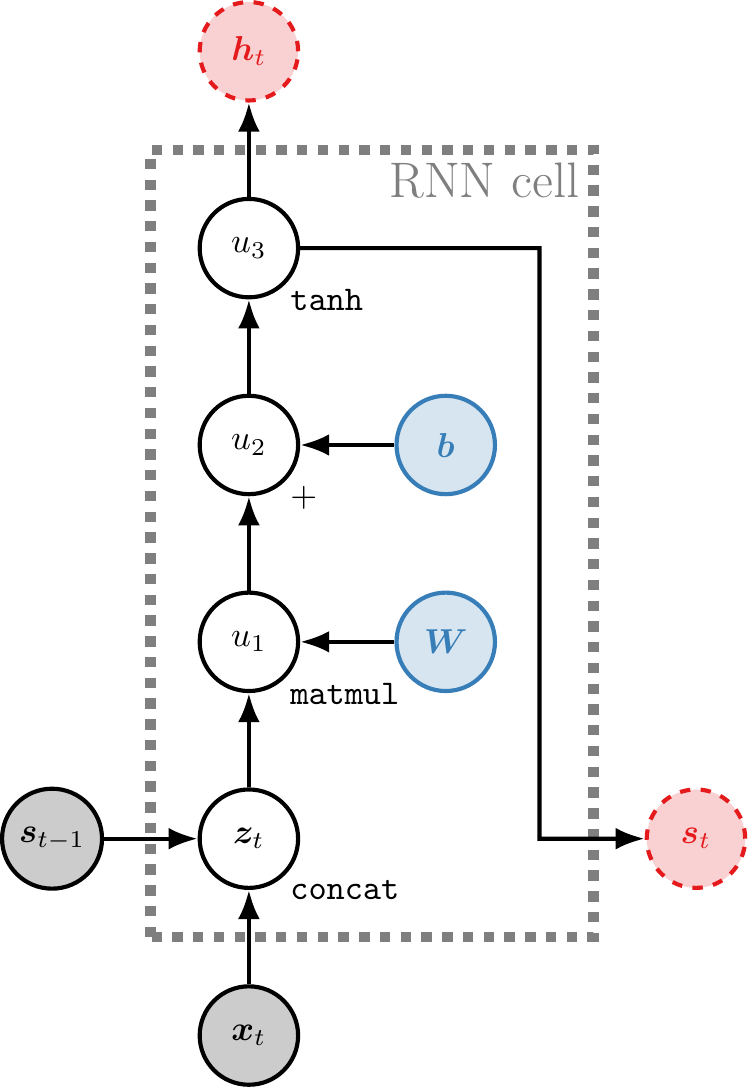}
  \caption{Computation graph for the fully-connected RNN cell.}
  \label{fig:rnn_cell}
\end{figure}

An \gls{RNN} is parameterized by a differentiable \emph{cell function} $f_\theta: \R^k\times\R^s \to \R^o\times\R^s$ that maps an input vector $\vec{x}_t \in \R^k$ and previous state $\vec{s}_{t-1}\in\R^s$ to an output vector $\vec{h}_t\in\R^o$ and next state $\vec{s}_t\in\R^s$. 
For simplicity, we assume that the input size and output size are the same, \ie{}, $\vec{x}_t,\vec{h}_t \in\R^k$.
By applying the cell function at time-step $t$, we obtain an output vector $\vec{h}_t$ and the next state $\vec{s}_t$:
\begin{equation}
\vec{h}_t, \vec{s}_t = f_\theta(\vec{x}_t, \vec{s}_{t-1}). 
\end{equation}
Given a sequence of input representations $\ls{x} = [\vec{x}_1,\ldots,\vec{x}_T]$ and a start state $\vec{s}_0$, the output of the \gls{RNN} over the entire input sequence is then obtained by recursively applying the cell function:
\begin{equation}
\label{eq:rnn}
\rnn(f_\theta, \ls{x}, \vec{s}_0) = [f_\theta(\vec{x}_1, \vec{s}_0), \ldots, f_\theta(\vec{x}_T, \vec{s}_{T-1})].
\end{equation}
Usually, this output is separated into a list of output vectors $\ls{h} = [\vec{h}_1,\ldots,\vec{h}_T]$ and states $\ls{s} = [\vec{s}_1,\ldots,\vec{s}_T]$.
Note that $\rnn$ can be applied to input sequences of varying length, such as sequences of word representations.

\subsubsection{Fully-connected Recurrent Neural Network}
The most basic \gls{RNN} is parameterized by the following cell function
 \begin{align}
    \vec{z}_t &= \left[{
      \begin{array}{*{20}c}
        \vec{x}_t \\
        \vec{s}_{t-1}
      \end{array} }
    \right]\\
    \vec{h}_t &= \tanh(\mat{W}\vec{z}_t+\vec{b})\\
    \vec{s}_t &= \vec{h}_t\\
    f_\theta^{\rnn}(\vec{x}_t,\vec{s}_{t-1}) &= (\vec{h}_t, \vec{s}_t)
  \end{align}
where $\mat{W}\in\R^{2k\times k}$ is a trainable transition matrix, $\vec{b}\in\R^{k}$ a trainable bias, and $\tanh$ the element-wise application of the hyperbolic function.
We call this a fully-connected \gls{RNN}, as the cell function is modeled by a dense layer.
We illustrate the computation graph for a single application of the fully-connected \gls{RNN} cell in \cref{fig:rnn_cell}.
Note that for recurrent application of this cell function to a sequence of inputs, all parameters (transition matrix and bias) are shared between all time steps.
\maybe{vanishing and exploding gradients}

\subsubsection{Long Short-Term Memory}
\label{sec:lstm}
\begin{figure}[t!]
  \centering
  \includegraphics[width=\textwidth]{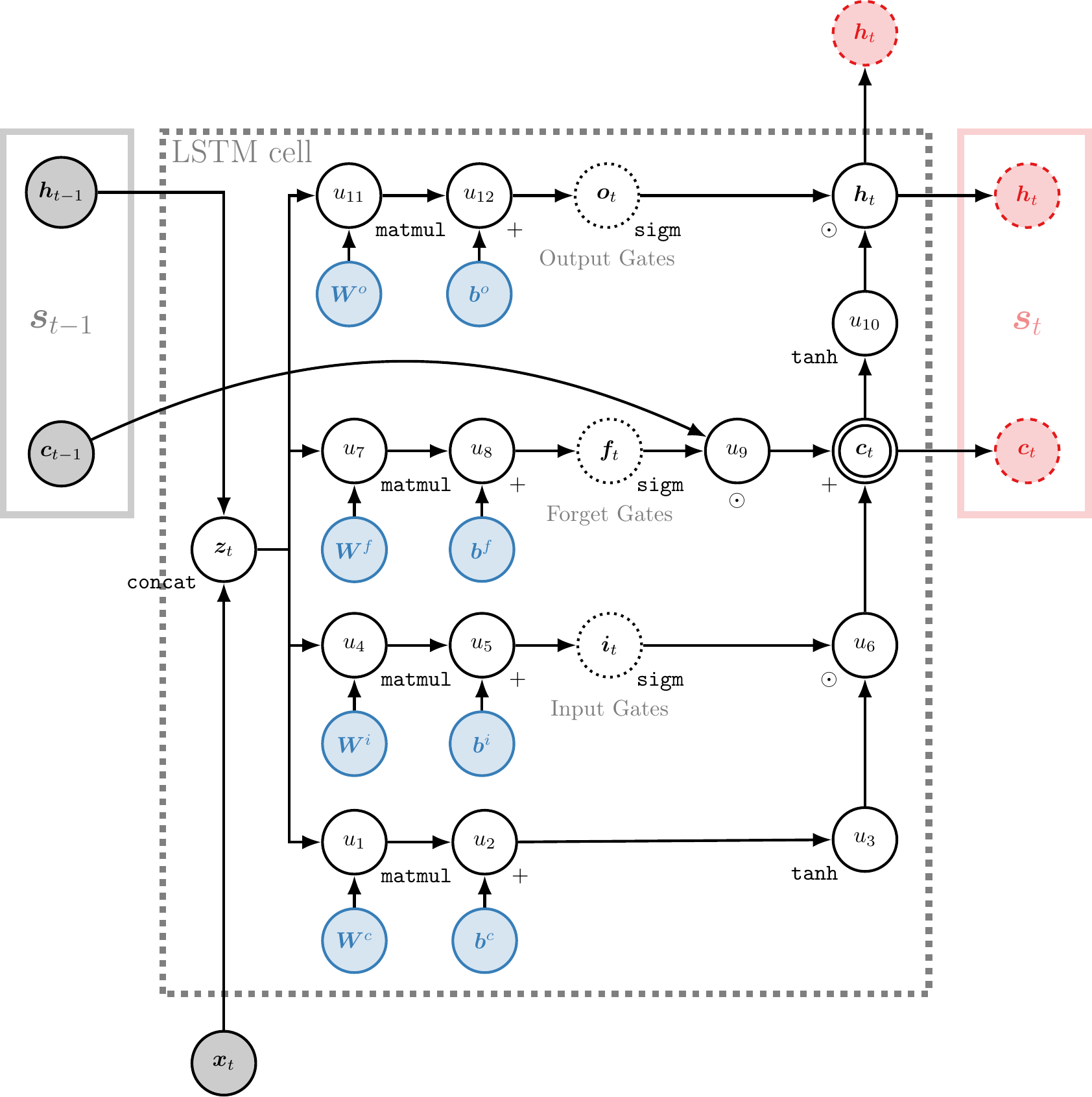}
  \caption{Computation graph for the LSTM cell.}
  \label{fig:lstm_cell}
\end{figure}

\glspl{RNN} with \gls{LSTM} units~\citep{hochreiter1997long} have been successfully applied to a wide range of \gls{NLP} tasks, such as machine translation \citep{sutskever2014sequence}, constituency parsing \citep{vinyals2014grammar}, language modeling \citep{zaremba2014recurrent} and recently \gls{RTE} \citep{bowman2015large}.
\glspl{LSTM} encompass memory cells that can store information for a long period of time, as well as three types of gates that control the flow of information into and out of these cells: \emph{input gates} (\cref{input}), \emph{forget gates} (\cref{forget}) and \emph{output gates} (\cref{output}). Given an input vector $\mathbf{x}_t$ at time step $t$, the previous output $\mathbf{h}_{t-1}$ and cell state $\mathbf{c}_{t-1}$, an LSTM with hidden size $k$ computes the next output $\mathbf{h}_t$ and cell state $\mathbf{c}_t$ as
  \begin{align}
    \vec{z}_t &= \left[{
      \begin{array}{*{20}c}
        \vec{x}_t \\
        \vec{h}_{t-1}
      \end{array} }
    \right]\\
    \vec{i}_t &= \sigma(\mat{W}^i\vec{z}_t+\vec{b}^i) \label{input}\\
    \vec{f}_t &= \sigma(\mat{W}^f\vec{z}_t+\vec{b}^f) \label{forget}\\
    \vec{o}_t &= \sigma(\mat{W}^o\vec{z}_t+\vec{b}^o) \label{output}\\
    \vec{c}_t &= \vec{f}_t \odot \vec{c}_{t-1} + \vec{i}_t \odot
    \tanh(\mat{W}^c\vec{z}_t+\vec{b}^c)\\
    \vec{h}_t &= \vec{o}_t \odot \tanh(\vec{c}_t)\\
    \vec{s}_t &= \left[{
      \begin{array}{*{20}c}
        \vec{h}_t \\
        \vec{c}_t
      \end{array} }
    \right]\\
    f_\theta^{\lstm}(\vec{x}_t,\vec{s}_{t-1}) &= (\vec{h}_t, \vec{s}_t)
  \end{align}
where $\mat{W}^i,\mat{W}^f,\mat{W}^o,\mat{W}^c\in\R^{2k\times k}$ are trained matrices and $\vec{b}^i, \vec{b}^f, \vec{b}^o, \vec{b}^c \in \R^k$ trained biases that parameterize the gates and transformations of the input.
As in previous chapters, $\sigma$ denotes the element-wise application of the sigmoid function and $\odot$ the element-wise multiplication of two vectors.
The corresponding computation graph is illustrated in \cref{fig:lstm_cell}.

\subsection{Independent Sentence Encoding}
\label{sec:independent}

\begin{figure*}[t]
\centering
\includegraphics[width=0.85\textwidth]{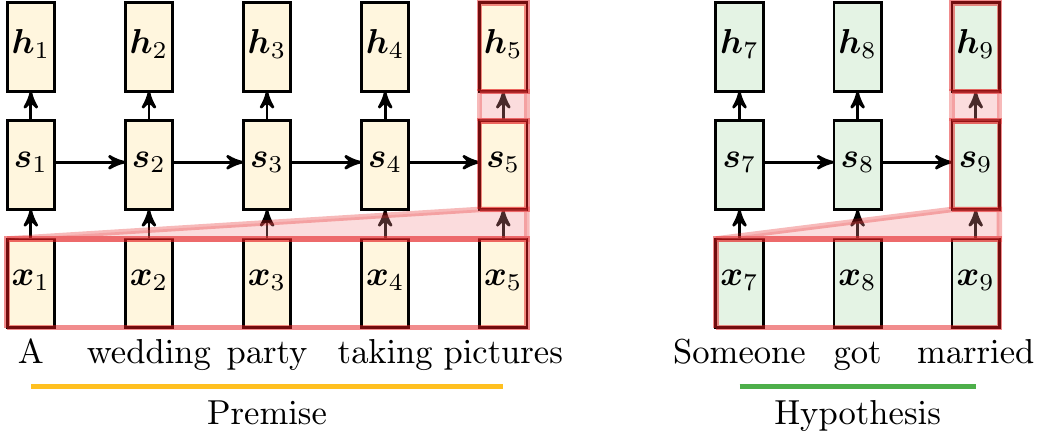}
\caption{Independent encoding of the premise and hypothesis using the same LSTM. Note that both sentences are compressed as dense vectors as indicated by the red mapping.}
\label{fig:independent}
\end{figure*}

\glspl{LSTM} can readily be used for \gls{RTE} by independently encoding the premise and hypothesis as dense vectors and taking their concatenation as input to an \gls{MLP} classifier \citep{bowman2015large}. 

Let $\ls{x}^P = [\vec{x}^P_1,\ldots,\vec{x}^P_{N}]$ be a sequence of words representing the premise and let $\ls{x}^H = [\vec{x}^H_1,\ldots,\vec{x}^H_{M}]$ represent the hypothesis.
Both, the premise and the hypothesis, can be encoded as fixed-dimensional vectors by taking the last output vector when applying the \gls{RNN} function (\cref{eq:rnn}) with an \gls{LSTM} cell function $f^{\lstm}_\theta$. 
Subsequently, the prediction for the three \gls{RTE} classes is obtained by an \gls{MLP} (\cref{eq:rte_mlp,eq:rte_mlp2}) followed by a softmax (\cref{eq:rte_softmax}):
\begin{align}
   \ls{h}^P, \ls{s}^P &= \rnn(f^{\lstm}_\theta, \ls{x}^P, \vec{s}_0)\label{eq:rte_bowman_p}\\
   \ls{h}^H, \ls{s}^H &= \rnn(f^{\lstm}_\theta, \ls{x}^H, \vec{s}_0)\label{eq:rte_bowman_h}\\
   \vec{h} &= \tanh\left(\mat{W}_2\tanh\left(\mat{W}_1\left[{
      \begin{array}{*{20}c}
        \vec{h}^P_N \\[0.25em]
        \vec{h}^H_M
      \end{array} }
    \right]+\vec{b}_1\right)+\vec{b}_2\right)\label{eq:rte_mlp}\\
  \vec{h}^* &= \tanh\left(\mat{W}_3\vec{h}+\vec{b}_3\right)\label{eq:rte_mlp2}\\
  \vec{y}_i &= \softmax(\vec{h}^*)_i = \frac{e^{\vec{h}^*_i}}{\sum_je^{\vec{h}^*_j}}\label{eq:rte_softmax} 
\end{align}
where $\mat{W}_1,\mat{W}_2\in\R^{2k\times 2k},\mat{W}_3 \in\R^{2k \times 3},\vec{b}_1,\vec{b}_2\in\R^{2k}$ and $\vec{b}_3\in\R^3$.
Furthermore, $\vec{s}_0$ is a trainable start state and $\vec{h}^P_N$ denotes the last element from the list of premise output vectors (similarly for $\vec{h}^H_M$).
The independent sentence encoding (\cref{eq:rte_bowman_p,eq:rte_bowman_h}) is visualized in \cref{fig:independent}.

Given the predicted output distribution over the three \gls{RTE} classes $\vec{y}$  (\textsc{Entailment}, \textsc{Neutral} or \textsc{Contradiction}) and a target one-hot vector $\vec{\hat y}$ encoding the correct class, a cross-entropy loss is commonly used:
\begin{align}
  \label{eq:rte_loss}
  \loss(\vec{y},\vec{\hat y}) = -\sum_i \vec{\hat y}_i\log(\vec{y}_i).
\end{align}

Independent sentence encoding is a straightforward model for \gls{RTE}.
However, it is questionable how efficiently an entire sentence can be represented in a single fixed-dimensional vector.
Hence, in the next section, we investigate various neural architectures that are tailored towards more fine-grained comparison of the premise and hypothesis and thus do not require to represent entire sentences as fixed-sized vectors in an embedding space.

\section{Methods}
First, we propose to encode the hypothesis conditioned on a representation of the premise (\cref{sec:cond}).
Subsequently, we introduce an extension of an \gls{LSTM} for \gls{RTE} with neural attention (\cref{sec:att}) and word-by-word attention (\cref{sec:iatt}).
Finally, we show how such attentive models can easily be used for attending both ways: over the premise conditioned on the hypothesis and over the hypothesis conditioned on the premise (\cref{sec:two}).
All these models are trained using the loss in \cref{eq:rte_loss}, and predict the probability of the \gls{RTE} class using \cref{eq:rte_softmax}, but they differ in the way $\vec{h}^*$ (\cref{eq:rte_mlp2}) is calculated.

\subsection{Conditional Encoding}
\label{sec:cond}

\begin{figure*}[t]
\centering
\includegraphics[width=0.85\textwidth]{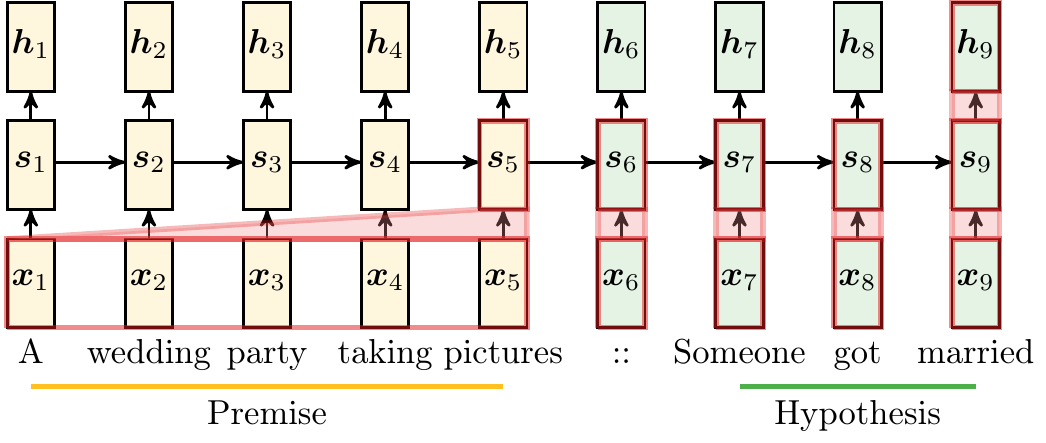}
\caption{Conditional encoding with two LSTMs. The first LSTM encodes the premise, and then the second LSTM processes the hypothesis conditioned on the representation of the premise ($\vec{s}_5$).}
\label{fig:conditional}
\end{figure*}

In contrast to learning sentence representations, we are interested in neural models that read both sentences to determine entailment, thereby comparing pairs of words and phrases.
Figure \ref{fig:conditional} shows the high-level structure of this model. The premise (left) is read by an LSTM.
A second LSTM with different parameters is reading a delimiter and the hypothesis (right), but its memory state is initialized with the last state of the previous LSTM ($\vec{s}_5$ in the example).
That is, it processes the hypothesis conditioned on the representation that the first LSTM built for the premise.
Formally, we replace \crefrange{eq:rte_bowman_p}{eq:rte_mlp} with
\begin{align}
   \ls{h}^P, \ls{s}^P &= \rnn(f^{\lstm}_{\theta_P}, \ls{x}^P, \vec{s}_0)\label{eq:rte_bowman_p_cond}\\
   \ls{h}^H, \ls{s}^H &= \rnn(f^{\lstm}_{\theta_H}, \ls{x}^H, \vec{s}^P_N)\label{eq:rte_bowman_h_cond}\\
   \vec{h} &= \vec{h}^H_M\label{eq:rte_h}
\end{align}
where $\vec{s}^P_N$ denotes the last state of the LSTM that encoded the premise and $\vec{h}^P_M$ denotes the last element from the list of hypothesis output vectors.

This model is illustrated for an \gls{RTE} example in \cref{fig:conditional}.
Note that while the premise still has to be encoded in a fixed-dimensional vector, the \gls{LSTM} that processes the hypothesis has to only keep track of whether incoming words contradict the premise, whether they are entailed by it, or whether they are unrelated.
This is inspired by finite-state automata proposed for natural logic inference \citep{angeli2014naturalli}.

\subsection{Attention}
\label{sec:att}

\begin{figure*}[t]
\centering
\includegraphics[width=1.0\textwidth]{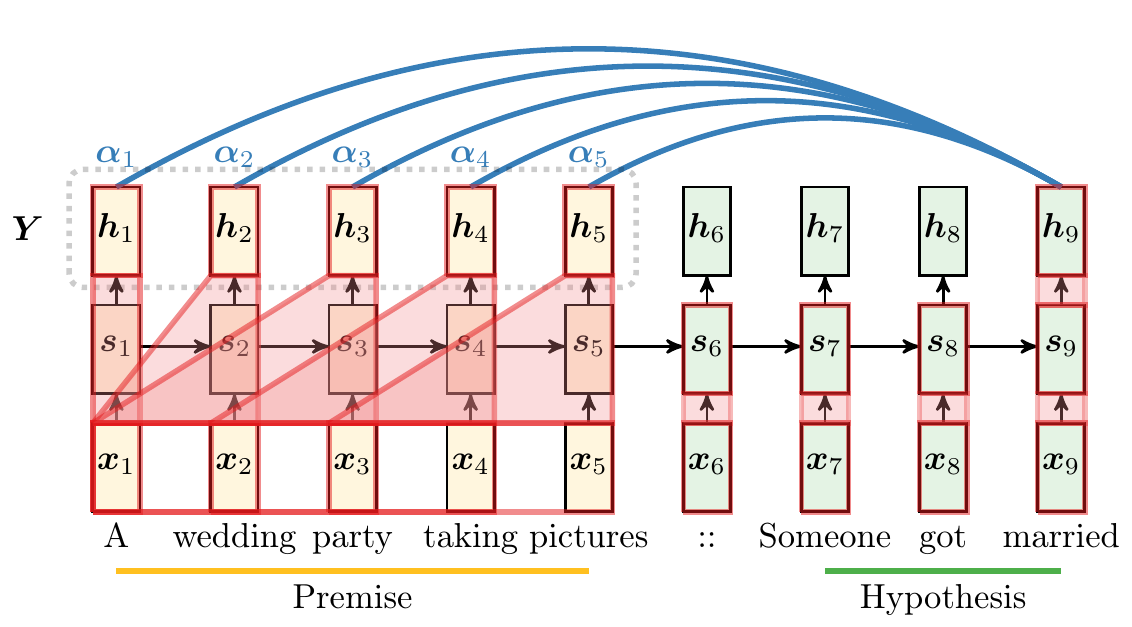}
\caption{Attention model for RTE. Compared to \cref{fig:conditional}, this model does not have to represent the entire premise in its cell state, but can instead output context representations (informally visualized by the red mapping) that are later queried by the attention mechanism (blue). Also note that now $\vec{h}_1$ to $\vec{h}_5$ are used.}
\label{fig:attention}
\end{figure*}

Attentive neural networks have recently demonstrated success in a wide range of tasks ranging from handwriting synthesis \citep{graves2013generating}, digit classification \citep{mnih2014recurrent}, machine translation \citep{bahdanau2014neural}, image captioning \citep{xu2015show}, speech recognition \citep{chorowski2015attention}, sentence summarization \citep{rush2015neural}, and code summarization \citep{alamanis2016convolutional} to geometric reasoning \citep{vinyals2015pointer}.
The idea is to allow the model to attend over past output vectors.
For \glspl{LSTM} this mitigates the cell state bottleneck, \ie{}, the fact that a standard \gls{LSTM} has to store all relevant information for future time steps in its internal memory $\vec{c}_t$ (see $\vec{c}_t$ in \cref{fig:lstm_cell} and compare \cref{fig:conditional} with \cref{fig:attention}).

An \gls{LSTM} with attention for \gls{RTE} does not have to capture the entire content of the premise in its cell state.
Instead, it is sufficient to output vectors while reading the premise (\ie{} populating a differentiable memory of the premise) and accumulating a representation in the cell state that informs the second \gls{LSTM} which of the output vectors of the premise it needs to attend over to determine the \gls{RTE} class.

Formally, let $\mat{Y} \in \R^{k\times N}$ be a matrix consisting of output vectors $[\vec{h}_1 \cdots \vec{h}_N]$ that the first \gls{LSTM} produced when reading the $N$ words of the premise.
Furthermore, let $\vec{1} \in \R^N$ be a vector of ones and $\vec{h}^H_M$ be the last output vector after the premise and hypothesis were processed by the two \glspl{LSTM}.
The attention mechanism will produce a vector $\vec{\alpha} \in \R^N$ of attention weights and a weighted representation $\vec{r}$ of the premise via
\begin{align}
  \mat{M} &= \tanh\left(\mat{W}^y\mat{Y}+\left(\mat{W}^h\vec{h}^H_M\right)\vec{1}^T\right)&\mat{M}&\in\R^{k \times N}\\
  \vec{\alpha} &= \softmax(\vec{w}^T\mat{M})&\vec{\alpha}&\in\R^{1\times N}\\
  \vec{r} &= \mat{Y}\vec{\alpha}^T&\vec{r}&\in\R^k
\end{align}
where $\mat{W}^y, \mat{W}^h \in \R^{k \times k}$ are trainable projection matrices, and $\vec{w} \in \R^k$ is a trainable parameter vector.
Note that the outer product $(\mat{W}^h\vec{h}^H_M)\vec{1}^\top$ is repeating the linearly transformed $\vec{h}^H_M$ as many times as there are words in the premise (\emph{i.e.} $N$ times).
Hence, the intermediate attention representation $\vec{m}_i$ ($i$th column vector in $\mat{M}$) of the $i$th word in the premise is obtained from a non-linear combination of the premise's output vector $\vec{h}_i$ ($i$th column vector in $\mat{Y}$) and the transformed $\vec{h}^H_M$.
The attention weight for the $i$th word in the premise is the result of a weighted combination (parameterized by $\vec{w}$) of values in $\vec{m}_i$.

The final sentence pair representation is obtained from a non-linear combination of the attention-weighted representation $\mathbf{r}$ of the premise and the last output vector $\mathbf{h}^H_M$, thus replacing \cref{eq:rte_h} by
\begin{align}
  \vec{h} & = \tanh(\mat{W}^p\vec{r}+\mat{W}^x\vec{h}^H_M)\label{eq:att}
\end{align}
where $\mat{W}^p, \mat{W}^x \in \R^{k\times k}$ are trainable projection matrices.

The attention model is illustrated in \cref{fig:attention}. 
Note that this model does not have to represent the entire premise in its cell state, but can instead output context representations that are later queried by the attention mechanism. 
This is informally illustrated by the red mapping of input phrases to output context representations.\footnote{Note that the number of word representations that are used to output a context representation is not known. For illustration purposes we depicted the case that information of three words contribute to an output context representation.}

\subsection{Word-by-word Attention}
\label{sec:iatt}

\begin{figure*}[t]
\centering
\includegraphics[width=1.0\textwidth]{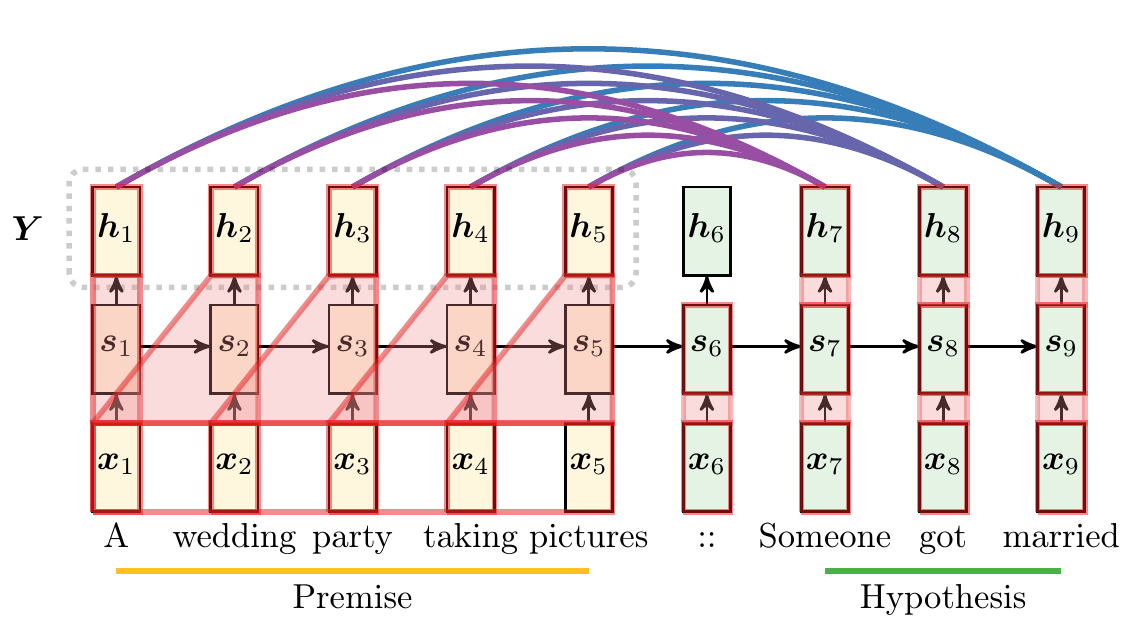}
\caption{Word-by-word attention model for RTE. Compared to \cref{fig:attention}, querying the memory $\mat{Y}$ multiple times allows the model to store more fine-grained information in its output vectors when processing the premise. Also note that now also $\vec{h}_6$ to $\vec{h}_8$ are used.}
\label{fig:wordbyword}
\end{figure*}

For determining whether one sentence entails another it is desirable to check for entailment or contradiction of individual word and phrase pairs.
To encourage such behavior we employ neural word-by-word attention similar to \cite{bahdanau2014neural}, \cite{hermann2015teaching} and \cite{rush2015neural}.
The difference is that we do not use attention to generate words, but to obtain a sentence pair encoding from fine-grained comparison via soft-alignment of word and phrase pairs in the premise and hypothesis.
In our case, this amounts to attending over the first \gls{LSTM}'s output vectors for the premise while the second \gls{LSTM} processes the hypothesis one word at a time.
Consequently, we obtain attention weight-vectors $\vec{\alpha}_t$ over premise output vectors for every word in the hypothesis.
This can be modeled as follows:
\begin{align}
  \mat{M}_t &= \tanh\left(\mat{W}^y\mat{Y}+\left(\mat{W}^h\vec{h}^H_t + \mat{W}^r\vec{r}_{t-1}\right)\vec{1}^T\right)&\mat{M}_t&\in\R^{k \times N}\label{eq:att-rec}\\
  \vec{\alpha}_t &= \softmax\left(\vec{w}^T\mat{M}_t\right)&\vec{\alpha}_t&\in\R^{1\times N}\\
  \vec{r}_t &= \mat{Y}\vec{\alpha}^T_t + \tanh\left(\mat{W}^t\vec{r}_{t-1}\right)&\vec{r}_t&\in\R^k\label{eq:att-final}
\end{align}
where $\mat{W}^r,\mat{W}^t\in\R^{k\times k}$ are trainable projection matrices.
Note that $\mathbf{r}_t$ is dependent on the previous attention representation $\mathbf{r}_{t-1}$ to inform the model about what was attended over in previous steps (see \cref{eq:att-rec,eq:att-final}).

As in the previous section, the final sentence pair representation is obtained from a non-linear combination of the last attention-weighted representation of the premise (here conditioned on the last word of the hypothesis) $\mathbf{r}_N$ and the last output vector using
\begin{align}
  \vec{h} & = \tanh(\mat{W}^p\vec{r}_M+\vec{W}^x\vec{h}^H_M).\label{eq:iatt}
\end{align}

The word-by-word attention model is illustrated in \cref{fig:wordbyword}. 
Compared to the attention model introduced earlier, querying the memory $\mat{Y}$ multiple times allows the model to store more fine-grained information in its output vectors when processing the premise. 
We informally illustrate this by fewer words contributing to the output context representations (red mapping).

\subsection{Two-way Attention}
\label{sec:two}
Inspired by bidirectional LSTMs that read a sequence and its reverse for improved encoding \citep{graves2005framewise}, we experiment with two-way attention for RTE.
The idea is to use the same model (\emph{i.e.} same structure and weights) to attend over the premise conditioned on the hypothesis, as well as to attend over the hypothesis conditioned on the premise, by simply swapping the two sequences.
This produces two sentence pair representations that we concatenate for classification.

\section{Experiments}
We conduct experiments on the Stanford Natural Language Inference corpus \cite[SNLI,][]{bowman2015large}.
This corpus is two orders of magnitude larger than other existing RTE corpora such as Sentences Involving Compositional Knowledge \citep[SICK,][]{marelli2014semeval}.
Furthermore, a large part of training examples in SICK were generated heuristically from other examples. In contrast, all sentence pairs in SNLI stem from human annotators.
The size and quality of SNLI make it a suitable resource for training neural architectures such as the ones proposed in this chapter.

\subsection{Training Details}
We use pretrained word2vec vectors \citep{mikolov2013distributed} as word representations, which we keep fixed during training.
Out-of-vocabulary words in the training set are randomly initialized by uniformly sampling values from $[-0.05,0.05]$ and are optimized during training.\footnote{We found $12.1$k words in SNLI for which we could not obtain word2vec embeddings, resulting in $3.65$M tunable parameters.}
Out-of-vocabulary words encountered at inference time on the validation and test corpus are set to fixed random vectors.
By not tuning representations of words for which we have word2vec vectors, we ensure that at test time their representation stays close to unseen similar words that are contained in word2vec.\todo{example: mouse and rat}

We use ADAM \citep{kingma2014adam} for optimization with a first momentum coefficient of $0.9$ and a second momentum coefficient of $0.999$.\footnote{Standard configuration recommended by \citeauthor{kingma2014adam}.}
For every model we perform a grid search over combinations of the initial learning rate [1\textsc{e}-4, 3\textsc{e}-4, 1\textsc{e}-3], dropout\footnote{As in \cite{zaremba2014recurrent}, we apply dropout only on the inputs and outputs of the network.} [0.0, 0.1, 0.2] and $\ell_2$-regularization strength [0.0, 1\textsc{e}-4, 3\textsc{e}-4, 1\textsc{e}-3].
Subsequently, we take the best configuration based on performance on the validation set, and evaluate only that configuration on the test set.

\section{Results and Discussion}

\begin{table*}
  \caption{Results on the SNLI corpus.}
  \label{tab:results}
  \centering
\resizebox{1.0\textwidth}{!}{
  \begin{tabular}{lllllll}
  	\toprule
  	Model & $k$ & $|\theta|_\text{W+M}$ & $|\theta|_\text{M}$ & Train & Dev & Test\\
  	\midrule
  	Lexicalized classifier \citep{bowman2015large} & - & - & - &  99.7 & - & 78.2\\
  	LSTM \citep{bowman2015large} & 100 & $\approx 10$M & $221$k & 84.4 & - & 77.6\\
  	\midrule
  	Conditional encoding, shared & 100 & $3.8$M & $111$k & 83.7 & 81.9 & 80.9\\
    Conditional encoding, shared & 159 & $3.9$M & $252$k & 84.4 & 83.0 & 81.4\\
    Conditional encoding & 116 & $3.9$M & $252$k & 83.5 & 82.1 & 80.9\\
  	\midrule
  	Attention & 100 & $3.9$M & $242$k & 85.4 & 83.2 & 82.3\\
    Attention, two-way & 100 & $3.9$M & $242$k & 86.5 & 83.0 & 82.4 \\
  	\midrule
  	Word-by-word attention & 100 & $3.9$M & $252$k & 85.3 & \textbf{83.7} & \textbf{83.5}\\
    Word-by-word attention, two-way & 100 & $3.9$M & $252$k & 86.6 & 83.6 & 83.2 \\
  	\bottomrule
  \end{tabular}
}
\end{table*}

Results on the SNLI corpus are summarized in \cref{tab:results}.
The total number of model parameters, including tunable word representations, is denoted by $|\theta|_\text{W+M}$ (without word representations $|\theta|_\text{M}$).
To ensure a comparable number of parameters to \citeauthor{bowman2015large}'s model that encodes the premise and hypothesis independently using one LSTM, we also run experiments for conditional encoding where the parameters between both LSTMs are shared (``Conditional encoding, shared'' with $k=100$), as opposed to using two independent LSTMs.
In addition, we compare our attentive models to two benchmark LSTMs whose hidden sizes were chosen so that they have at least as many parameters as the attentive models ($k$ set to $159$ and $116$ respectively).
Since we are not tuning word vectors for which we have word2vec embeddings, the total number of parameters $|\theta|_\text{W+M}$ of our models is considerably smaller.
We also compare our models against the benchmark lexicalized classifier used by \citeauthor{bowman2015large}, which uses features based on the BLEU score between the premise and hypothesis, length difference, word overlap, uni- and bigrams, part-of-speech tags, as well as cross uni- and bigrams.

\paragraph{Conditional Encoding}
We found that processing the hypothesis conditioned on the premise instead of encoding both sentences independently gives an improvement of $3.3$ percentage points in accuracy over \citeauthor{bowman2015large}'s LSTM.
We argue this is due to information being able to flow from the part of the model that processes the premise to the part that processes the hypothesis.
Specifically, the model does not waste capacity on encoding the hypothesis (in fact it does not need to encode the hypothesis at all), but can read the hypothesis in a more focused way by checking words and phrases for contradiction or entailment based on the semantic representation of the premise (see \cref{fig:conditional}).
One interpretation is that the LSTM is approximating a finite-state automaton for RTE \cite[\emph{c.f.}][]{angeli2014naturalli}.
Another difference to \citeauthor{bowman2015large}'s model is that we are using word2vec instead of GloVe for word representations and, more importantly, do not fine-tune these word embeddings.
The drop in accuracy from the train to the test set is less severe for our models, which suggest that fine-tuning word embeddings could be a cause of overfitting.

Our \gls{LSTM} outperforms a simple lexicalized classifier by $2.7$ percentage points.
To the best of our knowledge, at the time of publication this was the first instance of a neural end-to-end differentiable model outperforming a hand-crafted \gls{NLP} pipeline on a textual entailment dataset.

\paragraph{Attention} By incorporating an attention mechanism we observe a $0.9$ percentage point improvement over a single \gls{LSTM} with a hidden size of $159$ and a $1.4$ percentage point increase over a benchmark model that uses two \glspl{LSTM} for conditional encoding (one for the premise and one for the hypothesis conditioned on the representation of the premise).
The attention model produces output vectors summarizing contextual information of the premise that is useful to attend over later when reading the hypothesis.
Therefore, when reading the premise, the model does not have to build up a semantic representation of the whole premise, but instead a representation that helps attending over the premise's output vectors when processing the hypothesis (see \cref{fig:attention}).
In contrast, the output vectors of the premise are not used by the baseline conditional model.
Thus, these models have to build up a representation of the entire premise and carry it over through the cell state to the part that processes the hypothesis---a bottleneck that can be overcome to some degree by using attention.

\paragraph{Word-by-word Attention} Enabling the model to attend over output vectors of the premise for every word in the hypothesis yields another $1.2$ percentage point improvement compared to attending only once.
We argue that this can be explained by the model being able to check for entailment or contradiction of individual word and phrase pairs, and we demonstrate this effect in the qualitative analysis below.

\paragraph{Two-way Attention}
Allowing the model to also attend over the hypothesis based on the premise did not improve performance for \gls{RTE} in our experiments.
We suspect that this is due to entailment being an asymmetric relation.
Hence, using the same \gls{LSTM} to encode the hypothesis (in one direction) and the premise (in the other direction) might lead to noise in the training signal.
This could be addressed by training different \glspl{LSTM} at the cost of doubling the number of model parameters.

\subsection{Qualitative Analysis}
\label{sec:analysis}
It is instructive to analyze which output representations the model is attending over when deciding the class of an \gls{RTE} example.
Note that interpretations based on attention weights have to be taken with care  since the model is not forced to solely rely on representations obtained from attention (see $\vec{h}^H_M$ in \cref{eq:att,eq:iatt}).
In the following, we visualize and discuss attention patterns of the presented attentive models.
For each attentive model, we hand-picked examples from ten sentence pairs that were randomly drawn from the development set.

\paragraph{Attention}
\begin{figure}[t!]
  \begin{subfigure}[t]{0.5\textwidth}
    \frame{
      \includegraphics[height=7em]{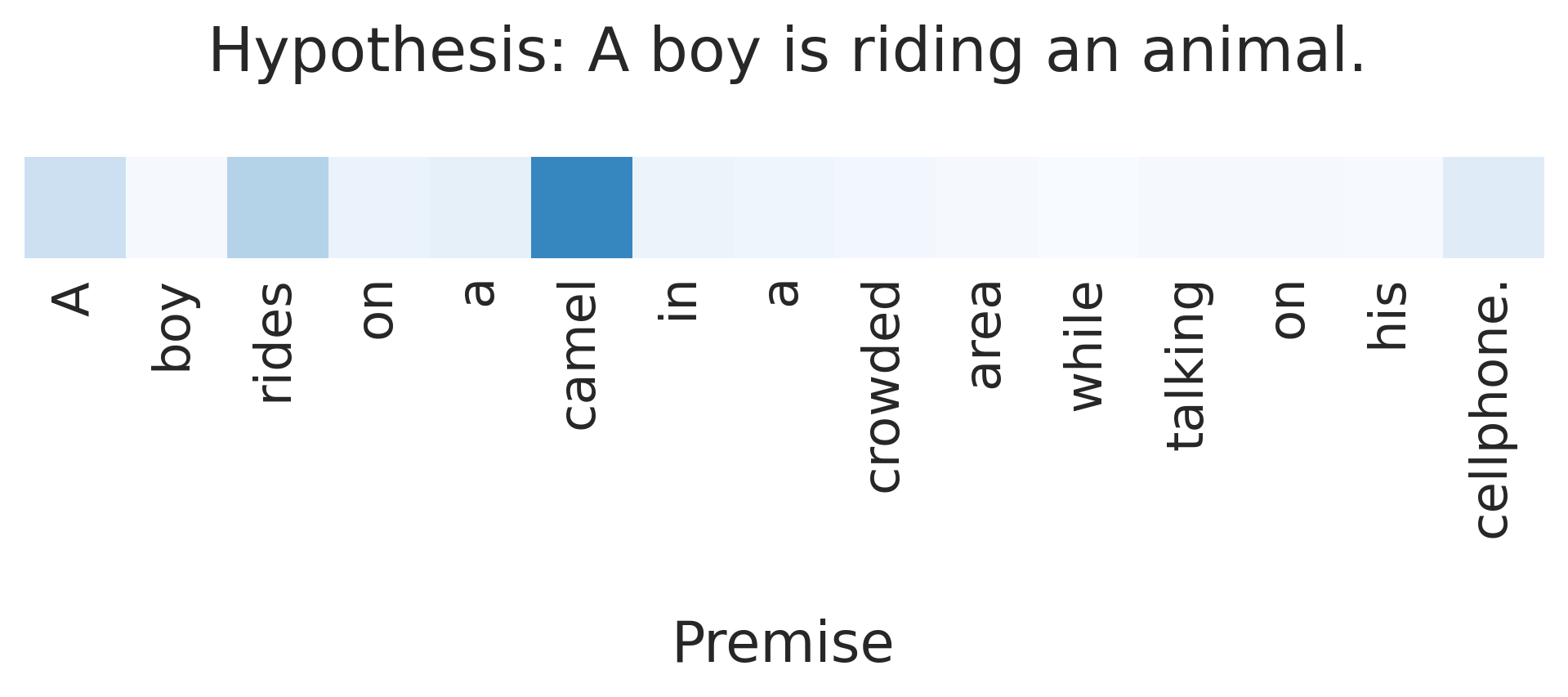}
    }
    \caption{}
    \label{fig:att:a}
  \end{subfigure}
  \begin{subfigure}[t]{0.5\textwidth}
    \hfill
    \frame{
      \includegraphics[height=7em]{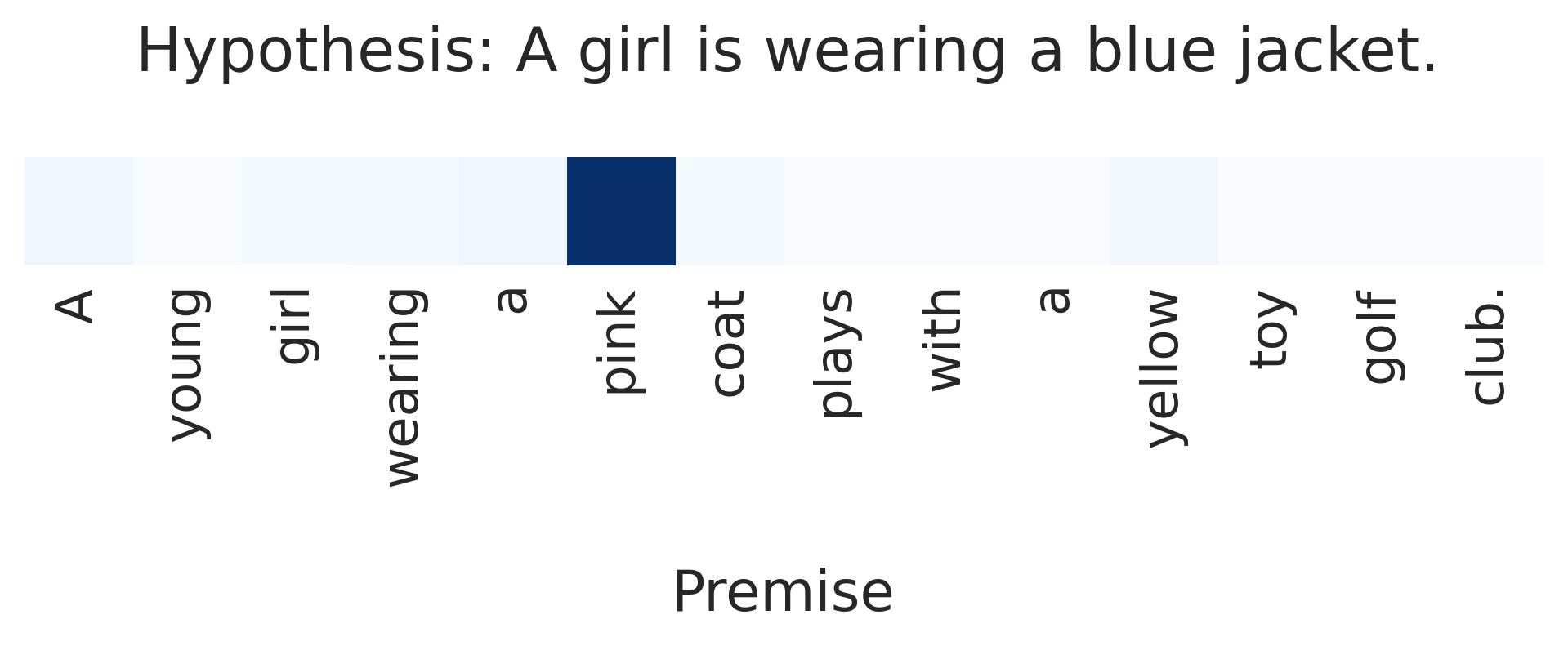}
    }
    \caption{\vspace{1.5em}}
    \label{fig:att:b}
  \end{subfigure}
  \begin{subfigure}[t]{0.47\textwidth}
    \frame{
      \includegraphics[height=7em]{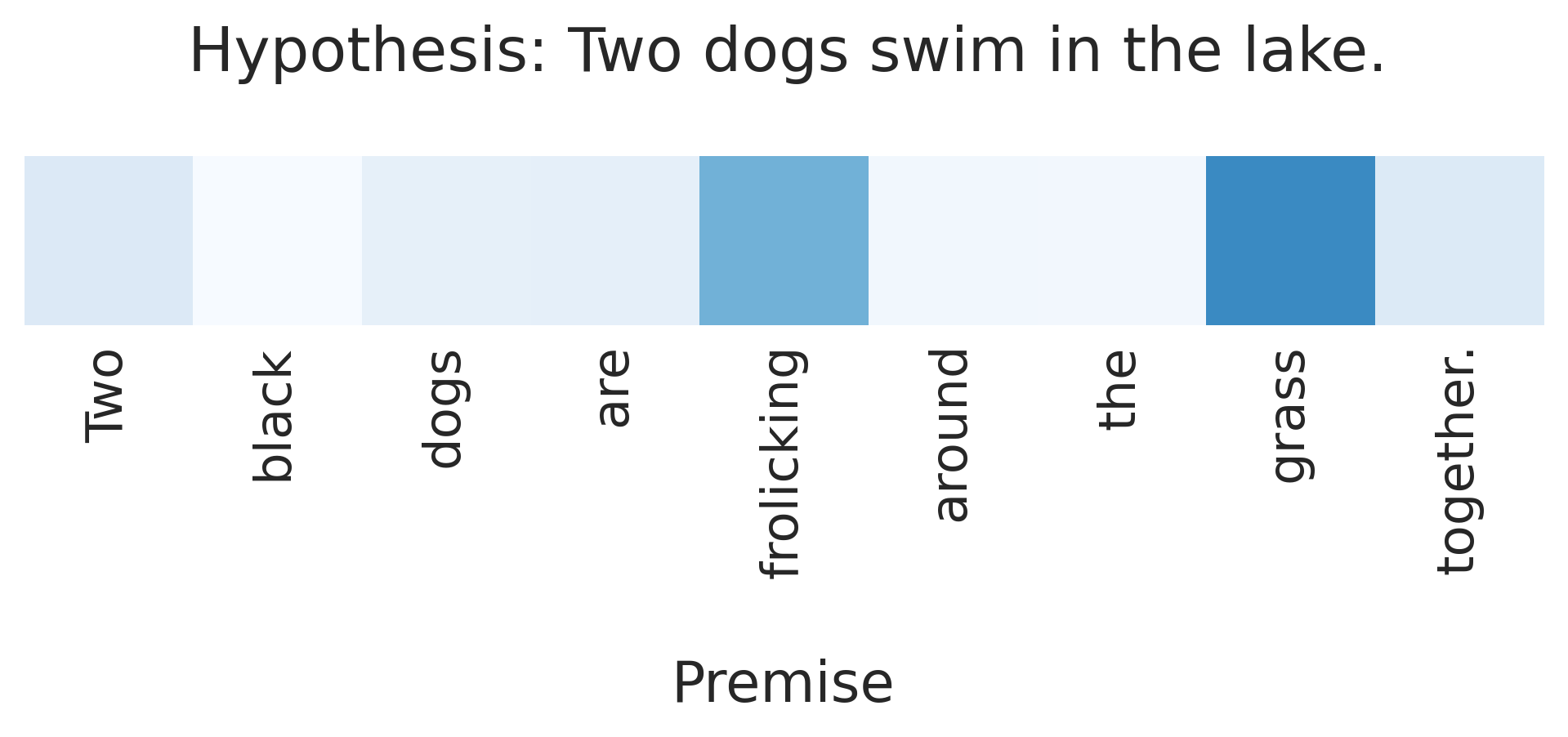}
    }
    \caption{}
    \label{fig:att:c}
  \end{subfigure}
  \begin{subfigure}[t]{0.53\textwidth}
    \hfill
    \frame{
      \includegraphics[height=7em]{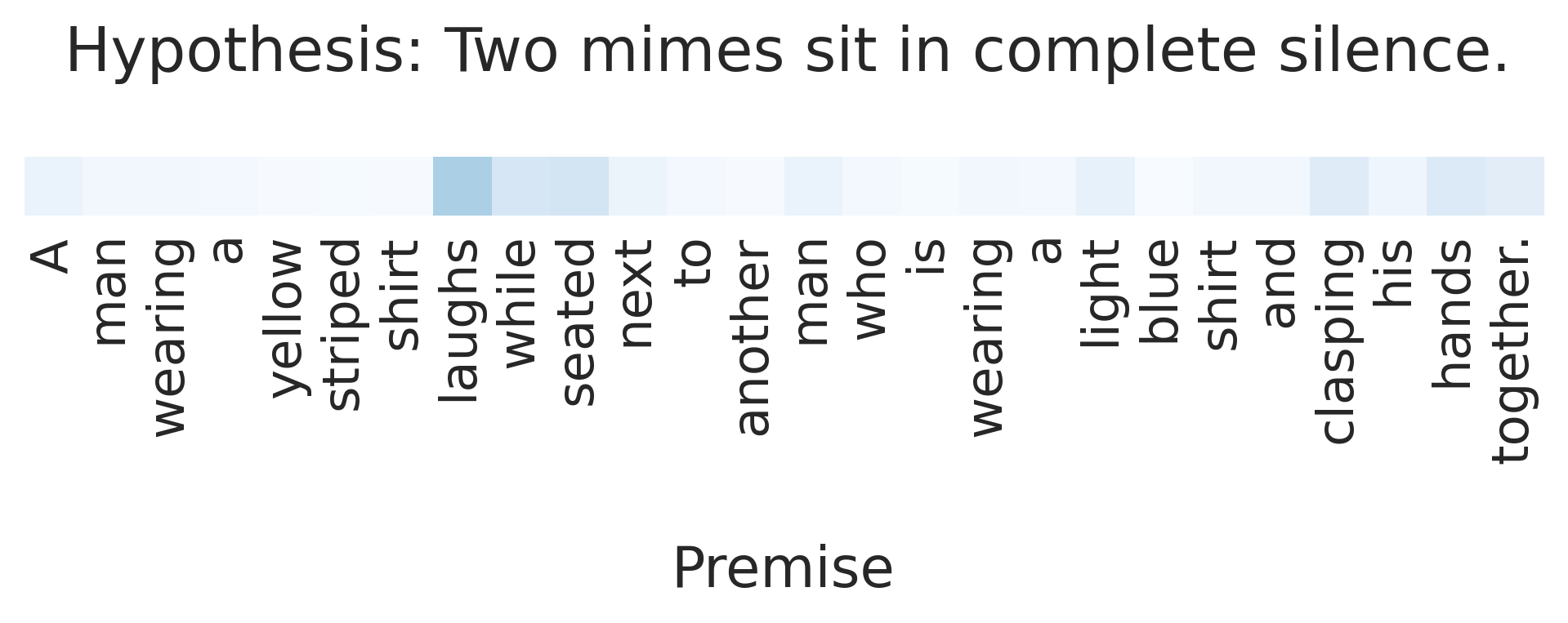}
    }
    \caption{}
    \label{fig:att:d}
  \end{subfigure}
  \caption{Attention visualizations.}
  \label{fig:att}
\end{figure}

\begin{figure}[h!]
  \begin{subfigure}[t]{0.4\textwidth}
    \frame{
      \includegraphics[height=13.25em]{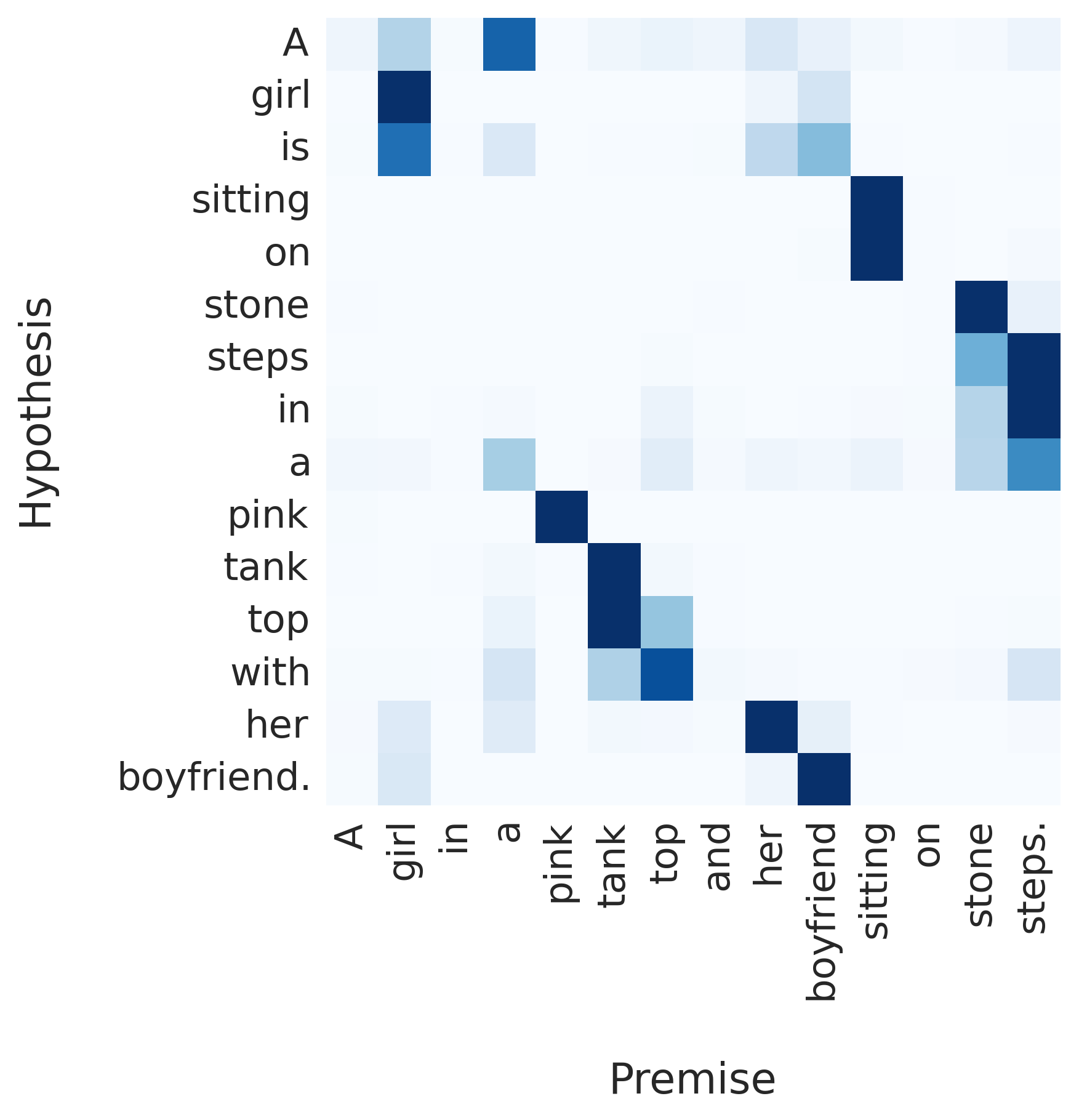}
    }
    \caption{}
    \label{fig:iatt1:a}
  \end{subfigure}
  \begin{subfigure}[t]{0.6\textwidth}
    \hfill
    \frame{
      \includegraphics[height=13.25em]{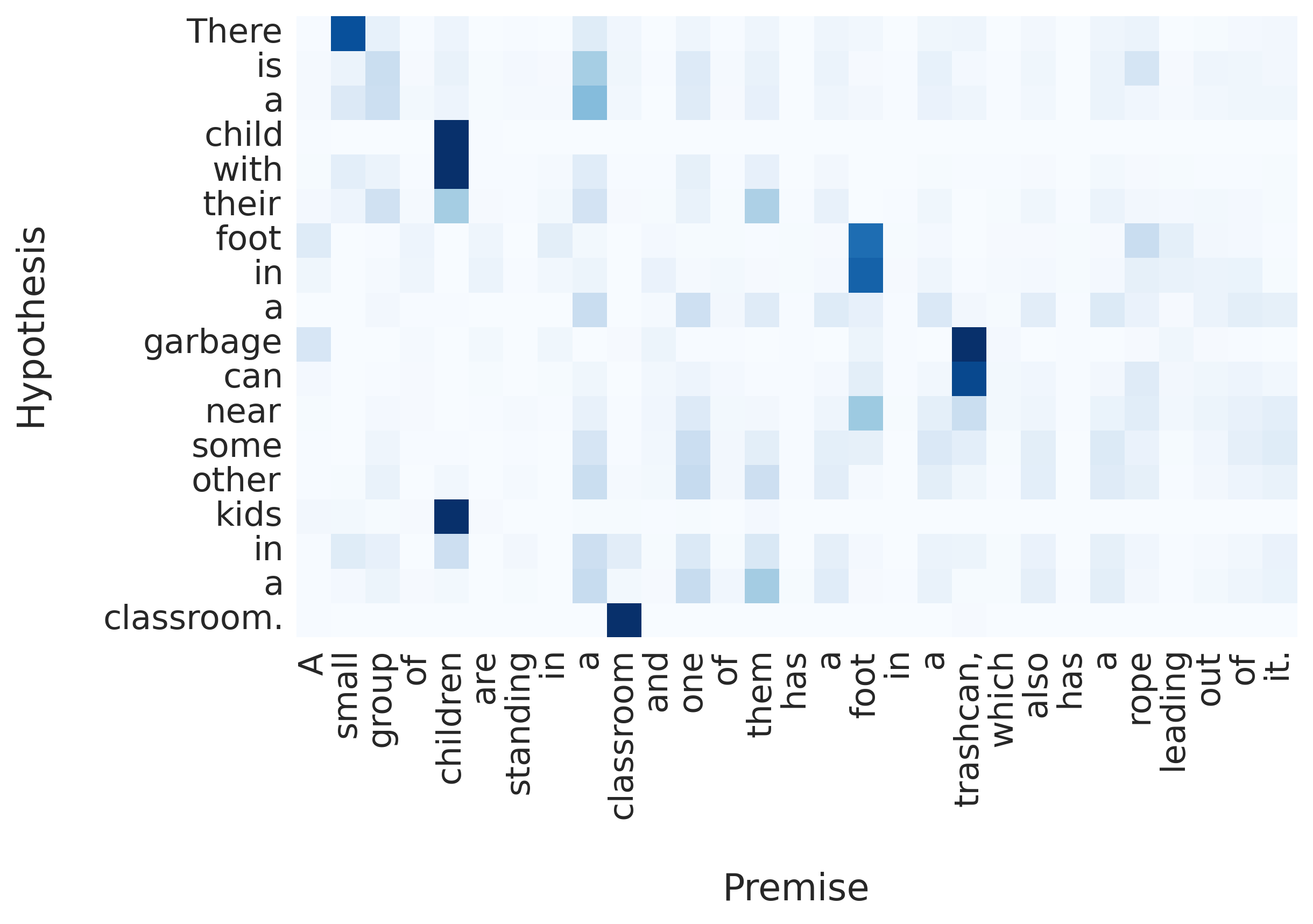}
    }
    \caption{\vspace{1.5em}}
    \label{fig:iatt1:c}
  \end{subfigure}
  \begin{subfigure}[t]{0.55\textwidth}
    \frame{
      \includegraphics[height=10.25em]{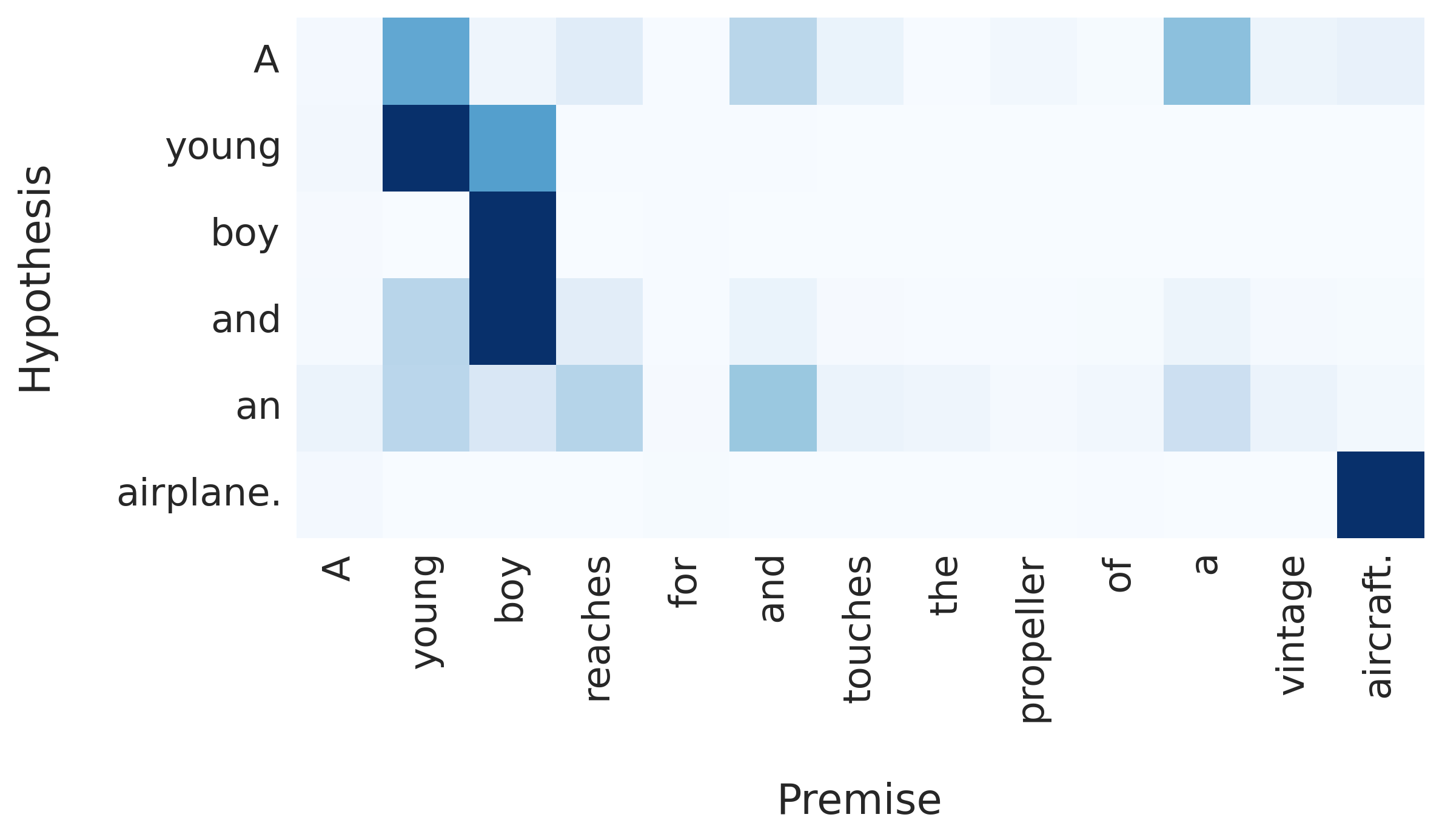}
    }
    \caption{}
    \label{fig:iatt1:b}
  \end{subfigure}
  \begin{subfigure}[t]{0.45\textwidth}
    \hfill
    \frame{
      \includegraphics[height=10.25em]{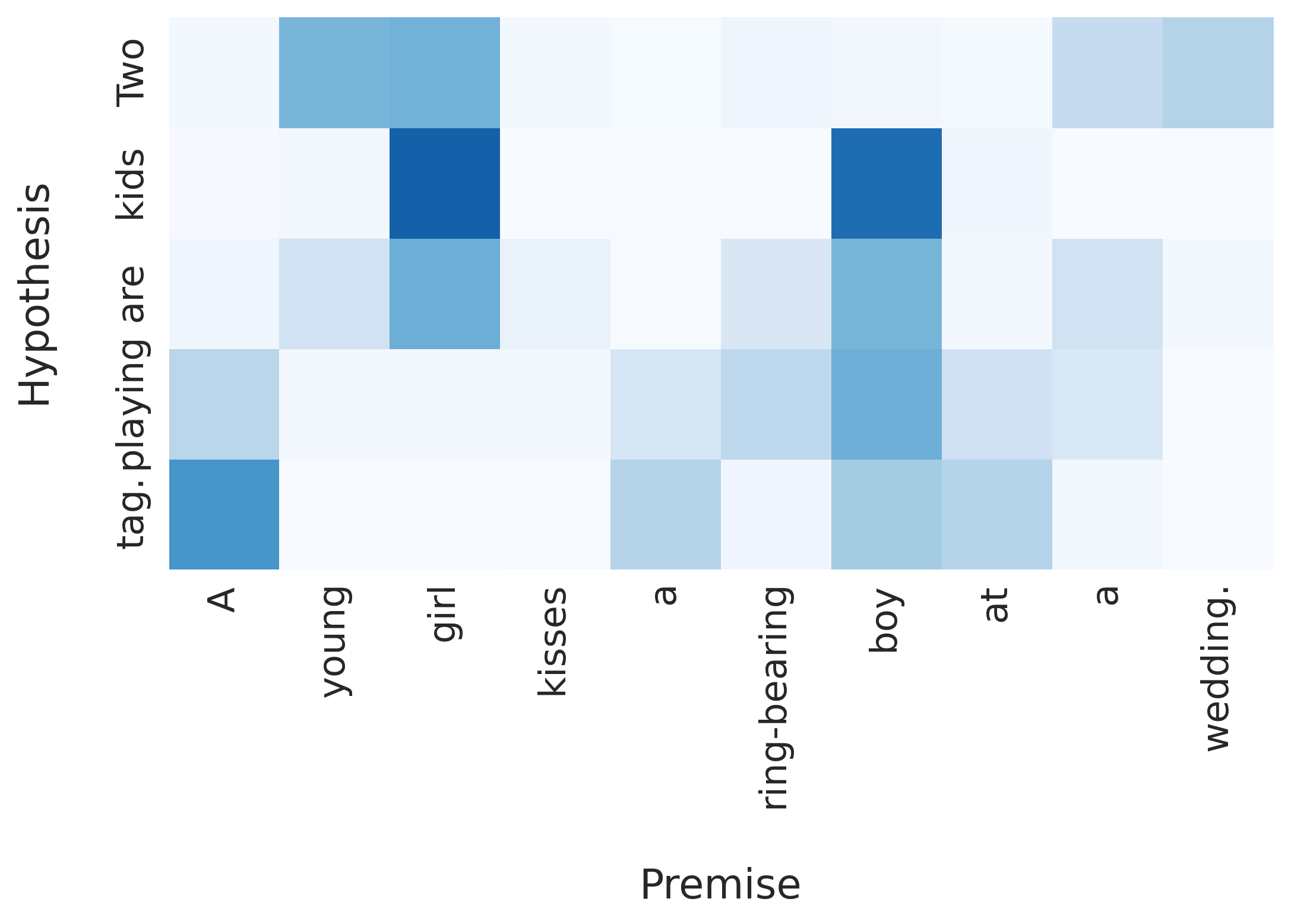}
    }
    \caption{\vspace{1.5em}}
    \label{fig:iatt2:c}
  \end{subfigure}
  \begin{subfigure}[t]{0.5\textwidth}
    \frame{
      \includegraphics[height=7.5em]{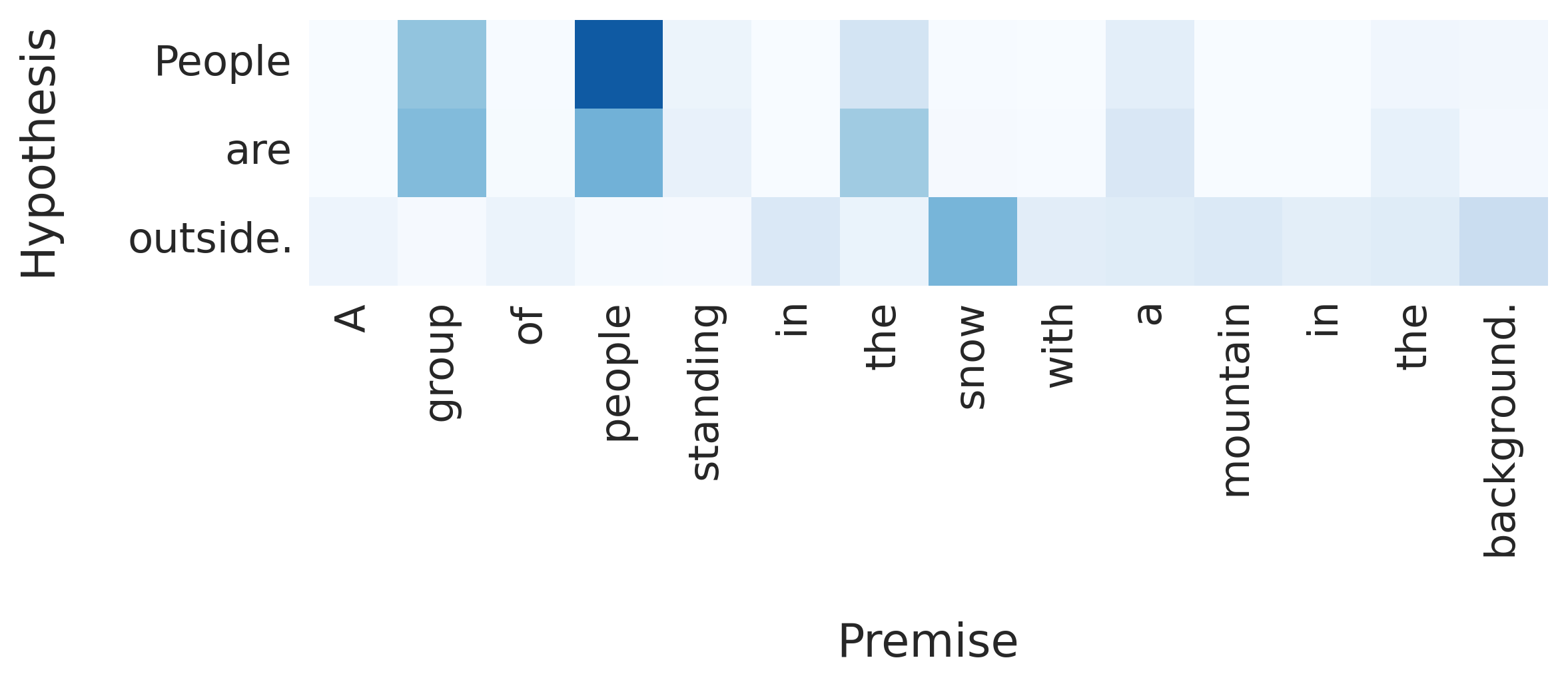}
    }
    \caption{}
    \label{fig:iatt2:a}
  \end{subfigure}
  \begin{subfigure}[t]{0.5\textwidth}
      \hfill
      \frame{
      \includegraphics[height=7.5em]{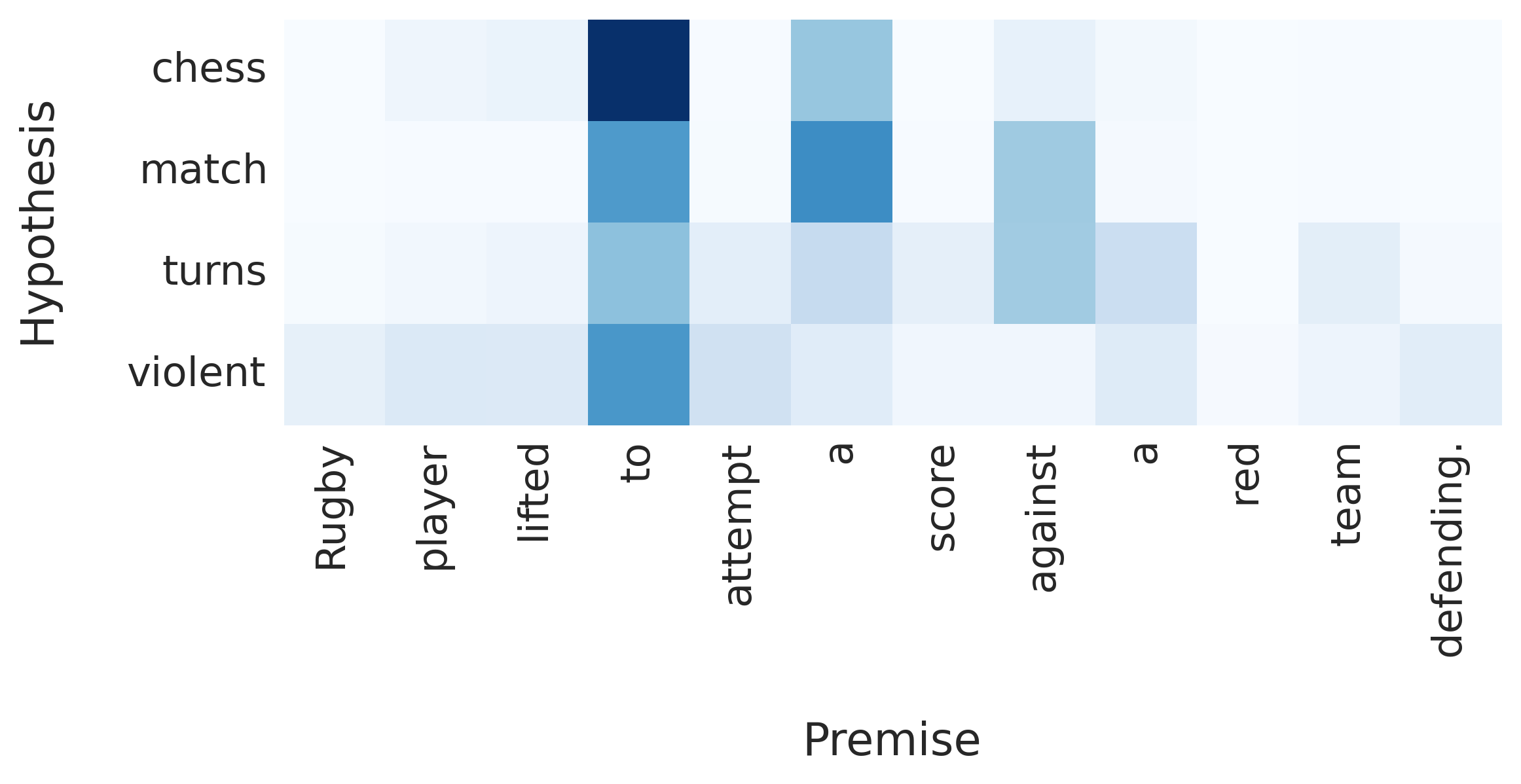}
      }
      \caption{\vspace{1.5em}}
      \label{fig:iatt3}
  \end{subfigure}
  \begin{subfigure}[t]{1.0\textwidth}
    \centering
    \frame{
      \includegraphics[height=13.75em]{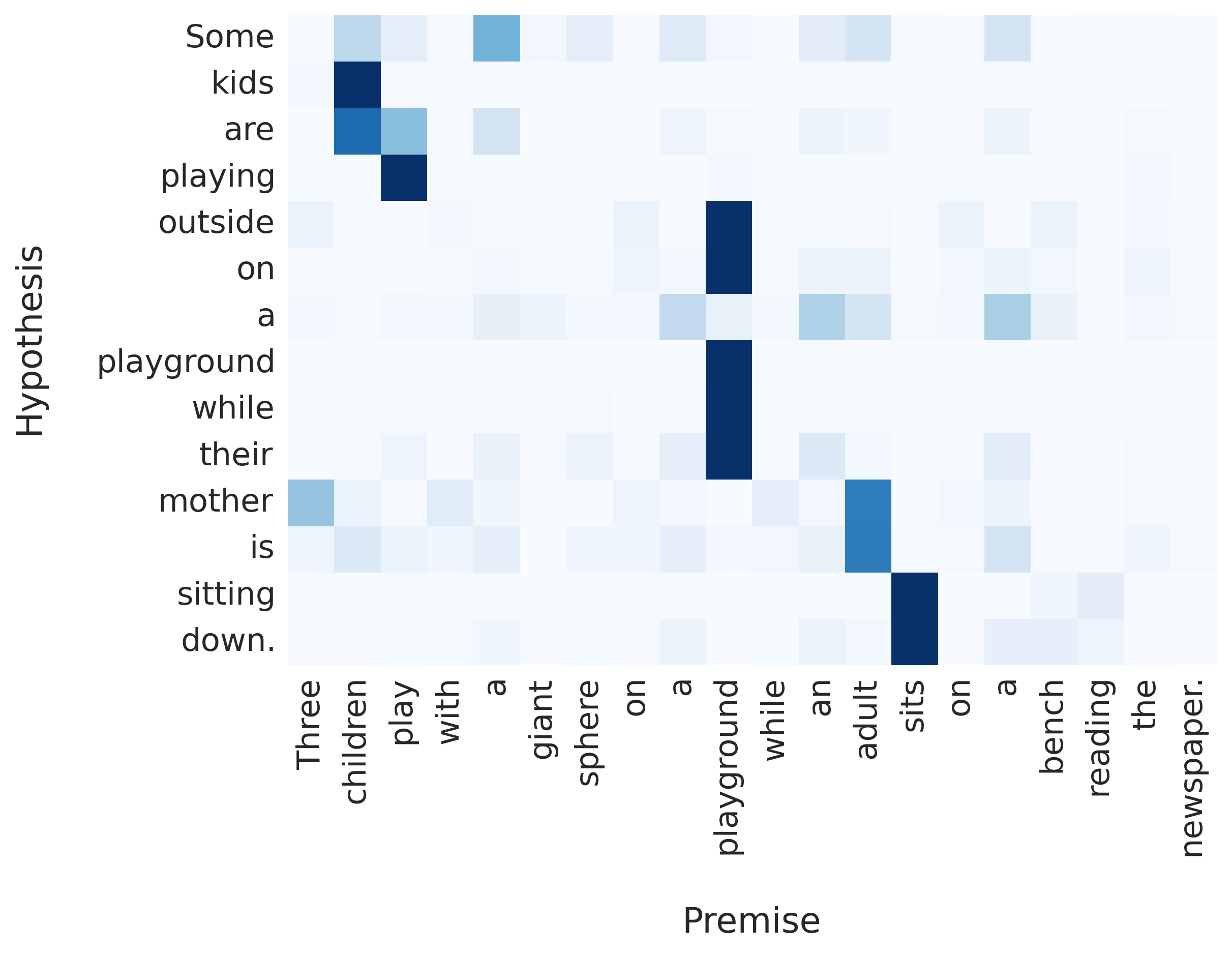}
    }
    \caption{}
    \label{fig:iatt2:b}
  \end{subfigure}
  \caption{Word-by-word attention visualizations.}
  \label{fig:iatt1}
\end{figure}

\Cref{fig:att} shows to what extent the attentive model focuses on contextual representations of the premise after both \glspl{LSTM} processed the premise and hypothesis, respectively.
Note how the model pays attention to output vectors of words that are semantically coherent with the premise (``riding'' and ``rides'', ``animal'' and ``camel'', \ref{fig:att:a}) or in contradiction, as caused by a single word (``blue'' vs. ``pink'', \ref{fig:att:b}) or multiple words (``swim'' and ``lake'' vs. ``frolicking'' and ``grass'', \ref{fig:att:c}).
Interestingly, the model shows sensitivity to context by not attending over ``yellow'', the color of the toy, but ``pink'', the color of the coat.
\todo{remove this sentence? the model should not be able to understand this, as it can't look ahead / is not bidirectional}
However, for more involved examples with longer premises, we found that attention is more uniformly distributed (\ref{fig:att:d}).
This suggests that conditioning attention only on the last output representation has limitations when multiple words need to be considered for deciding the RTE class.

\paragraph{Word-by-word Attention}
Visualizations of word-by-word attention are depicted in \cref{fig:iatt1}.
We found that word-by-word attention can easily detect if the hypothesis is simply a reordering of words in the premise (\ref{fig:iatt1:a}).
Furthermore, it is able to resolve synonyms (``airplane'' and ``aircraft'', \ref{fig:iatt1:b}) and capable of matching multi-word expressions to single words (``garbage can'' to ``trashcan'', \ref{fig:iatt1:c}).
It is also noteworthy that irrelevant parts of the premise, such as words capturing little meaning or whole uninformative relative clauses, are correctly neglected for determining entailment (``which also has a rope leading out of it'', \ref{fig:iatt1:c}).

Word-by-word attention seems to also work well when words in the premise and hypothesis are connected via deeper semantics\todo{tone down} (``snow'' can be found ``outside'' and a ``mother'' is an ``adult'', \ref{fig:iatt2:a} and \ref{fig:iatt2:b}).
Furthermore, the model is able to resolve one-to-many relationships (``kids'' to ``boy'' and ``girl'', \ref{fig:iatt2:c}).

Attention can fail, for example when the two sentences and their words are entirely unrelated (\ref{fig:iatt3}).
In such cases, the model seems to back off to attending over function words, and the sentence pair representation is likely dominated by the last output vector ($\vec{h}^H_M$) instead of the attention-weighted representation (see \cref{eq:iatt}).

\section{Related Work}
The methods in this chapter were published in \cite{rocktaschel2016reasoning} and since then many new models have been proposed.
They can be roughly classified into sentence encoding models which extend the independent encoding \gls{LSTM} by \cite{bowman2015large} (\cref{sec:independent}),
and models that are related to the conditional encoding architecture presented in \cref{sec:cond}. 
Results for these follow-up works are collected in a leaderboard at \url{http://nlp.stanford.edu/projects/snli/}.
The current best result\footnote{Checked last on 26th of April, 2017.} is held by a bidirectional \gls{LSTM} with matching and aggregation layers introduced by \cite{wang2017bilateral}. 
It achieves a test accuracy of $88.8\%$ and outperforms the best independent encoding model by $4.2$ percentage points.
In fact, most independent encoding models \citep[\eg{}][]{vendrov2016order,mou2016natural,bowman2016fast,munkhdalai2016neural} do not reach the performance of our conditional model with word-by-word attention. 
Exceptions are the recently introduced two-stage bidirectional \gls{LSTM} model by \cite{liu2016learning} and the Neural Semantic Encoder by \cite{munkhdalai2016neural_b}.

\subsection{Bidirectional Conditional Encoding}
\begin{figure}[t!]
  \centering
  \includegraphics[trim=4cm 0cm 4cm 0cm, width=1.0\textwidth]{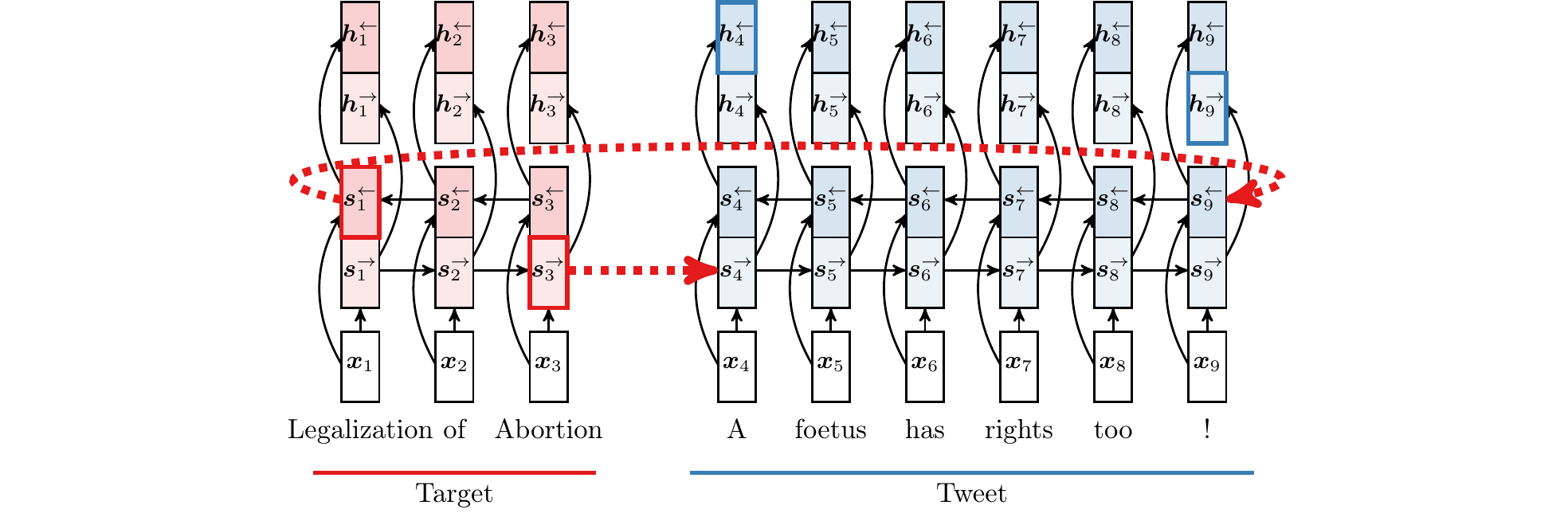}
  \caption{Bidirectional encoding of a tweet conditioned on bidirectional encoding of a target ($[\vec{s}^\rightarrow_3\;\vec{s}^\leftarrow_1]$). The stance is predicted using the last forward and reversed output representations ($[\vec{h}^\rightarrow_{9}\;\vec{h}^\leftarrow_4]$).}
  \label{fig:bidirectional}
\end{figure}

As the models presented in this chapter make little assumptions about the input data, they can be applied in other domains too.
\cite{augenstein2016stance} introduced a conditional encoding model for determining the stance of a tweet (\eg{} ``A foetus has rights too!'') with respect to a target (\eg{} ``Legalization of Abortion'').
They further extended the conditional encoding model with bidirectional \glspl{LSTM} \citep{graves2005framewise}, thus replacing \crefrange{eq:rte_bowman_p_cond}{eq:rte_h} with 
\begin{align}
   \ls{h}^{\overrightarrow{P}}, \ls{s}^{\overrightarrow{P}} &= \rnn\left(f^{\lstm}_{\theta_{\overrightarrow{P}}}, \ls{x}^{\overrightarrow{P}}, {\overrightarrow{\vec{s}}}_0\right)\\
   \ls{h}^{\overrightarrow{H}}, \ls{s}^{\overrightarrow{H}} &= \rnn\left(f^{\lstm}_{\theta_{\overrightarrow{H}}}, \ls{x}^{\overrightarrow{H}}, \vec{h}^{\overrightarrow{P}}_N\right)\\
   \ls{h}^{\overleftarrow{H}}, \ls{s}^{\overleftarrow{H}} &= \rnn\left(f^{\lstm}_{\theta_{\overleftarrow{H}}}, \ls{x}^{\overleftarrow{H}}, {\overleftarrow{\vec{s}}}_0\right)\\
   \ls{h}^{\overleftarrow{P}}, \ls{s}^{\overleftarrow{P}} &= \rnn\left(f^{\lstm}_{\theta_{\overleftarrow{P}}}, \ls{x}^{\overleftarrow{P}}, \vec{h}^{\overleftarrow{H}}_M\right)\\
   \vec{h} &= \tanh\left(\overrightarrow{\mat{W}}\vec{h}^{\overrightarrow{H}}_M + \overleftarrow{\mat{W}}\vec{h}^{\overleftarrow{P}}_N\right)
\end{align}
where $\overrightarrow{P}$ and  $\overrightarrow{H}$ denotes the forward, and  $\overleftarrow{P}$ and  $\overleftarrow{H}$ the reversed sequence, $\overrightarrow{\mat{W}}, \overleftarrow{\mat{W}}\in\R^{k\times k}$ are trainable projection matrices, and ${\overrightarrow{\vec{s}}}_0$ and ${\overleftarrow{\vec{s}}}_0$ are the trainable forward and reverse start state, respectively. 
This architecture is illustrated in \cref{fig:bidirectional}.
It achieved the second best result on the SemEval 2016 Task 6 Twitter Stance Detection corpus \citep{mohammad2016semeval}.

\subsection{Generating Entailing Sentences}
Instead of predicting the logical relationship between sentences, \cite{kolesnyk2016generating} used entailment pairs of the \gls{SNLI} corpus to learn to generate an entailed sentence given a premise. 
Their model is an encoder-decoder with attention, as used in neural machine translation \citep{bahdanau2014neural}.
On a manually annotated test corpus of $100$ generated sentences, their model generated correct entailed sentences in $82\%$ of the cases.
By recursively applying this encoder-decoder to the produced outputs, their model is able to generate natural language inference chains such as
``A wedding party looks happy on the picture $\Rightarrow$ A bride and groom smiles $\Rightarrow$ Couple smiling $\Rightarrow$ Some people smiling.''

\section{Summary}
In this chapter, we
demonstrated that \glspl{LSTM} that read pairs of sequences to produce a final representation from which a simple classifier predicts entailment, outperform an \gls{LSTM} baseline encoding the two sequences independently, as well as a classifier with hand-engineered features.
Besides contributing this conditional model for \gls{RTE}, our main contribution is to extend this model with attention over the premise which provides further improvements to the predictive abilities of the system.
\maybe{refer back to Raymond Mooney's quote in the beginning}
In a qualitative analysis, we showed that a word-by-word attention model is able to compare word and phrase pairs for deciding the logical relationship sentences.
With an accuracy of $84.5\%$, it held the state-of-the-art on a large \gls{RTE} corpus at the time of publication.

\chapter{Conclusions}
\label{conclusions}
\glsresetall

\section{Summary of Contributions}
In this thesis, we have presented various combinations of representation learning models with logic.

First, we proposed a way to calculate the gradient of propositional logic rules with respect to parameters of a neural link prediction model (\cref{log}).
By stochastically grounding first-order logic rules, we were able to use these rules as regularizers in a matrix factorization neural link prediction model for automated \gls{KB} completion.
This allowed us to embed background knowledge in form of logical rules in the vector space of predicate and entity pair representations.
Using this method, we were able to train relation extractors for predicates with provided rules but little or no known training facts.

In \cref{foil}, we identified various shortcomings of stochastic grounding and proposed a model in which implication rules are only used to regularize predicate representations.
This has the two advantages that the method becomes independent of the size of the domain of entity pairs, and that we can guarantee that the provided rules will hold for any test entity pair.
By restricting the entity pair embedding space to be non-negative, we were able to impose implications as a partial order on the predicate representation space similar to Order Embeddings \citep{vendrov2016order}.
We showed empirically that by restricting the entity pair embedding space, the model generalizes better to predicting facts in the test set, which we attribute to a regularization effect.
Furthermore, we showed that incorporating implication rules with this method scales well with the number of rules.

After investigating two ways of regularizing symbol representations based on rules, in \cref{ntp} we proposed a differentiable prover that performs \gls{KB} inference with symbol representations in a more explicit way.
To this end, we used Prolog's backward chaining algorithm as a recipe for recursively constructing neural networks that can be used to prove facts in a \gls{KB}. 
Specifically, we proposed a differentiable unification operation between symbol representations.
The constructed neural network allows us to compute the gradient of a proof success with respect to symbol representations, and thus train symbol representation end-to-end from the proof outcome.
Furthermore, given templates for unknown rules of predefined structure, we can induce first-order logic rules using gradient descent.
We proposed three optimizations for this model: 
(i) we implemented unification of multiple symbol representations as batch-operation which allows us to make use of modern \glspl{GPU} for efficient proving,
(ii) we proposed an approximation of the gradient by only following $K\max$ proofs, 
and (iii) we used neural link prediction models as regularizers for the prover to learn better symbol representations more quickly. 
On three out of four benchmark knowledge bases, our method outperforms ComplEx, a state-of-the-art neural link prediction model, while at the same time inducing interpretable rules.

Lastly, we developed neural models for \gls{RTE}, \ie, for determining the logical relationship between two natural language sentences (\cref{rte}).
We used one \gls{LSTM} to encode the first sentence, and then conditioned on that representation encoded the second sentence using a second \gls{LSTM} for deciding the label of the sentence pair.
Furthermore, we extended this model with a neural word-by-word attention mechanism that enables more fine-grained comparison of word and phrase pairs.
On a large \gls{RTE} corpus, these models outperform a classifier with hand-engineered features and a strong \gls{LSTM} baseline. 
In addition, we qualitatively analyzed the attention the model pays to words in the first sentence, and we were able to confirm the presence of fine-grained reasoning patterns.

\section{Limitations and Future Work}
The integration of neural representations with symbolic logic and reasoning remains an exciting and open research area, and we except to see much more systems improving representation learning models by taking inspiration from formal logic in the future.
While we demonstrated the benefit of regularizing symbol representations by logical rules for automated \gls{KB} completion, we were only able to do this efficiently for simple implication rules.
For future work, it would be interesting to use more general first-order logic rules as regularizers on predicate representations in a lifted way, for instance, by a more informed grounding of first-order rules.
However, it is likely that the approach of regularizing predicate representations using rules has theoretical limitations that need to be investigated further.
Thus, we believe an interesting alternative direction is synthesizing symbolic reasoning and neural representations in more explicit ways.

The end-to-end differentiable \gls{NTP} introduced in this thesis is only a first proposal towards a tight integration of symbolic reasoning systems with trainable rules and symbol representations.
The major obstacle that we encountered has to do with the computational complexity of making the proof success differentiable so that we can calculate the gradient with respect to symbol representations.
While it is possible to approximate the gradient by only maintaining the $K\max$ proofs for a given a query, at some point a unification of a query with all facts in a \gls{KB} is necessary. 
As real-world \glspl{KB} can contain millions of facts, this grounding becomes impossible to do efficiently without applying further heuristics even when using modern \glspl{GPU}.
A possible future direction could be the use of hierarchical attention \citep{andrychowicz2016learning}, or recent methods for reinforcement learning such as Monte Carlo tree search \citep{coulom2006efficient,kocsis2006bandit} as used, for instance, for learning to play Go \citep{silver2016mastering} or chemical synthesis planning \citep{segler2017towards}.
Specifically, the idea would be to train a model that learns to select promising rules instead of trying all rules for proving a goal.
Orthogonal to that, more flexible individual components of end-to-end differentiable provers are conceivable.
For instance, unification, rule selection, and rule application could be modeled as parameterized functions, and thus could be used to learn a more optimal behavior from data in a \gls{KB} than the behavior that we specified by closely following the backward chaining algorithm.
Furthermore, while the \gls{NTP} is constructed from Prolog’s backward chaining, we currently only support Datalog logic programs, \ie{}, function-free first-order logic.
An open question is how we can enable support for function terms in end-to-end differentiable provers.

Another open research direction is the extension of automated provers to handle natural language sentences, questions, and perform multi-hop reasoning with natural language sentences.
A starting point could be the combination of models that we proposed for determining the logical relationship between two natural language sentences, and the differentiable prover.
As the end-to-end differentiable prover introduced in this thesis can be used to calculate the gradient of proof success with respect to symbol representations, these symbol representations can itself be composed by an \acrshort{RNN} encoder that is trained jointly with the prover.
The vision is a prover that directly operates on natural language statements and explanations, avoiding the need for semantic parsing \citep{zettlemoyer2005learning}, \ie{}, parsing text into logical form.
As \glspl{NTP} decompose inference in a more explicit way, it would be worthwhile to investigate whether we can obtain  interpretable natural language proofs.
Furthermore, it would be interesting to scale the methods presented here to larger units of text such as entire documents.
Again, this needs model extensions such as hierarchical attention to ensure computational efficiency. 
In addition, it would be worthwhile exploring how other, more structured forms of attention \citep[\emph{e.g.}][]{graves2014neural,sukhbaatar2015end}, or other forms of differentiable memory \citep[\emph{e.g.}][]{grefenstette2015learning,joulin2015inferring} could help improve performance of neural networks for \gls{RTE} and differentiable proving.
Lastly, we are interested in applying \glspl{NTP} to automated proving of mathematical theorems, either in logical or natural language form, similar to the recent work by \cite{kaliszyk2017holstep} and \cite{loos2017deep}.

\addcontentsline{toc}{chapter}{Appendices}

\appendix

\chapter{Annotated Rules}
\label{appendixA}
Manually filtered rules and their score for experiments in \cref{log}.\\
\small
0.97	\rel{organization/parent/child}(\var{X},\var{Y}) \lif\\
\phantom{AAAA} \rel{\#2-nn<-unit->prep->of->pobj-\#1}(\var{X},\var{Y}).\\
0.97	\rel{organization/parent/child}(\var{X},\var{Y}) \lif\\
\phantom{AAAA} \rel{\#2->appos->subsidiary->prep->of->pobj-\#1}(\var{X},\var{Y}).\\
0.97	\rel{organization/parent/child}(\var{X},\var{Y}) \lif\\
\phantom{AAAA} \rel{\#2->rcmod->own->prep->by->pobj-\#1}(\var{X},\var{Y}).\\
0.97	\rel{location/location/containedby}(\var{X},\var{Y}) \lif\\
\phantom{AAAA} \rel{\#2-nn<-city->prep->of->pobj-\#1}(\var{X},\var{Y}).\\
0.97	\rel{organization/parent/child}(\var{X},\var{Y}) \lif\\
\phantom{AAAA} \rel{\#2->appos->subsidiary->nn-\#1}(\var{X},\var{Y}).\\
0.97	\rel{people/person/nationality}(\var{X},\var{Y}) \lif\\
\phantom{AAAA} \rel{\#2-poss<-minister->appos-\#1}(\var{X},\var{Y}).\\
0.97	\rel{organization/parent/child}(\var{X},\var{Y}) \lif\\
\phantom{AAAA} \rel{\#2->appos->unit->prep->of->pobj-\#1}(\var{X},\var{Y}).\\
0.96	\rel{organization/parent/child}(\var{X},\var{Y}) \lif\\
\phantom{AAAA} \rel{\#2->appos->division->prep->of->pobj-\#1}(\var{X},\var{Y}).\\
0.96	\rel{business/person/company}(\var{X},\var{Y}) \lif\\
\phantom{AAAA} \rel{\#2-poss<-executive->appos-\#1}(\var{X},\var{Y}).\\
0.96	\rel{business/company/founders}(\var{X},\var{Y}) \lif\\
\phantom{AAAA} \rel{\#2->appos->co-founder->prep->of->pobj-\#1}(\var{X},\var{Y}).\\
0.96	\rel{book/author/works\_written}(\var{X},\var{Y}) \lif\\
\phantom{AAAA} \rel{\#2-dobj<-review->prep->by->pobj-\#1}(\var{X},\var{Y}).\\
0.95	\rel{business/company/founders}(\var{X},\var{Y}) \lif\\
\phantom{AAAA} \rel{\#2->appos->founder->prep->of->pobj-\#1}(\var{X},\var{Y}).\\
0.95	\rel{location/location/containedby}(\var{X},\var{Y}) \lif\\
\phantom{AAAA} \rel{\#2-nn<-town->prep->of->pobj-\#1}(\var{X},\var{Y}).\\
0.95	\rel{location/neighborhood/neighborhood\_of}(\var{X},\var{Y}) \lif\\
\phantom{AAAA} \rel{\#2-nn<-neighborhood->prep->of->pobj-\#1}(\var{X},\var{Y}).\\
0.95	\rel{film/film/directed\_by}(\var{X},\var{Y}) \lif\\
\phantom{AAAA} \rel{\#2->appos->director->dep-\#1}(\var{X},\var{Y}).\\
0.95	\rel{location/location/containedby}(\var{X},\var{Y}) \lif\\
\phantom{AAAA} \rel{\#2-poss<-region->nn-\#1}(\var{X},\var{Y}).\\
0.94	\rel{film/film/produced\_by}(\var{X},\var{Y}) \lif\\
\phantom{AAAA} \rel{\#2->appos->producer->dep-\#1}(\var{X},\var{Y}).\\
0.94	\rel{film/film/directed\_by}(\var{X},\var{Y}) \lif\\
\phantom{AAAA} \rel{\#2-poss<-film->dep-\#1}(\var{X},\var{Y}).\\
0.94	\rel{location/location/containedby}(\var{X},\var{Y}) \lif\\
\phantom{AAAA} \rel{\#2-nsubj<-professor->prep->at->pobj-\#1}(\var{X},\var{Y}).\\
0.94	\rel{film/film/directed\_by}(\var{X},\var{Y}) \lif\\
\phantom{AAAA} \rel{\#2-poss<-movie->dep-\#1}(\var{X},\var{Y}).\\
0.93	\rel{people/person/nationality}(\var{X},\var{Y}) \lif\\
\phantom{AAAA} \rel{\#2-poss<-leader->appos-\#1}(\var{X},\var{Y}).\\
0.93	\rel{film/film/directed\_by}(\var{X},\var{Y}) \lif\\
\phantom{AAAA} \rel{\#2-nn<-film->dep-\#1}(\var{X},\var{Y}).\\
0.93	\rel{location/location/containedby}(\var{X},\var{Y}) \lif\\
\phantom{AAAA} \rel{\#2-nn<-suburb->prep->of->pobj-\#1}(\var{X},\var{Y}).\\
0.93	\rel{people/person/parents}(\var{X},\var{Y}) \lif\\
\phantom{AAAA} \rel{\#1->appos->daughter->prep->of->pobj-\#2}(\var{X},\var{Y}).\\
0.93	\rel{business/person/company}(\var{X},\var{Y}) \lif\\
\phantom{AAAA} \rel{\#2-poss<-chairman->appos-\#1}(\var{X},\var{Y}).\\
0.93	\rel{location/location/containedby}(\var{X},\var{Y}) \lif\\
\phantom{AAAA} \rel{\#2-nn<-side->prep->of->pobj-\#1}(\var{X},\var{Y}).\\
0.93	\rel{people/deceased\_person/place\_of\_death}(\var{X},\var{Y}) \lif\\
\phantom{AAAA} \rel{\#1-nsubj<-die->prep->in->pobj-\#2}(\var{X},\var{Y}).\\
0.93	\rel{location/neighborhood/neighborhood\_of}(\var{X},\var{Y}) \lif\\
\phantom{AAAA} \rel{\#2-poss<-neighborhood->nn-\#1}(\var{X},\var{Y}).\\
0.91	\rel{location/location/containedby}(\var{X},\var{Y}) \lif\\
\phantom{AAAA} \rel{\#2-nsubj<-professor->appos-\#1}(\var{X},\var{Y}).\\
0.91	\rel{people/deceased\_person/place\_of\_death}(\var{X},\var{Y}) \lif\\
\phantom{AAAA} \rel{\#1-nsubj<-die->prep->at->pobj->hospital->prep->in->pobj-\#2}(\var{X},\var{Y}).\\
0.91	\rel{book/author/works\_written}(\var{X},\var{Y}) \lif\\
\phantom{AAAA} \rel{\#1-poss<-book->dep-\#2}(\var{X},\var{Y}).\\
0.90	\rel{business/person/company}(\var{X},\var{Y}) \lif\\
\phantom{AAAA} \rel{\#2-nsubj<-name->dobj-\#1}(\var{X},\var{Y}).\\
0.90	\rel{people/person/place\_of\_birth}(\var{X},\var{Y}) \lif\\
\phantom{AAAA} \rel{\#1-nsubjpass<-bear->prep->in->pobj-\#2}(\var{X},\var{Y}).\\
0.90	\rel{people/person/nationality}(\var{X},\var{Y}) \lif\\
\phantom{AAAA} \rel{\#1->appos->minister->poss-\#2}(\var{X},\var{Y}).\\
0.88	\rel{location/location/containedby}(\var{X},\var{Y}) \lif\\
\phantom{AAAA} \rel{\#1->appos->capital->prep->of->pobj-\#2}(\var{X},\var{Y}).\\
0.87	\rel{location/location/containedby}(\var{X},\var{Y}) \lif\\
\phantom{AAAA} \rel{\#1-nsubj<-city->prep->in->pobj-\#2}(\var{X},\var{Y}).\\
\normalsize

\chapter{Annotated WordNet Rules}
\label{appendixB}
Manually filtered rules derived from WordNet for experiments in \cref{foil}.\\
\small
\rel{\#1->appos->organization->amod-\#2}(\var{X},\var{Y}) \lif\\
\phantom{AAAA} \rel{\#1->appos->party->amod-\#2}(\var{X},\var{Y}).\\
\rel{\#1-nsubj<-push->dobj-\#2}(\var{X},\var{Y}) \lif\\
\phantom{AAAA} \rel{\#1-nsubj<-press->dobj-\#2}(\var{X},\var{Y}).\\
\rel{\#1->appos->artist->amod-\#2}(\var{X},\var{Y}) \lif\\
\phantom{AAAA} \rel{\#1->appos->painter->amod-\#2}(\var{X},\var{Y}).\\
\rel{\#1->appos->artist->nn-\#2}(\var{X},\var{Y}) \lif\\
\phantom{AAAA} \rel{\#1->appos->painter->nn-\#2}(\var{X},\var{Y}).\\
\rel{\#1->appos->writer->amod-\#2}(\var{X},\var{Y}) \lif\\
\phantom{AAAA} \rel{\#1->appos->journalist->amod-\#2}(\var{X},\var{Y}).\\
\rel{\#1->appos->writer->amod-\#2}(\var{X},\var{Y}) \lif\\
\phantom{AAAA} \rel{\#1->appos->poet->amod-\#2}(\var{X},\var{Y}).\\
\rel{\#1-poss<-parent->appos-\#2}(\var{X},\var{Y}) \lif\\
\phantom{AAAA} \rel{\#1-poss<-father->appos-\#2}(\var{X},\var{Y}).\\
\rel{\#1->appos->lawyer->nn-\#2}(\var{X},\var{Y}) \lif\\
\phantom{AAAA} \rel{\#1->appos->prosecutor->nn-\#2}(\var{X},\var{Y}).\\
\rel{\#1->appos->expert->nn-\#2}(\var{X},\var{Y}) \lif\\
\phantom{AAAA} \rel{\#1->appos->specialist->nn-\#2}(\var{X},\var{Y}).\\
\rel{\#1->appos->newspaper->amod-\#2}(\var{X},\var{Y}) \lif\\
\phantom{AAAA} \rel{\#1->appos->daily->amod-\#2}(\var{X},\var{Y}).\\
\rel{\#1->appos->leader->nn-\#2}(\var{X},\var{Y}) \lif\\
\phantom{AAAA} \rel{\#1->appos->boss->nn-\#2}(\var{X},\var{Y}).\\
\rel{\#1->appos->firm->nn-\#2}(\var{X},\var{Y}) \lif\\
\phantom{AAAA} \rel{\#1->appos->publisher->nn-\#2}(\var{X},\var{Y}).\\
\rel{\#1->appos->journalist->nn-\#2}(\var{X},\var{Y}) \lif\\
\phantom{AAAA} \rel{\#1->appos->correspondent->nn-\#2}(\var{X},\var{Y}).\\
\rel{\#1->appos->company->amod-\#2}(\var{X},\var{Y}) \lif\\
\phantom{AAAA} \rel{\#1->appos->subsidiary->amod-\#2}(\var{X},\var{Y}).\\
\rel{\#1-nsubj<-purchase->dobj-\#2}(\var{X},\var{Y}) \lif\\
\phantom{AAAA} \rel{\#1-nsubj<-buy->dobj-\#2}(\var{X},\var{Y}).\\
\rel{\#1->appos->leader->nn-\#2}(\var{X},\var{Y}) \lif\\
\phantom{AAAA} \rel{\#1->appos->chief->nn-\#2}(\var{X},\var{Y}).\\
\rel{\#1->appos->player->poss-\#2}(\var{X},\var{Y}) \lif\\
\phantom{AAAA} \rel{\#1->appos->scorer->poss-\#2}(\var{X},\var{Y}).\\
\rel{\#1->appos->organization->nn-\#2}(\var{X},\var{Y}) \lif\\
\phantom{AAAA} \rel{\#1->appos->institution->nn-\#2}(\var{X},\var{Y}).\\
\rel{\#1->appos->center->amod-\#2}(\var{X},\var{Y}) \lif\\
\phantom{AAAA} \rel{\#1->appos->capital->amod-\#2}(\var{X},\var{Y}).\\
\rel{\#1->appos->center->poss-\#2}(\var{X},\var{Y}) \lif\\
\phantom{AAAA} \rel{\#1->appos->capital->poss-\#2}(\var{X},\var{Y}).\\
\rel{\#1->appos->representative->poss-\#2}(\var{X},\var{Y}) \lif\\
\phantom{AAAA} \rel{\#1->appos->envoy->poss-\#2}(\var{X},\var{Y}).\\
\rel{\#1->appos->expert->amod-\#2}(\var{X},\var{Y}) \lif\\
\phantom{AAAA} \rel{\#1->appos->specialist->amod-\#2}(\var{X},\var{Y}).\\
\rel{\#1->appos->center->nn-\#2}(\var{X},\var{Y}) \lif\\
\phantom{AAAA} \rel{\#1->appos->capital->nn-\#2}(\var{X},\var{Y}).\\
\rel{\#1->appos->writer->amod-\#2}(\var{X},\var{Y}) \lif\\
\phantom{AAAA} \rel{\#1->appos->novelist->amod-\#2}(\var{X},\var{Y}).\\
\rel{\#1->appos->diplomat->amod-\#2}(\var{X},\var{Y}) \lif\\
\phantom{AAAA} \rel{\#1->appos->ambassador->amod-\#2}(\var{X},\var{Y}).\\
\rel{\#1->appos->expert->amod-\#2}(\var{X},\var{Y}) \lif\\
\phantom{AAAA} \rel{\#1->appos->analyst->amod-\#2}(\var{X},\var{Y}).\\
\rel{\#1->appos->scholar->nn-\#2}(\var{X},\var{Y}) \lif\\
\phantom{AAAA} \rel{\#1->appos->historian->nn-\#2}(\var{X},\var{Y}).\\
\rel{\#1->appos->maker->amod-\#2}(\var{X},\var{Y}) \lif\\
\phantom{AAAA} \rel{\#1->appos->producer->amod-\#2}(\var{X},\var{Y}).\\
\rel{\#1->appos->maker->amod-\#2}(\var{X},\var{Y}) \lif\\
\phantom{AAAA} \rel{\#1->appos->manufacturer->amod-\#2}(\var{X},\var{Y}).\\
\rel{\#1->appos->official->amod-\#2}(\var{X},\var{Y}) \lif\\
\phantom{AAAA} \rel{\#1->appos->diplomat->amod-\#2}(\var{X},\var{Y}).\\
\rel{\#1->appos->trainer->poss-\#2}(\var{X},\var{Y}) \lif\\
\phantom{AAAA} \rel{\#1->appos->coach->poss-\#2}(\var{X},\var{Y}).\\
\rel{\#1->appos->member->amod-\#2}(\var{X},\var{Y}) \lif\\
\phantom{AAAA} \rel{\#1->appos->commissioner->amod-\#2}(\var{X},\var{Y}).\\
\rel{\#1->appos->institution->nn-\#2}(\var{X},\var{Y}) \lif\\
\phantom{AAAA} \rel{\#1->appos->company->nn-\#2}(\var{X},\var{Y}).\\
\rel{\#1->appos->representative->amod-\#2}(\var{X},\var{Y}) \lif\\
\phantom{AAAA} \rel{\#1->appos->envoy->amod-\#2}(\var{X},\var{Y}).\\
\rel{\#1->appos->scientist->nn-\#2}(\var{X},\var{Y}) \lif\\
\phantom{AAAA} \rel{\#1->appos->physicist->nn-\#2}(\var{X},\var{Y}).\\
\rel{\#1->appos->representative->nn-\#2}(\var{X},\var{Y}) \lif\\
\phantom{AAAA} \rel{\#1->appos->envoy->nn-\#2}(\var{X},\var{Y}).\\
\normalsize

\addcontentsline{toc}{chapter}{Bibliography}
\bibliographystyle{plainnat}
\bibliography{refs,./2_rte,./3_log,./4_foil,./4_ntp,refs2,refs3,extra}

\end{document}